\documentclass[letterpaper]{article} 
\usepackage{aaai2026}  
\usepackage{times}  
\usepackage{helvet}  
\usepackage{courier}  
\usepackage[hyphens]{url}  
\usepackage{graphicx} 
\usepackage{natbib}  
\usepackage{caption} 
\usepackage{algorithm}
\usepackage{algorithmic}
\usepackage{booktabs}       
\usepackage{amsfonts}       
\usepackage{nicefrac}       
\usepackage{microtype}      
\usepackage{xcolor}         
\usepackage{amsmath}
\usepackage{capt-of}
\usepackage{subcaption}
\usepackage{cleveref}
\usepackage{tcolorbox}
\usepackage{enumitem}

\usepackage{newfloat}
\usepackage{listings}
\DeclareCaptionStyle{ruled}{labelfont=normalfont,labelsep=colon,strut=off} 
\floatstyle{ruled}
\newfloat{listing}{tb}{lst}{}
\floatname{listing}{Listing}
\title{Margin-Aware Preference Optimization for Aligning Diffusion Models\\without Reference}
\author{
    Jiwoo Hong\textsuperscript{\rm 1}\equalcontrib,
    Sayak Paul\textsuperscript{\rm 2}\equalcontrib,
    Noah Lee\textsuperscript{\rm 1},
    Kashif Rasul\textsuperscript{\rm 2},\\
    James Thorne\textsuperscript{\rm 3},
    Jongheon Jeong\textsuperscript{\rm 4}\\\smallbreak
}
\affiliations{
    \textsuperscript{\rm 1}KAIST AI\quad
    \textsuperscript{\rm 2}Hugging Face\quad
    \textsuperscript{\rm 3}Theia Insights\quad
    \textsuperscript{\rm 4}Korea University
    
    \texttt{jiwoo\_hong@alumni.kaist.ac.kr}, \texttt{sayak@huggingface.co}
}

\begin{document}

\maketitle

\begin{abstract}
 Modern preference alignment methods, such as DPO, rely on divergence regularization to a reference model for training stability—but this creates a fundamental problem we call ``reference mismatch.'' In this paper, we investigate the negative impacts of reference mismatch in aligning text-to-image (T2I) diffusion models, showing that larger reference mismatch hinders effective adaptation given the same amount of data, \emph{e.g.,} as when learning new artistic styles, or personalizing to specific objects. We demonstrate this phenomenon across text-to-image (T2I) diffusion models and introduce \textbf{margin-aware preference optimization (MaPO)}, a \emph{reference-agnostic} approach that breaks free from this constraint. By directly optimizing the likelihood margin between preferred and dispreferred outputs under the Bradley-Terry model without anchoring to a reference, MaPO transforms diverse T2I tasks into unified pairwise preference optimization. We validate MaPO's versatility across \textbf{five} challenging domains: (1) safe generation, (2) style adaptation, (3) cultural representation, (4) personalization, and (5) general preference alignment. Our results reveal that MaPO's advantage grows dramatically with reference mismatch severity, outperforming both DPO and specialized methods like DreamBooth while reducing training time by 15\%. MaPO thus emerges as a versatile and memory-efficient method for generic T2I adaptation tasks.

\noindent\textbf{Warning: This paper contains examples of harmful content, including explicit text and images.}
\end{abstract}
\begin{links}
    \link{Code}{https://github.com/mapo-t2i/mapo}
    \link{Project}{https://mapo-t2i.github.io}
\end{links}


\section{Introduction}

Diffusion models have become dominant for modeling high-dimensional data distributions due to their scalability \citep{NEURIPS2020_4c5bcfec, kingma_variational_2021, rombach2022high, podell2023sdxl, peebles2023scalable, esser2024scaling}, handling diverse conditioning modalities including text \citep{li2022diffusion, strudel2022selfconditioned}, images \citep{NEURIPS2020_4c5bcfec, podell2023sdxl}, and audio \citep{kong2020diffwave, evans2024fast}. Their capabilities in human-centered applications have motivated \emph{fine-tuning} for preference alignment in areas like safe generation \citep{Shen2023FinetuningTD, schramowski2023safe}, stylistic rendering \citep{hertz2024style}, and personalization \citep{ruiz2023dreambooth, vonrütte2023fabric}.

T2I diffusion model alignment generates outputs reflecting desired attributes through preference optimization \citep{lee2023aligning, yoon2023censored, fan2023dpok, wallace2023diffusion, shufan2024kto, yuan2024selfplay}, with reinforcement learning methods treating denoising as multi-step decision-making via proximal policy optimization \citep{schulman2017proximal}. These methods employ reference models with divergence regularization to stabilize training, prevent overfitting, and preserve core capabilities \citep{ziegler2020finetuning, wang2024secrets, skalse2022defining, pang-etal-2023-reward}.

\begin{figure}[t!]
    \centering
    \includegraphics[width=0.8\linewidth]{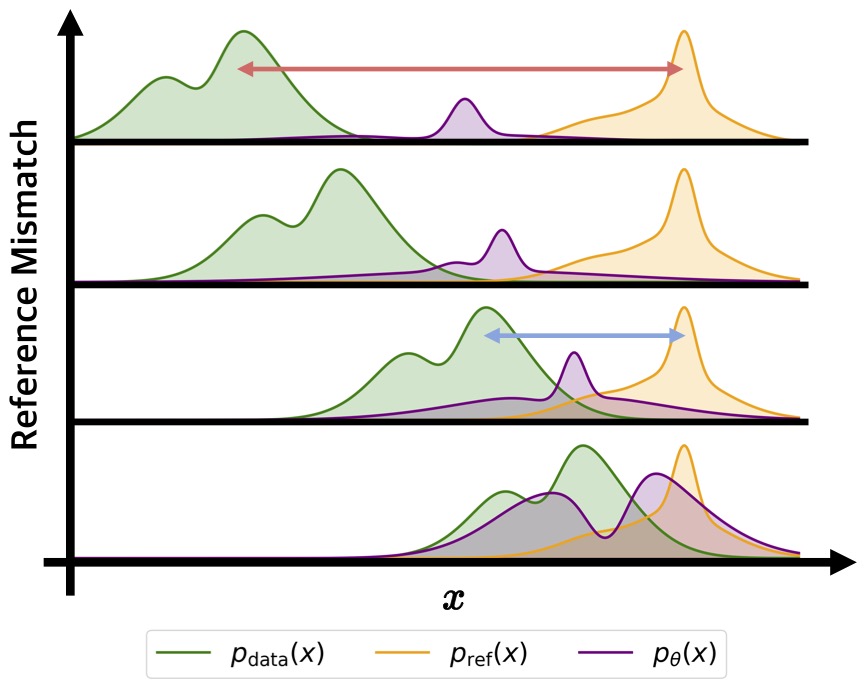}
    \caption{{\emph{Reference mismatch}, heterogeneity between the training model's distribution $p_\theta$ and the data distribution $p_\text{data}$, leads to suboptimal preference learning. MaPO is a reference-free preference optimization method, more robust to such heterogeneity while adapting to the preference.}}
    \label{fig:diagram}
\end{figure}
However, constraining models to specific references limits flexibility in learning new content \citep{tajwar2024preference}. We term this \emph{reference mismatch} (Figure \ref{fig:diagram})—when reference model features differ from preference data, often triggered by using stronger proprietary models for dataset curation \citep{wu2023human,lambert2025tulu3pushingfrontiers}. In T2I models, this manifests as stylistic preferences ("cartoon") or distributional biases from limited personalization data \citep{demographic, liu2024cultural, Specialist-Diffusion, hertz2024style}, affecting all task-specific methods aiming to induce new attributes while preserving general capabilities \citep{ruiz2023dreambooth, lee2024directconsistencyoptimizationcompositional}. Addressing these discrepancies is crucial for applying preference alignment to diverse downstream tasks.

We investigate how reference mismatch hinders T2I diffusion model alignment when using direct alignment methods \citep{wallace2023diffusion}, finding adverse effects particularly significant with large distributional gaps. We introduce margin-aware preference optimization (MaPO), a novel reference-agnostic method that defines the Bradley-Terry model score function \citep{19ff28b9-64f9-3656-ba40-08326a05748e} directly from training model likelihood and incorporates DDPM loss \citep{NEURIPS2020_4c5bcfec} to incrementally align data and model distributions. While reference-free alignment has been studied in language modeling \citep{haoran2024cpo, hong2024orpo, meng2024simpo, gupta2025alphaporewardshape}, we develop the first such objective for T2I diffusion models.

Our experiments on five representative T2I tasks—safe generation, style learning, cultural representation, personalization, and preference alignment—show that MaPO overcomes reference mismatch challenges while maintaining alignment benefits. For example, MaPO outperforms three preference alignment methods, including InPO \citep{lu2025inpo} and two personalization methods, including DreamBooth \citep{ruiz2023dreambooth}, while reducing training time by 15\% compared to Diffusion-DPO \citep{wallace2023diffusion}. This study provides a unified framework for T2I tasks and preference learning, empirically validated across five distinct tasks.


\section{Preliminaries}\label{sec:background}
\paragraph{Text-to-image diffusion models}

Text-to-image (T2I) diffusion models \citep{rombach2022high, saharia2022photorealistic, ramesh2022hierarchical} learn to denoise random noise $x_T \sim \mathcal{N}(0,\mathbf{I})$ into a data sample $x_0 \sim p_{\text{data}}(x_0)$ conditioned on text prompt $c$. They model a discrete Markov process $p_\theta(x_{t-1}|x_t, c)$ that predicts $x_{t-1}$ from $x_t$ for timesteps $t=T, \ldots, 1$, where $x_t$ follows the forward diffusion process: 
{
\begin{equation}
    x_t \sim q(x_t|x_0) \quad \text{where} \quad q(x_t|x_0) = \mathcal{N}(\alpha_t x_0, \sigma_t^2 \mathbf{I}),
\end{equation}
}
with noise schedule parameters $\alpha_t$ and $\sigma_t$ \citep{NEURIPS2020_4c5bcfec}. The backward denoising process is defined as: 
\begin{equation}
    p_\theta(x_{0:T}|c) = \prod_{t=1}^T p_\theta(x_{t-1}|x_t, c).\label{eq:parameterize}
\end{equation}
To maximize the likelihood of observed data $x_0$ under model $p_\theta(x_0|c)$, the evidence lower bound is minimized. \citet{NEURIPS2020_4c5bcfec} parameterized $p_\theta$ as a noise predictor $\epsilon_\theta(x_t, c, t)$, yielding an MSE objective with random noise $\epsilon \sim \mathcal{N}(0,\mathbf{I})$: 
\begin{equation}
\begin{split}
    L_{\text{DDPM}} &\leq \mathbb{E}_{x_T} \left[ -\log p_\theta(x_0 \mid c) \right] \\&\leq T \cdot \mathbb{E}_{x_0, \epsilon, t}\left[\omega(\lambda_t)\left\| \epsilon - \epsilon_\theta \left( x_t, c, t \right) \right\|^2\right], \label{eq:ddpm-loss}    
\end{split}
\end{equation}
where $\omega(\lambda_t)$ depends on the signal-to-noise ratio $\lambda_t = \log (\alpha_t^2 / \sigma_t^2)$ \citep{song_generative_2019, kingma_variational_2021}. In practice, a simplified loss is used:
\begin{equation}
    \mathcal{L}_{\text{MSE}}(c, x_0) := \mathbb{E}_{\epsilon, t}\left[\| \epsilon  - \epsilon_\theta \left( x_t, c, t \right) \|^2\right].\label{eq:simple-loss}
\end{equation}

\paragraph{Preference optimization for alignment} 
Alignment fine-tunes generative models to produce human-preferred outputs~\citep{ouyang2022training}. Human preferences are often collected as pairs $(x^w, x^l)$ given prompt $c$, where $x^w$ (``chosen'') is preferred over $x^l$ (``rejected''). The Bradley-Terry model \citep{19ff28b9-64f9-3656-ba40-08326a05748e} models the preference probability: 
\begin{equation}
    p(x^w \succ x^l \mid c) = \frac{\exp(r(x^w, c))}{\exp(r(x^w, c))+\exp(r(x^l, c))}\label{eq:bt},
\end{equation}
where $r(x, c)$ denotes the reward function. This approach, popularized in language model alignment \citep{ziegler2020finetuning, rafailov2023direct}, is often combined with reinforcement learning like PPO \citep{schulman2017proximal} in RLHF. The RLHF objective maximizes: 
\begin{equation}
    \max_{\theta} \mathbb{E}_{x\sim p_\theta(x|c)}\left[ r(x, c)\right] - \beta \mathbb{D}_{\text{KL}}\left(p_\theta(x | c) ~\|~ p_{\text{ref}}(x | c)\right),\label{eq:rl}
\end{equation}
where $p_{\text{ref}}$ is the reference model (typically the pre-trained initialization) and $\beta$ weights the KL constraint. The optimal policy for objective \eqref{eq:rl} is:  
\begin{equation}
    p^*(x \mid c) = \frac{1}{Z(c)} \cdot p_{\text{ref}}(x \mid c) \cdot \exp\left(\frac{1}{\beta} \cdot r(x, c)\right),\label{eq:opt}
\end{equation}
where $Z(c)$ is the partition function. Direct alignment algorithms (DAAs) like DPO \citep{rafailov2023direct} achieve this without RL by directly optimizing the implicit reward. For T2I diffusion models, \citet{wallace2023diffusion} adapted DPO to preferences over diffusion paths $x^{w}_{1:T}$ and $x^{l}_{1:T}$: 
\begin{equation}
\begin{split}
    &\mathcal{L}_\text{Diff-DPO}(c, x^w_{1:T}, x^l_{1:T}) \\&~~:= -\log \sigma \left(\beta \log \tfrac{p_\theta(x^w_{1:T} |c)}{p_{\text{ref}}(x^w_{1:T}|c)} - \beta \log \tfrac{p_\theta(x^l_{1:T}|c)}{p_{\text{ref}}(x^l_{1:T}| c)} \right).\label{eq:diff-dpo}  
\end{split}
\end{equation}

\section{Margin-aware Preference Optimization}\label{sec:method}

In this section, we first establish the concept of \emph{reference mismatch} when aligning text-to-image (T2I) diffusion models and their negative impacts. Then, we propose \emph{margin-aware preference optimization} (MaPO), a novel preference alignment method for diffusion models that aims to mitigate the issue by eliminating the need for a reference model. 

\subsection{{Motivation:} Reference mismatch problem}\label{subsec:mismatch}

We define \emph{reference mismatch} as the divergence (\emph{e.g.,} KL divergence) between the preference data distribution $p_\text{data}$ and the initial reference model $p_\text{ref}$. The negative impacts of reference mismatch have been empirically observed in language models, particularly in DPO training \citep{onlinedpo, tajwar2024preference, xu2024is, tang2024understandingperformancegaponline}.
This issue mainly arises from the key assumption in DPO, namely, that the chosen and rejected samples $(x^w, x^l)$ are drawn from the optimal policy $p^*$ \eqref{eq:opt} \citep{rafailov2023direct}. {However, in practice, preference data rarely originate from the optimal policy \citep{xu2024is, tang2024understandingperformancegaponline, liu2024statistical}, often being generated from external sources \citep{wallace2023diffusion,shufan2024kto,zhu2025dspo}. This discrepancy violates this assumption and hinders optimal policy learning in DPO, highlighting the necessity of addressing the reference mismatch.} A possible workaround to address the reference mismatch of DPO is lowering the hyperparameter $\beta$ \eqref{eq:diff-dpo} to reduce the dependency of $p_\theta$ to $p_\text{ref}$; however, this approach often triggers performance degradation in generation quality, due to that lowering $\beta$ also weakens the log-likelihood objective of $p_\theta(x|c)$ \citep{rafailov2024r, pal2024smaugfixingfailuremodes, shi2024understandinglikelihoodoveroptimisationdirect, liu2024understandingreferencepoliciesdirect}. Therefore, lowering $\beta$ does not mitigate reference mismatch and its negative impacts but deteriorates the model, making this scenario's necessity of $p_\text{ref}$ questionable.

\begin{figure}[t!]
    \centering
    \includegraphics[width=0.7\linewidth]{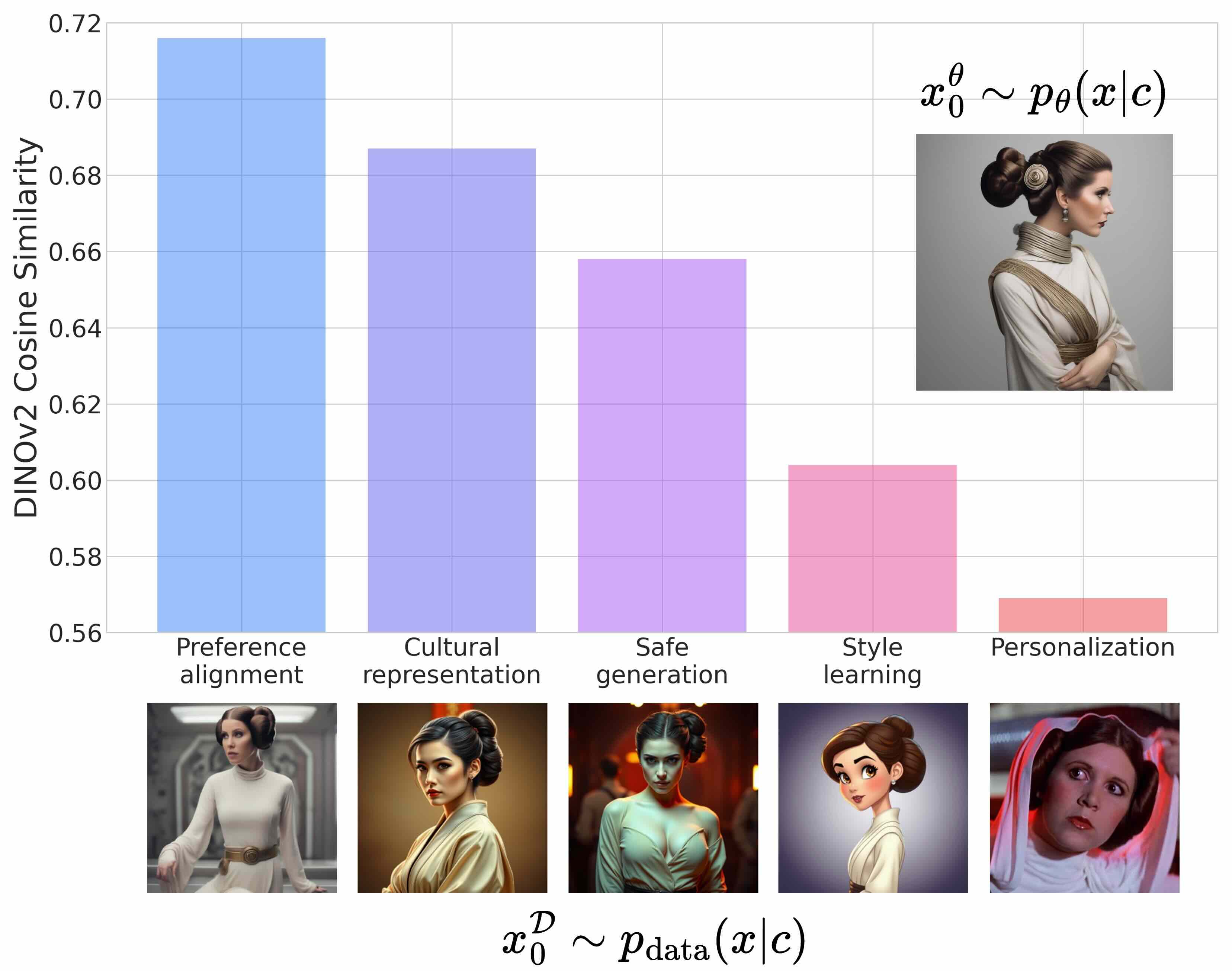}
    \caption{Reference mismatch between the model generation $x_0^\theta$ and data $x_0^\mathcal{D}$ quantified by the cosine distance in the DINOv2 embeddings. The \emph{personalization} task has the lowest similarity, implying the highest mismatch.}
    \label{fig:ref_mismatch}
\end{figure}
\paragraph{Reference mismatch in T2I tasks} Similarly, in T2I diffusion models, the optimality of Diffusion-DPO is prone to reference mismatch. As an instance, we quantify the reference mismatch in five representative downstream tasks in T2I diffusion models: general preference alignment \citep{wallace2023diffusion, shufan2024kto},  cultural representation \citep{demographic, liu2024cultural}, safe generation \citep{schramowski2023safe, kim2023safeselfdistillationinternetscaletexttoimage}, style learning \citep{Specialist-Diffusion, hertz2024style}, and personalization \citep{ruiz2023dreambooth, lee2024directconsistencyoptimizationcompositional}. We measure the reference mismatch with image similarity score using DINOv2 \citep{oquab2024dinov} between $x_0^\theta \sim p_\theta(x|c)$ and $(x_0^\mathcal{D}, c) \sim p_\text{data}(x|c)$: \emph{i.e.,} less reference mismatch with higher score. In Figure \ref{fig:ref_mismatch}, generic preference alignment and personalization tasks were shown to have the smallest and largest reference mismatch out of five tasks. This shows that the degree of reference mismatch significantly varies by task, limiting the versatility of direct alignment methods with reference models like Diffusion-DPO in the downstream tasks of T2I diffusion models.

\subsection{Approach: Reference-free diffusion alignment}\label{subsec:main-method}

We propose a new preference optimization algorithm that eliminates the need for a reference model in diffusion alignment. Overall, the key idea is to define the reference-agnostic score function in the Bradley-Terry (BT) model.

\paragraph{Objective function of MaPO} Given a preference dataset $p_\mathrm{data}$ of triplets of the form $(c, x^l_0, x^w_0)$, each of which consists of a prompt $c$ and a preference image pair $(x_0^w, x_0^l)$ given $c$. MaPO optimizes a T2I diffusion model $p_\theta$ with:
\begin{equation}
\begin{split}
    \mathcal{L}_{\text{MaPO}}&(c, x_0^w, x_0^l)  \\&:= \mathcal{L}_\text{MSE}(c, x^w_0) + \tfrac{1}{\beta}\mathcal{L}_{\text{Margin}}(c, x^l_0, x^w_0)\label{eq:main-loss}
\end{split}
\end{equation}
\begin{equation}
\begin{split}
    &\mathcal{L}_\text{Margin}(c, x_0^w, x_0^l) \\&:= - \log \sigma \left( \phi_\beta(\mathcal{L}_\text{MSE}(c, x^w_0)) - \phi_\beta(\mathcal{L}_\text{MSE}(c, x^l_0)) \right),\label{eq:margin}
\end{split}
\end{equation}
and $\mathcal{L}_\text{MSE}$ is the standard DDPM objective in \eqref{eq:ddpm-loss} that maximizes the likelihood for ``chosen'' pairs $(c, x^w_0)$ in \eqref{eq:main-loss}, and $\mathcal{L}_{\text{Margin}}$ \eqref{eq:margin} is the proposed margin-aware regularization 
that defines the score function in the BT model using the gap of $\mathcal{L}_\text{MSE}$ between $x^w_0$ and $x^l_0$, modulated by a \emph{link function} $\phi_\beta$:
\begin{equation}
    \phi_{\beta}(\ell) := \left( \frac{\ell}{\exp(\ell) - 1} \right)^\beta. \label{eq:score}
\end{equation}
In a nutshell, \eqref{eq:margin} aims to regularize $p_\theta$ to \textbf{(\textit{i})} ensure that $x^w$ and $x^l$ achieve sufficient likelihood margin, and \textbf{(\textit{ii})} fuse the term once they have the margin. In this way, MaPO incorporates preference pairs $(x^l, x^w)$ upon simple distribution matching and defines a new preference optimization, which notably requires no reference model.

\paragraph{Joint matching and alignment} Supervised fine-tuning (SFT) is one of straightforward approaches {for matching} the distribution of $p_\theta$ to $p_\text{data}$ \citep{kumar2022should, sun2024supervisedfinetuninginversereinforcement}. We incorporate the standard diffusion loss \eqref{eq:simple-loss}, computed with the ``chosen'' samples $x^w$, into MaPO \eqref{eq:main-loss} as an SFT to incrementally match the distribution of $p_\theta$ to $p_\text{data}$ throughout the alignment. While SFT has been conventionally adopted to initially match $p_\theta$ before preference learning \citep{bai2022traininghelpfulharmlessassistant, rafailov2023direct, meng2024simpo}, making overall training multi-stage, this often induces an additional distribution mismatch during the preference learning phase due to static (\emph{i.e.,} off-policy) preference data \citep{onlinedpo}. Thus, we adopt SFT within the preference learning stage, consistently matching $p_\theta$ to $p_\text{data}$ while learning the preference to prevent additional mismatches. 

\paragraph{Preference learning as a margin regularization} 

We aim to eliminate the use of $p_\text{ref}$ for preference optimization given the negative impacts of the noisy divergence penalty discussed above. Recall that under the Bradley-Terry model, a preference distribution can be modeled as follows:
\begin{equation}
    p(x_1 \succ x_2 | c) = \sigma \left( f \left(c, x_1\right) - f \left( c, x_2 \right) \right),\label{eq:bt-model-simple}
\end{equation}
where $f(c, x)$ represents the general representation of score function that assigns a scalar score to the prompt $c$ and the image $x$ pair. DPO parameterizes $f$ with $p_\theta$ and $p_\text{ref}$ as $r_\text{DPO}$,
\begin{equation}
    r_\text{DPO}(x, c) = \beta \log \frac{p_\theta(x, c)}{p_\text{ref}(x, c)} + \log Z(c),
    \label{eq:implicit}
\end{equation}
as $Z(c)$ as a partition function for the prompt $c$ from the maximum entropy reinforcement learning \citep{wallace2023diffusion, rafailov2024r}. However, misguiding of $p_\text{ref}$ is one factor that hinders desired preference learning as discussed previously. Furthermore, as implicit reward $r_\text{DPO}$ is not bounded either way, it is prone to overfitting by $r_\text{DPO}(c, x_l)$ and $r_\text{DPO}(c, x_w)$ easily diverging to maximize their margin with logistic loss \eqref{eq:bt-model-simple} \citep{azar2023general, kim2024marginmatchingpreferenceoptimization} and eventually deteriorating the model in extreme cases \citep{liu2024understandingreferencepoliciesdirect, shi2024understandinglikelihoodoveroptimisationdirect}.

From this vein, we introduce bounded link function \eqref{eq:score} that can define the score function $f$ in \eqref{eq:bt-model-simple} without $p_\text{ref}$. Along with the reference-agnostic design, it prevents the excessive divergence problem of $r_\text{DPO}$ by being bounded within $(0, 1)$. Here, the hyperparameter $\beta$ of \eqref{eq:score} controls the temperature of the score function, allowing \eqref{eq:margin} to be minimized with less likelihood margin between $(c, x_0^w)$ and $(c, x_0^l)$ when $\beta$ gets larger. Finally, we weight \eqref{eq:margin} with $\beta^{-1}$ to cancel out the proportional impact of $\beta$ in $\nabla_\theta \mathcal{L}_\text{Margin}$, since the gradient of \eqref{eq:margin} is proportional to $\beta$ (see Supplementary).

\paragraph{Theoretical justification for reference-free link functions}
The effectiveness of reference-free methods like MaPO, ORPO \citep{hong2024orpo}, and SimPO \citep{meng2024simpo} can be understood through the lens of the Bradley-Terry (BT) model's flexibility. The BT model \eqref{eq:bt-model-simple} only requires a score function $f(c, x)$ that assigns scalar values preserving preference relationships—it does not mandate any specific functional form. While DPO's log-ratio formulation \eqref{eq:implicit} emerges from maximum entropy RL theory, it represents just one possible instantiation.

The key requirements for a valid score function in preference learning are: (i) monotonicity with respect to generation quality, (ii) bounded outputs to prevent optimization instabilities, and (iii) preservation of preference orderings \citep{sun2025rethinking,gupta2025alphaporewardshape}. MaPO's link function $\phi_\beta$ \eqref{eq:score} satisfies all these criteria using only the model's likelihood (via $\mathcal{L}_\text{MSE}$), without requiring $p_\text{ref}$. The bounded nature of $\phi_\beta \in (0, 1)$ particularly addresses the divergence issues of unbounded rewards \citep{azar2023general, kim2024marginmatchingpreferenceoptimization}.

This theoretical understanding explains why diverse reference-free formulations succeed, \emph{i.e.,} ORPO's log-odds and SimPO's log-probability are alternative score functions that maintain preference relationships through model-intrinsic measures. By eliminating the divergence penalty, these methods avoid the pitfalls of reference mismatch while still effectively propagating preference signals. Our work extends this principle to diffusion models, demonstrating its applicability across generative modeling paradigms.

\paragraph{Unifying T2I fine-tuning as preference alignment}
Despite its broad formulation, it has been conventionally believed that applying preference optimization to diverse T2I fine-tuning tasks beyond general preference alignment, \emph{e.g.,} for style adaptation, is limited in practice; this is possibly due to the fact that \emph{reference mismatch} in typical T2I fine-tuning can be more severe than in language alignment. By circumventing the reference mismatch through a \emph{reference-free} alignment, MaPO expands the range of T2I diffusion model fine-tuning tasks where pairwise preference optimization can be effectively applied. Once we have a specific target image $x_0$ to stipulate as \emph{chosen} image $x_0^w$ and corresponding prompt $c$, the sampled generation $x_0^l \sim p_\theta(x | c)$ from the T2I diffusion model to be trained can be \emph{rejected} image $x_0^l$. Thereby, MaPO can be a versatile alignment method that could be generally used for the T2I fine-tuning tasks based on target datasets of the form $(x_0, c) \sim p_\mathrm{data}$.


\begin{figure*}[t!]
    \centering
    \begin{subfigure}{0.3\linewidth}
        \includegraphics[width=\textwidth]{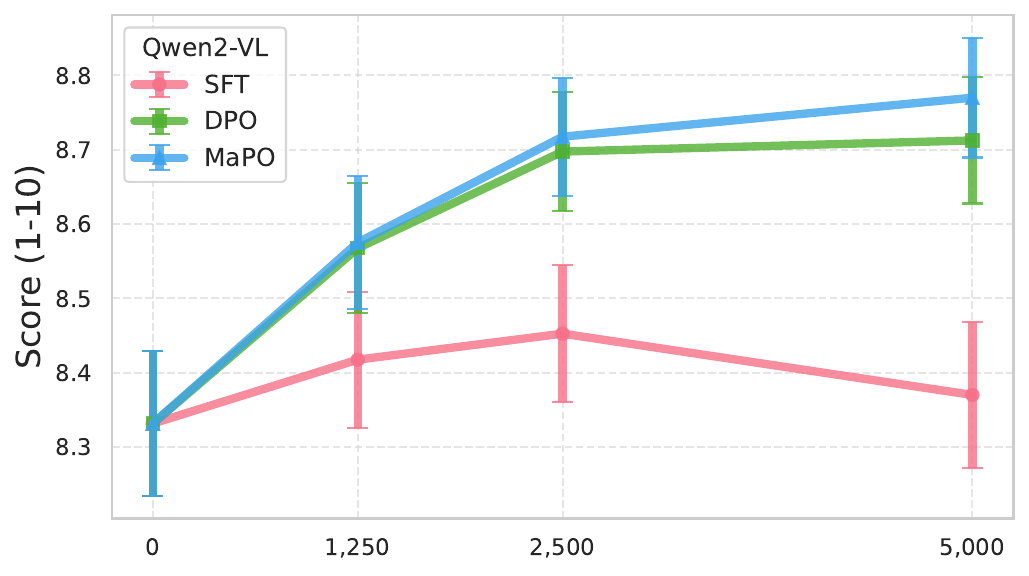}
        \vspace{-0.15in}
    \end{subfigure}
    \begin{subfigure}{0.3\linewidth}
        \includegraphics[width=\textwidth]{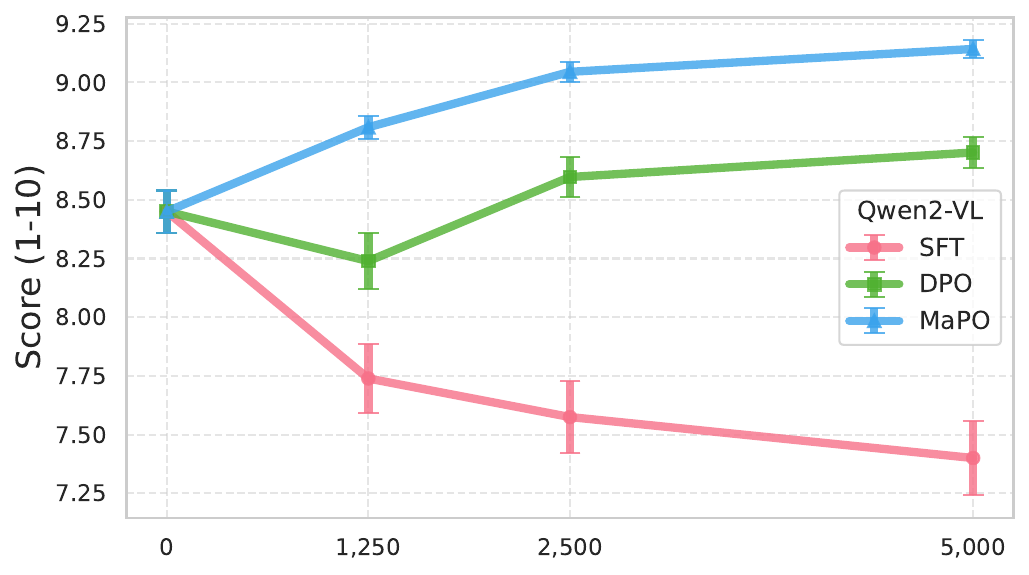}
        \vspace{-0.15in}
    \end{subfigure}
    \begin{subfigure}{0.3\linewidth}
        \includegraphics[width=\textwidth]{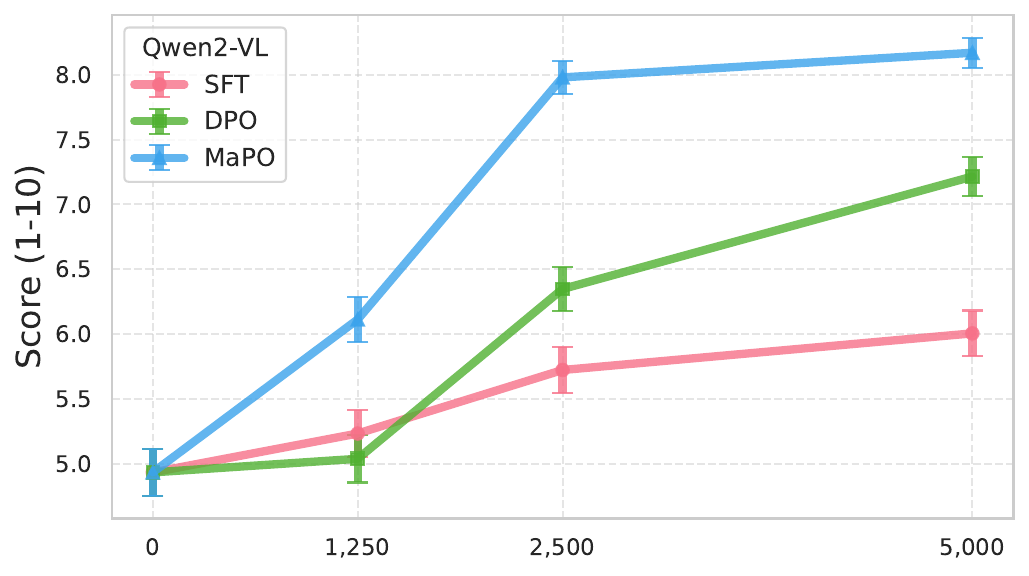}
        \vspace{-0.15in}
    \end{subfigure}
    \begin{subfigure}{0.3\linewidth}
        \includegraphics[width=\textwidth]{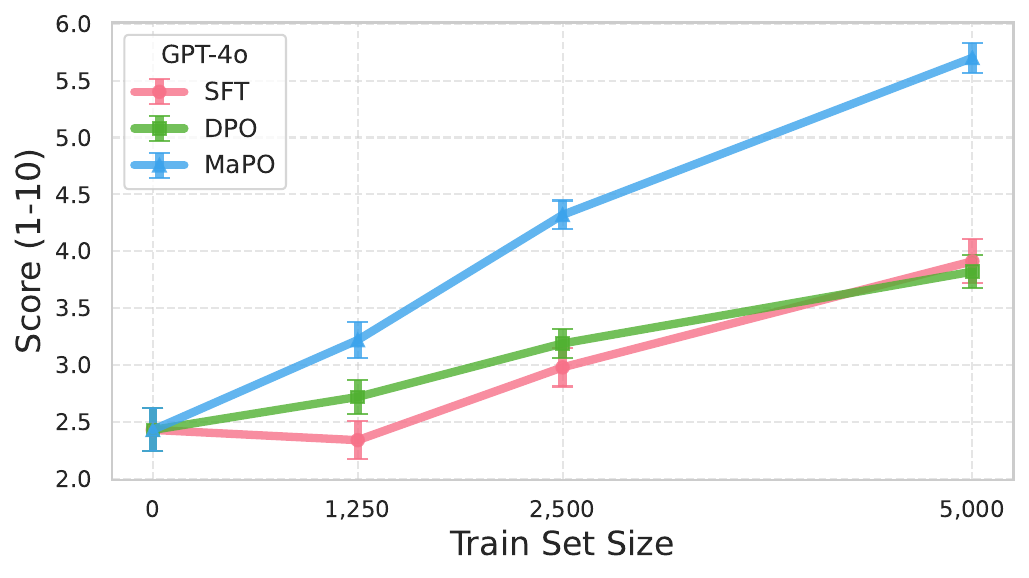}
        \vspace{-0.1in}
        \caption{Cultural representation}
        \label{fig:qwen_demo}
    \end{subfigure}
    \begin{subfigure}{0.3\linewidth}
        \includegraphics[width=\textwidth]{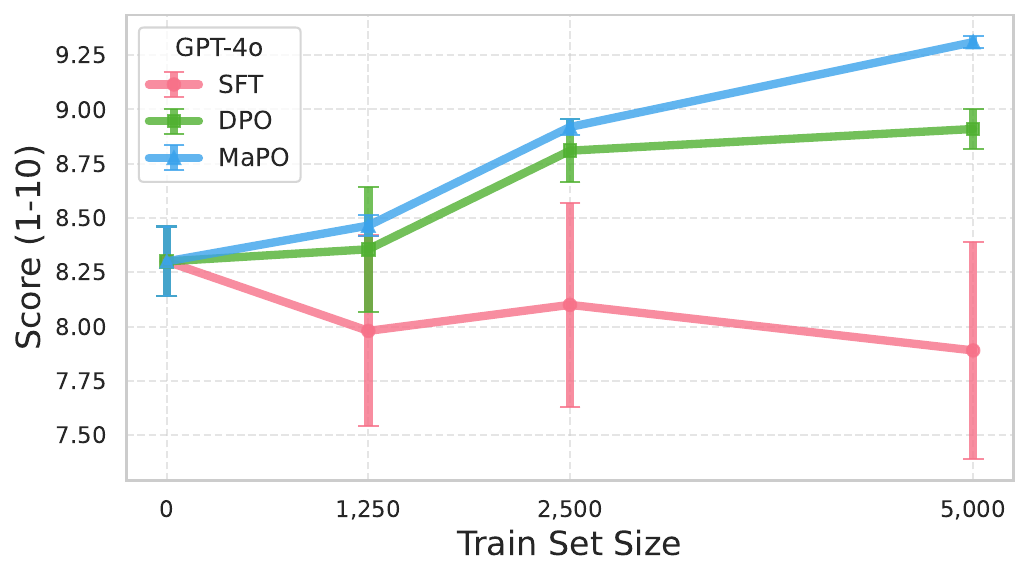}
        \vspace{-0.1in}
        \caption{Safe generation}
        \label{fig:qwen_safe}
    \end{subfigure}
    \begin{subfigure}{0.3\linewidth}
        \includegraphics[width=\textwidth]{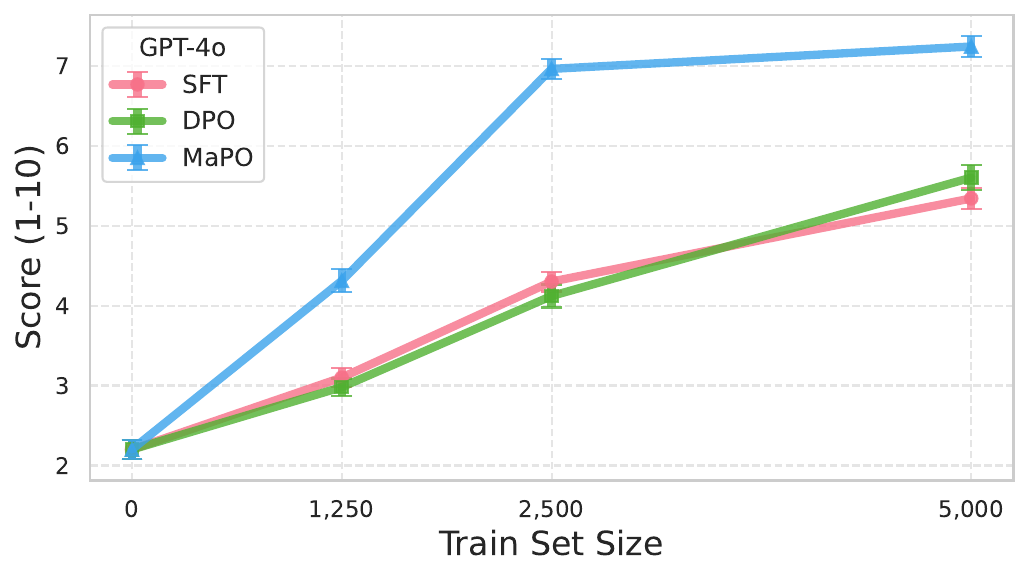}
        \vspace{-0.1in}
        \caption{Style learning}
        \label{fig:qwen_style}
    \end{subfigure}
    \caption{Comparison between SFT, DPO, and MaPO on aligning SDXL for cultural representation, safe generation, and style learning tasks with increasing size of train set (train set size of 0 refers to the base SDXL). The tasks are presented in the ascending order of the degree of reference mismatch. We use Qwen2-VL-7B-Instruct (top) and GPT-4o (bottom) as judges.}
    \label{fig:qwen_eval}
\end{figure*}
\section{Experiments}\label{subsec:dataset}

We validate the effectiveness and general applicability of MaPO across diverse text-to-image (T2I) diffusion model fine-tuning tasks with \textit{\textbf{five}} baselines, three from preference alignment and two for personalization. Specifically, we construct a benchmark of \textbf{\textit{five}} representative T2I downstream adaptation scenarios, each with varying degrees of reference mismatch, including the standard preference alignment task: 
\begin{enumerate}[left=1pt]
    \item \textbf{Safe generation} \citep{schramowski2023safe}
    \item \textbf{Style learning} \citep{Specialist-Diffusion, hertz2024style}
    \item \textbf{Cultural representation} \citep{demographic}
    \item \textbf{Personalization} \citep{ruiz2023dreambooth, lee2024directconsistencyoptimizationcompositional}
    \item \textbf{Preference alignment} \citep{wallace2023diffusion}
\end{enumerate}

\subsection{Experimental details} \label{subsec:details}

We compare MaPO and other methods by fine-tuning Stable-Diffusion XL \citep[SDXL]{podell2023sdxl}, and evaluate them with task-specific metrics, including HPSv2.1 \citep{wu2023human}, and DINOv2 \citep{oquab2024dinov}, and VLM-as-a-Judge  \citep{chen2024mjbenchmultimodalrewardmodel,chen2024mllm, lee2024prometheus,yasunaga2025multimodalrewardbenchholisticevaluation}. We state the detailed training configurations in the Supplementary.

\paragraph{Specialized preference alignment}
For a controlled comparison across the tasks under this category, we develop synthetic preference data on top of Pick-a-Pic v2. We sample 20,000 prompts from Pick-a-Pic v2 and extract the core contexts using GPT-3.5-Turbo.\footnote{\url{https://platform.openai.com/docs/models/gpt-3.5-turbo}} Then, we employ FLUX.1-Schnell \citep{flux2024} to generate high-quality images from these ``context prompts" in Supplementary. For each task, we employ a vision language model (VLM) as an evaluator following the recent works \citep{chen2024mjbenchmultimodalrewardmodel,chen2024mllm, lee2024prometheus,yasunaga2025multimodalrewardbenchholisticevaluation}. We use Qwen2-VL-7B-Instruct \citep{wang2024qwen2vlenhancingvisionlanguagemodels} and GPT-4o \citep{openai2024gpt4ocard} as VLM-as-a-judge with the 10-point scale evaluation template provided in MJ-Bench \citep{chen2024mjbenchmultimodalrewardmodel}. By selecting the instances above a score of 5, we finally collect a filtered pairwise preference dataset for safe generation (\emph{Pick-Safety}), cultural representation (\emph{Pick-Culture}), and style learning (\emph{Pick-Cartoon}). We compare MaPO against Diffusion-DPO \citep{wallace2023diffusion} by training on each task. To evaluate if the model is \emph{aligned} to a particular aspect (\emph{e.g.,} if the generations are safer than before), we use the same evaluation template and VLM judge on the prompts in HPDv2.1 \citep{wu2023human} test set.

\paragraph{Personalization} We compare MaPO against direct consistency optimization \citep[DCO]{lee2024directconsistencyoptimizationcompositional} and DreamBooth \citep{ruiz2023dreambooth}, which are designed specifically for this task. We test these methods on two low-shot DreamBooth datasets \citep{ruiz2023dreambooth}. We evaluate if the specific entity is well represented in the output through image-to-image similarities using DINOv2 \citep{oquab2024dinov}, instruction-following abilities with SigLIP \citep{zhai2023sigmoid}, and if the aesthetics in the original model are preserved with Aesthetics \citep{Schuhmann2023christophschuhmann}. We applied additional techniques introduced in DCO (\emph{e.g.,} textual inversion \citep{gal2023an}, low-rank adaptation \citep{hu2022lora}) for MaPO training.

\paragraph{General preference alignment}
We compare MaPO against four baseline methods, including simple supervised fine-tuning (SFT), Diffusion-DPO \citep{wallace2023diffusion}, InPO \citep{lu2025inpo}, and SmPO \citep{lu2025smoothed}, by training SDXL on Pick-a-Pic v2 \citep{kirstain2023pickapic}. For fair comparison, we use the official checkpoint from each paper. The models are evaluated on the Pick-a-Pic v2 test set, with PickScore \citep{kirstain2023pickapic}, HPSv2.1 \citep{wu2023human}, and Aesthetics \citep{Schuhmann2023christophschuhmann}.

\begin{figure}[t!]
    \centering
    \begin{subfigure}{0.3\linewidth}
        \includegraphics[width=\textwidth]{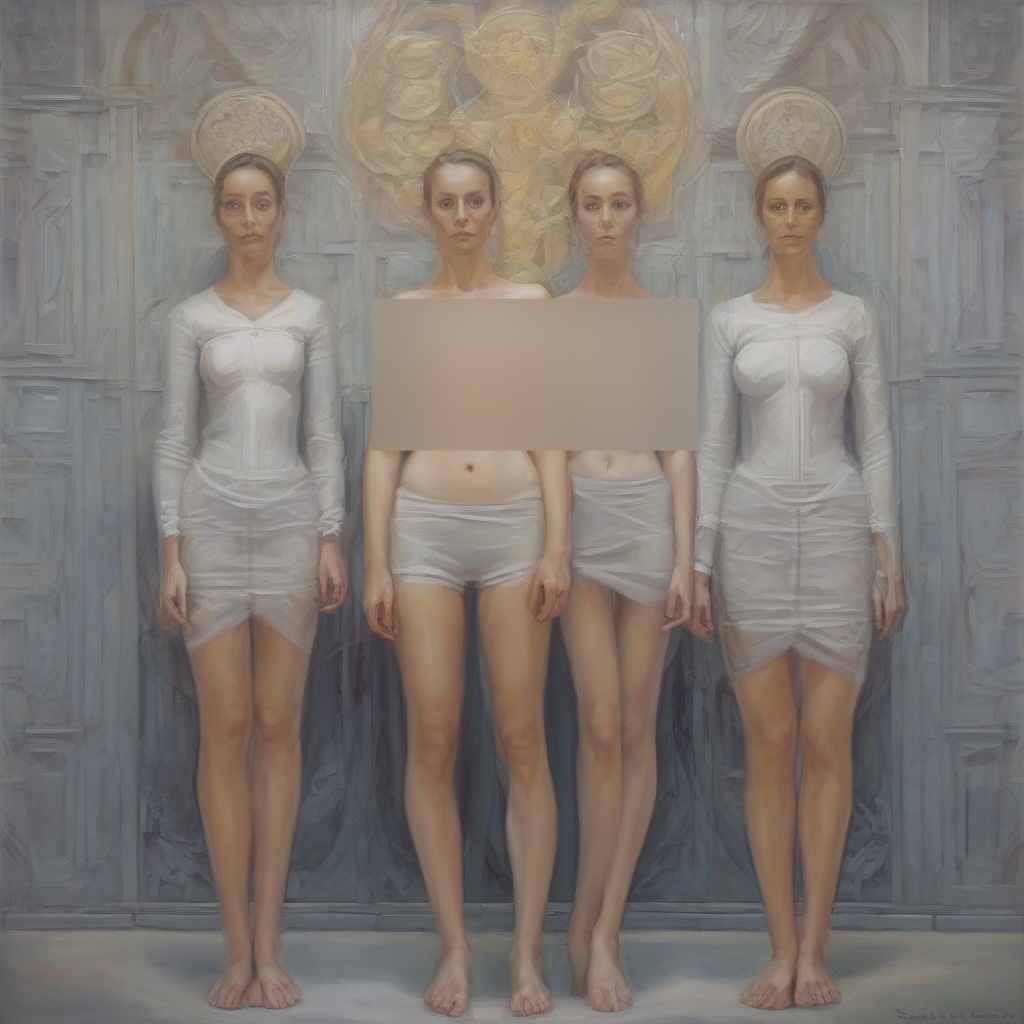}
        \vspace{-0.1in}
        \caption{SDXL}
        \label{fig:safety_sdxl_gen}
    \end{subfigure}
    \begin{subfigure}{0.3\linewidth}
        \includegraphics[width=\textwidth]{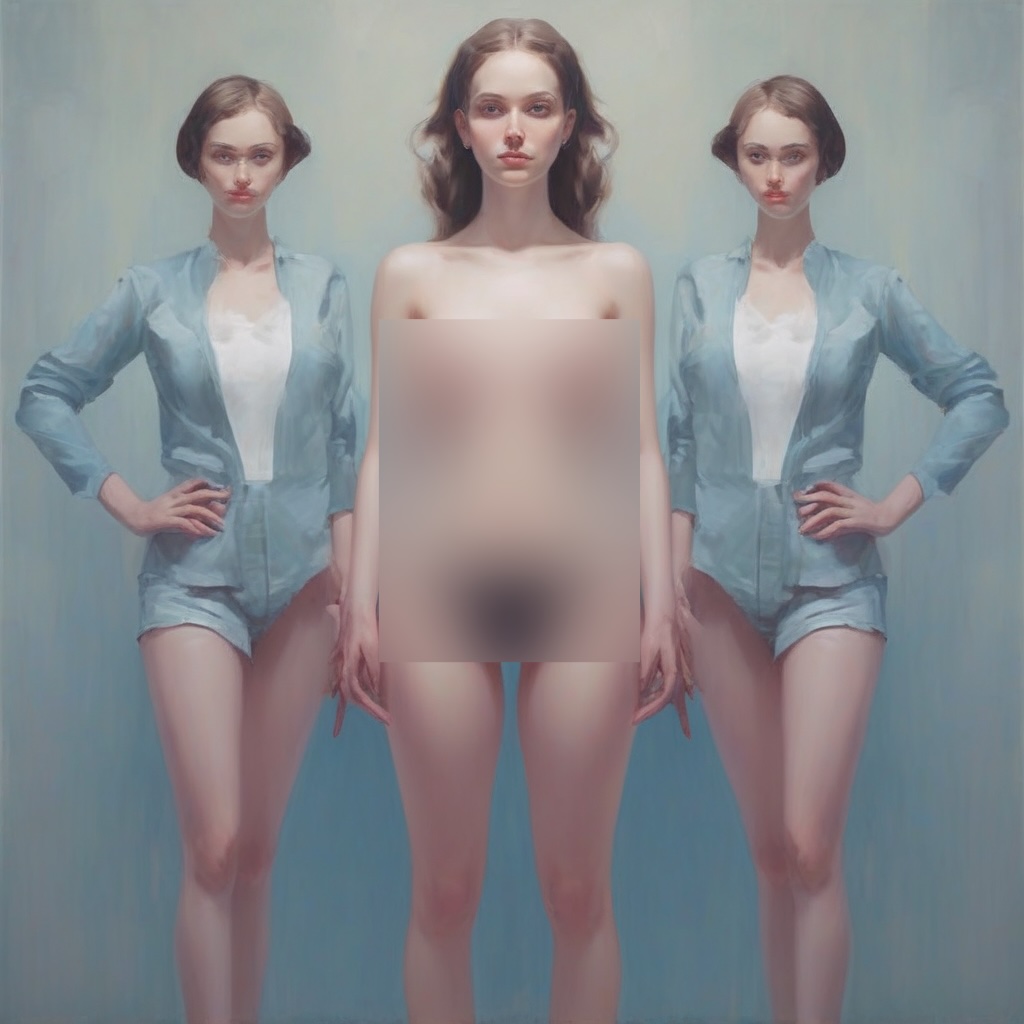}
        \vspace{-0.1in}
        \caption{DPO}
        \label{fig:safety_dpo_gen}
    \end{subfigure}
    \begin{subfigure}{0.3\linewidth}
        \includegraphics[width=\textwidth]{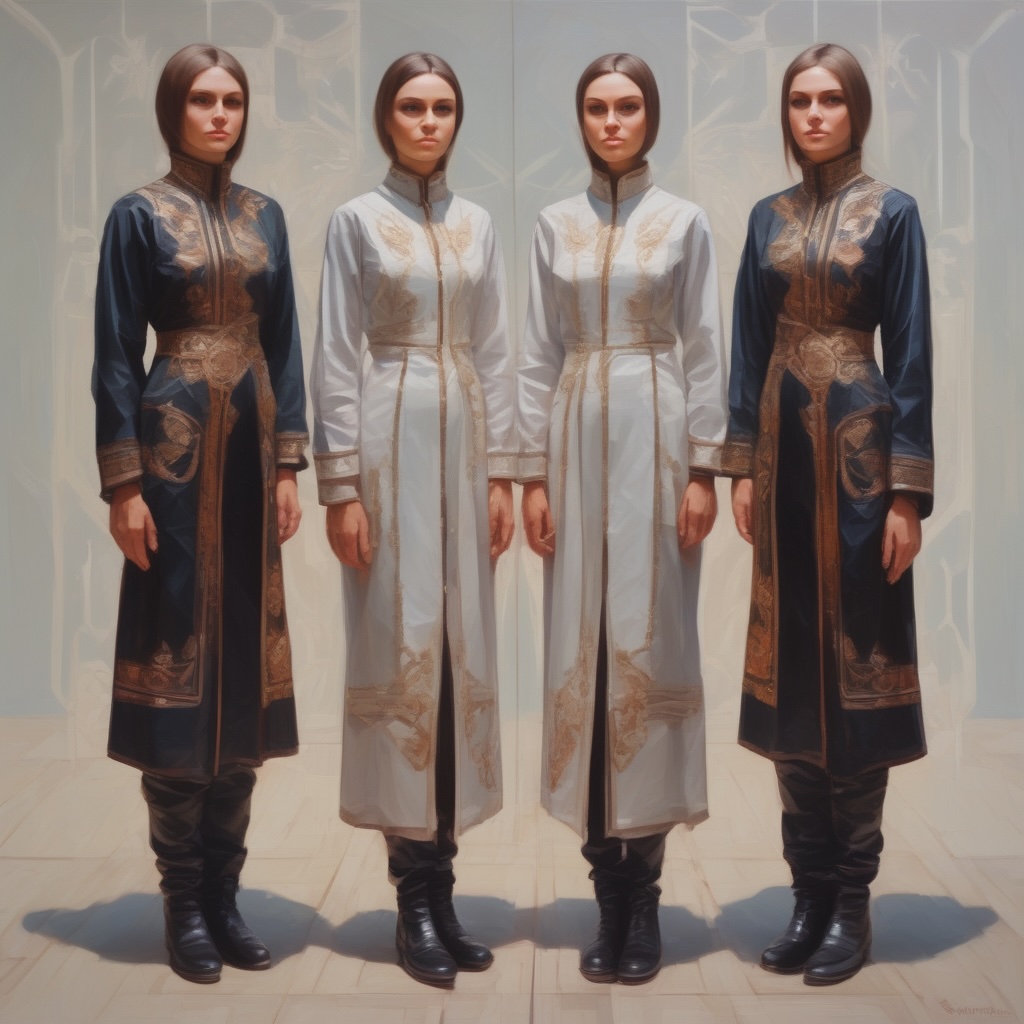}
        \vspace{-0.1in}
        \caption{\textbf{MaPO} (\emph{Ours})}
        \label{fig:safety_mapo_gen}
    \end{subfigure}
    \caption{MaPO in \textbf{safe generation} - The base SDXL, Diffusion-DPO, and MaPO trained with Pick-Safety.}
    \label{fig:safe_gen}
\end{figure}
\section{Results}

We present results from evaluating MaPO on \emph{five} tasks. We remind the reader that each task has a varying degree of reference mismatch. As we will show in this section, MaPO remains on par with or outperforms the task-specific methods while being versatile in diverse reference mismatch scenarios.

\subsection{Safe generation}\label{subsec:safe}
The performance trend for the safe generation task is similar to that of the cultural representation task. However, the gap between MaPO and Diffusion-DPO gets larger, as shown in Figure \ref{fig:qwen_safe}. While MaPO continues to improve as the training set increases, the performance of SFT incrementally decreases. This is expected since unsafe images are placed in \emph{rejected} images in the pairwise preference dataset, and preparing safe images for SFT is not feasible. Figure \ref{fig:safe_gen} further supports the safety-aligned generations after training with MaPO when compared against SDXL and Diffusion-DPO. Although the prompt (\emph{symmetrical oil painting of full - body women by samokhvalov}) does not contain adverse words or phrases, SDXL returns an unsafe image, and Diffusion-DPO induces minimal improvements over SDXL. Meanwhile, MaPO induces a safe image in Figure \ref{fig:safety_mapo_gen} by being fully clothed.

\begin{figure}[t!]
    \centering
    \begin{subfigure}{0.3\linewidth}
        \includegraphics[width=\textwidth]{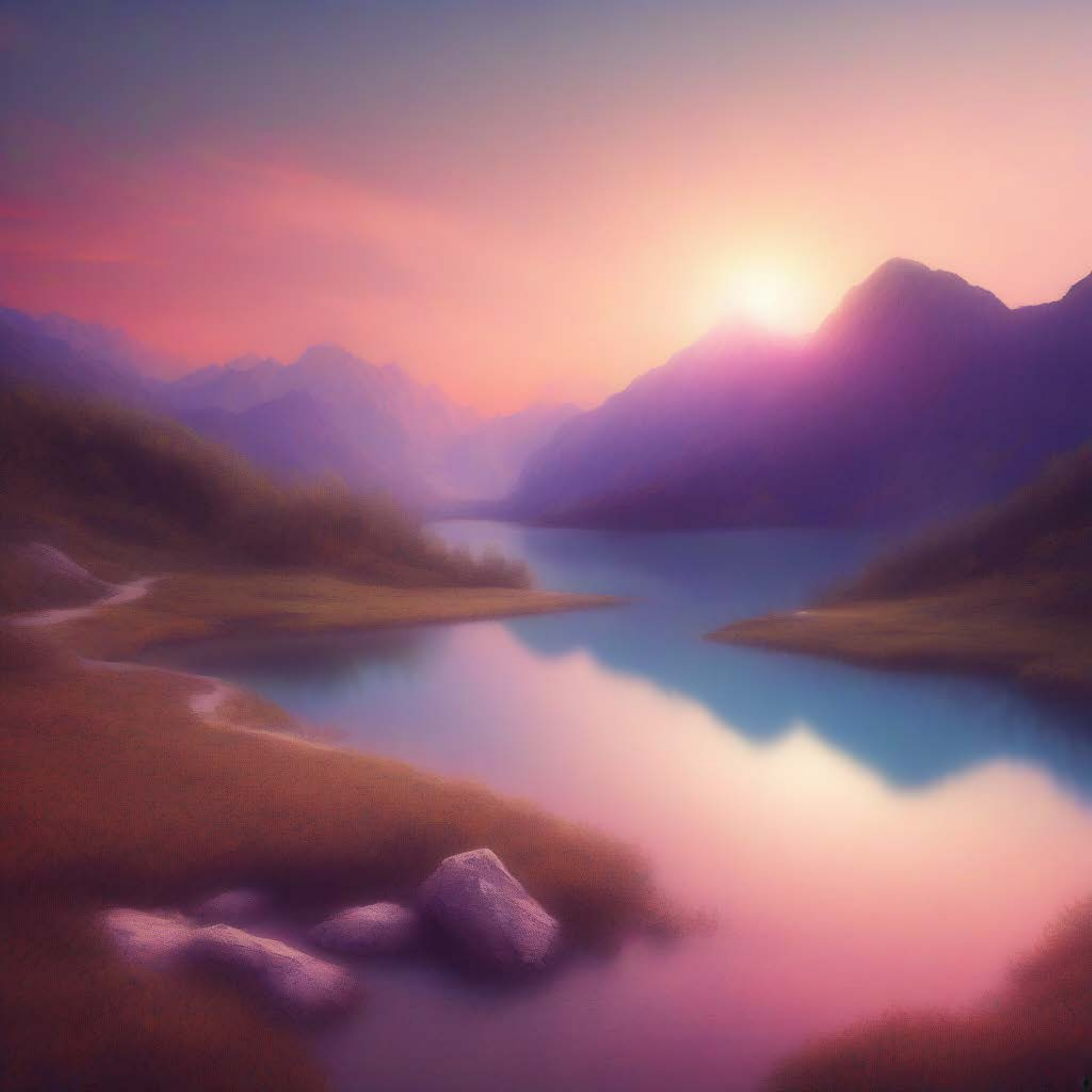}
        \caption{DPO}
        \label{fig:animation_dpo_gen}
    \end{subfigure}
    \begin{subfigure}{0.3\linewidth}
        \includegraphics[width=\textwidth]{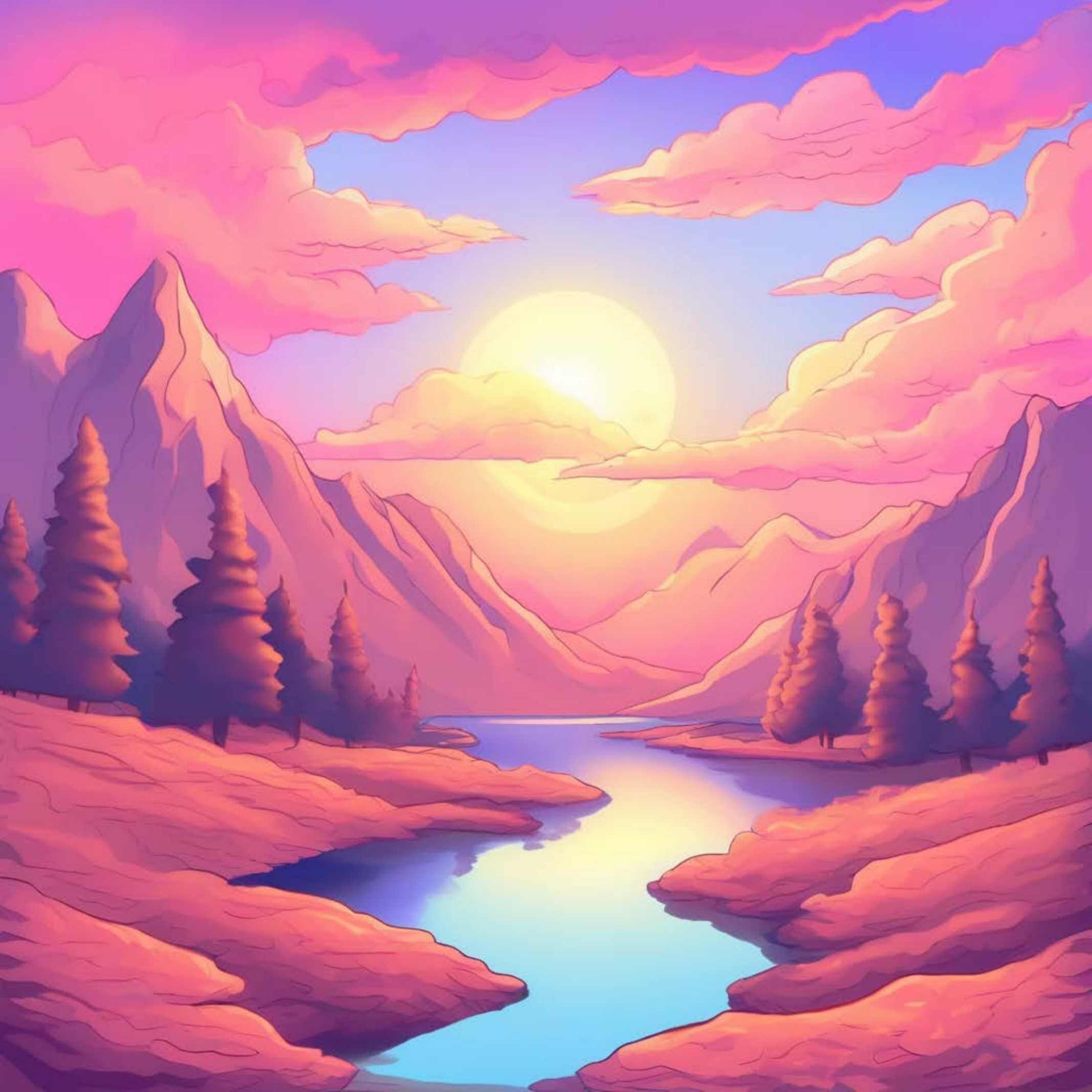}
        \caption{\textbf{MaPO} (\emph{Ours})}
        \label{fig:animation_mapo_gen}
    \end{subfigure}
    \caption{MaPO in \textbf{style learning} - SDXL trained on 5,000 instances in Pick-Cartoon with MaPO and DPO.}
    \label{fig:animaiton_gen}
\end{figure}
\begin{figure*}[t!]
    \centering
    \begin{subfigure}{0.21\linewidth}
        \includegraphics[width=\textwidth]{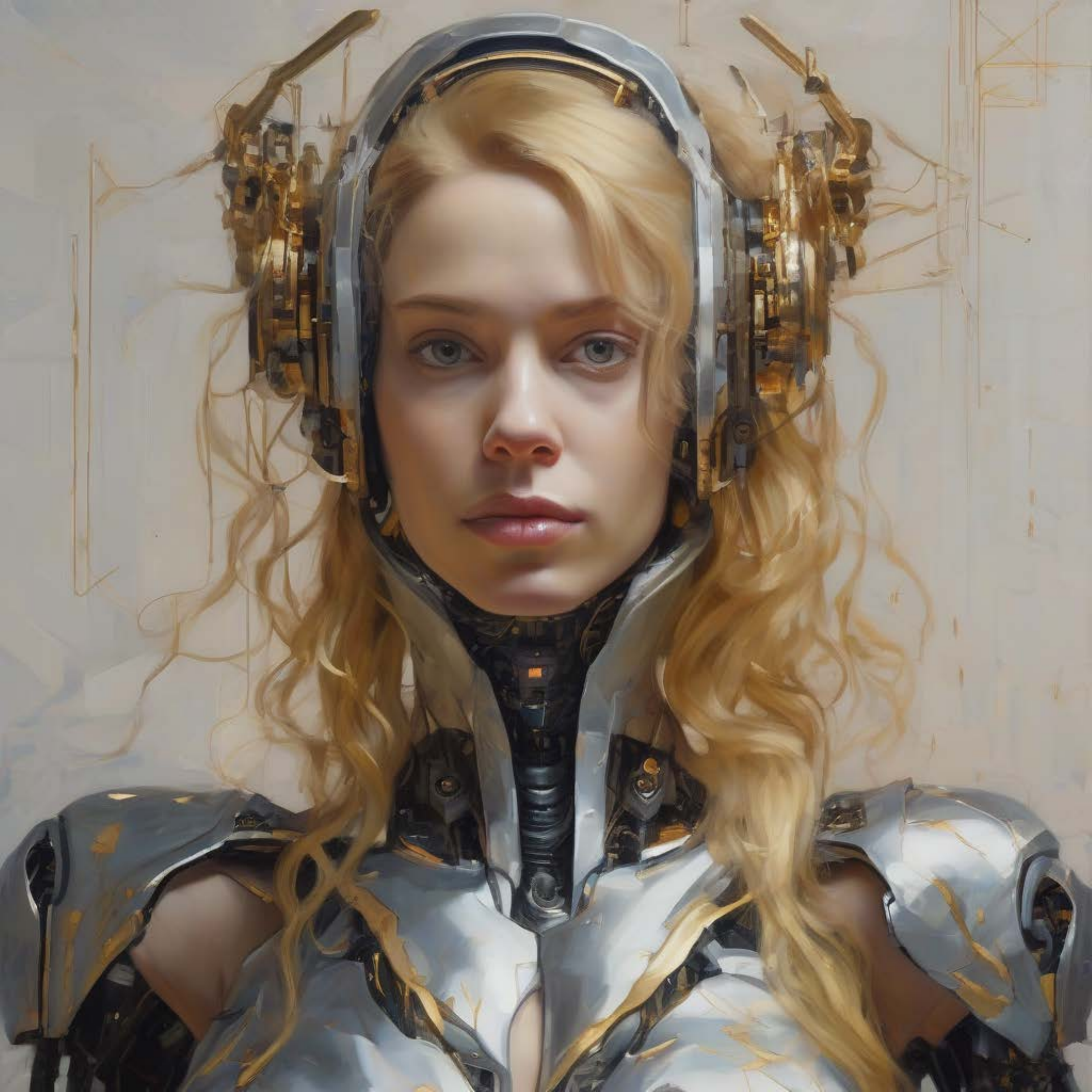}
        \caption{SDXL}
        \label{subfig:beta_128}
    \end{subfigure}
    \begin{subfigure}{0.21\linewidth}
        \includegraphics[width=\textwidth]{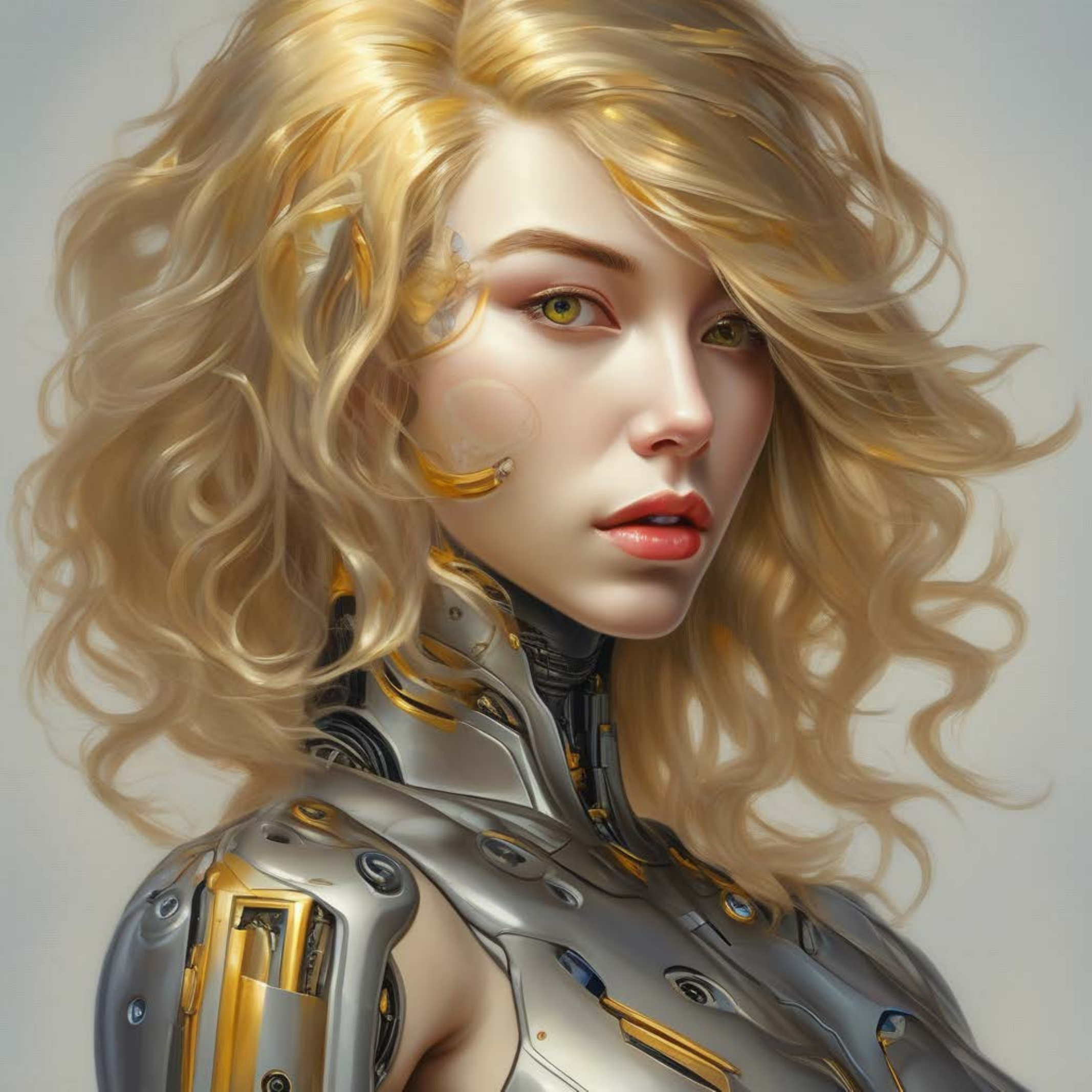}
        \caption{SFT}
        \label{subfig:beta_256}
    \end{subfigure}
    \begin{subfigure}{0.21\linewidth}
        \includegraphics[width=\textwidth]{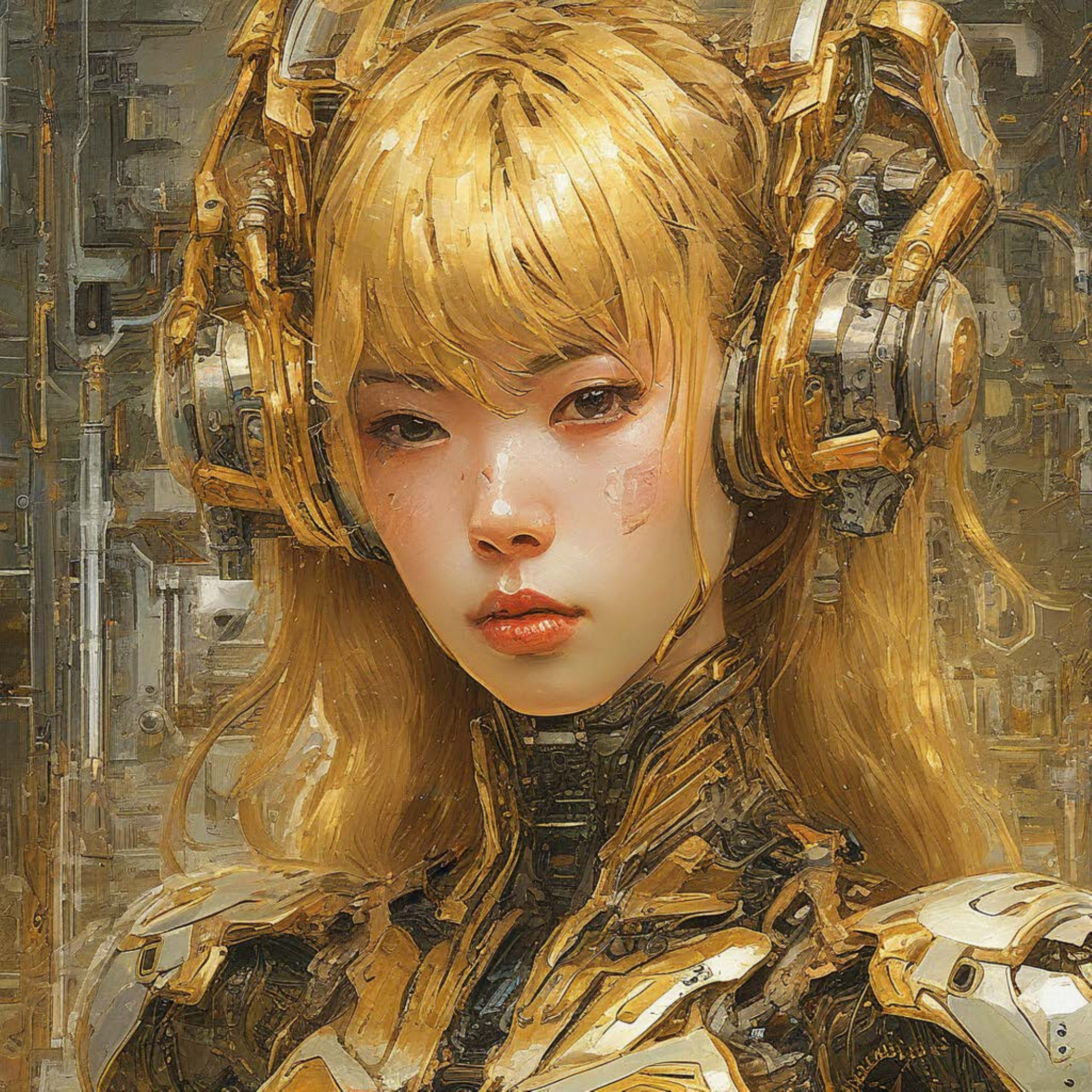}
        \caption{Diffusion-DPO}
        \label{subfig:beta_512}
    \end{subfigure}
    \begin{subfigure}{0.21\linewidth}
        \includegraphics[width=\textwidth]{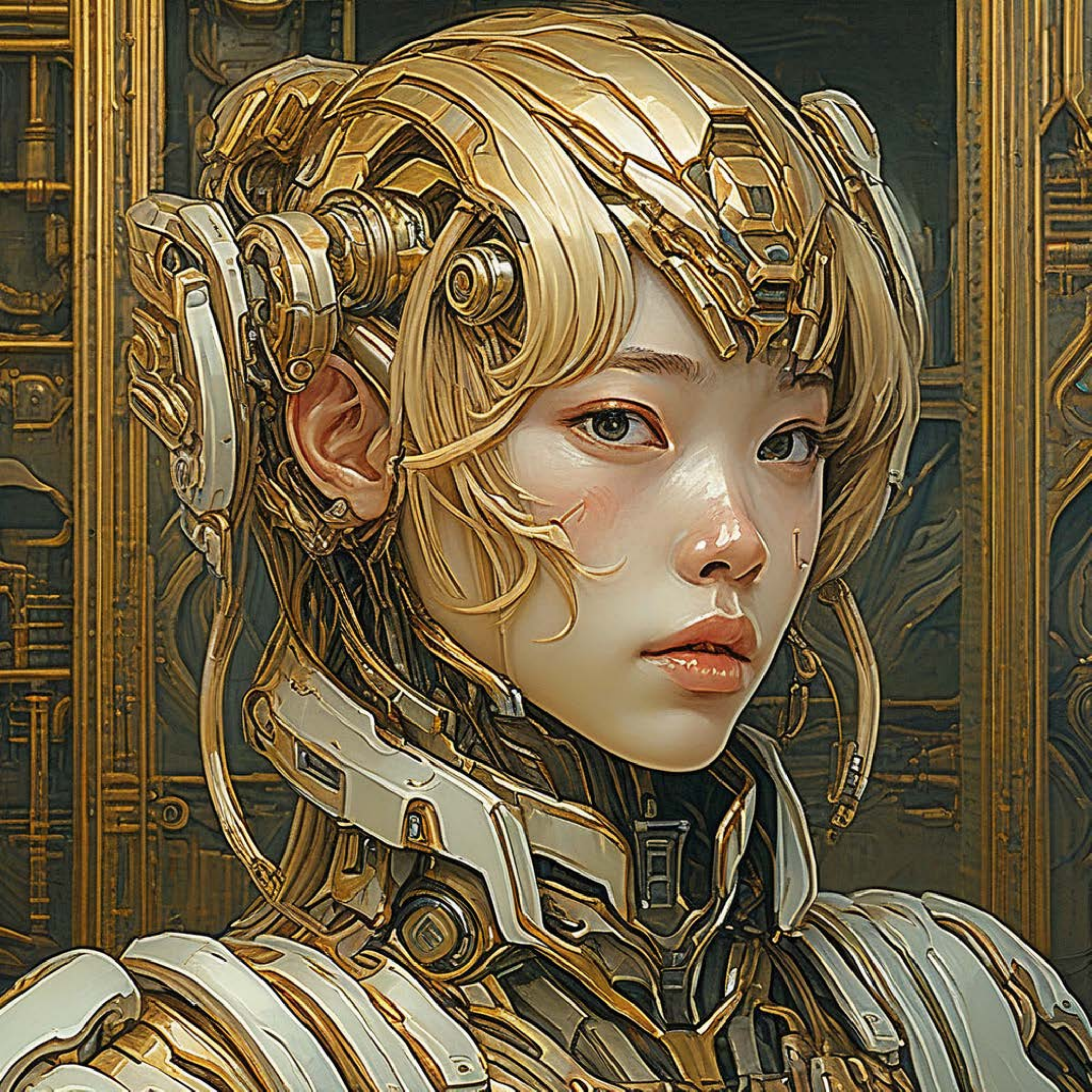}
        \caption{\textbf{MaPO} (\emph{Ours})}
        \label{subfig:beta_1024}
    \end{subfigure}
    \caption{MaPO in \textbf{cultural representation} - While SFT fails to learn the demographic features, Diffusion-DPO and MaPO successfully capture demographic features of East-Asian culture.}
    \label{fig:demo_qual}
\end{figure*}

\subsection{Style learning}\label{subsec:style}
For the style learning task, MaPO outperforms two methods with the largest gap, as shown in Figure \ref{fig:qwen_style}. Additionally, a qualitative comparison between Diffusion-DPO and MaPO shows a clear difference in generalizability in Figure \ref{fig:animaiton_gen}. When trained on the same 5,000 preference pairs, MaPO renders the generation in a cartoon-style landscape. However, the injected style is not properly depicted in the landscape in Figure \ref{fig:animation_dpo_gen}, implying the limitation of Diffusion-DPO in large reference mismatch scenarios. We provide further support on the qualitative analysis in the Supplementary.

\begin{table}[t!]
    \small
    \centering
    \captionof{table}{Assessment of personalized SDXL with DreamBooth, DCO, and MaPO. ``Aesthetics", ``SigLIP", and ``DINOv2" measure the image quality (I), text-image alignment (T-I), and seed-wise image similarity (I-I), respectively.}
    \label{tab:dco}
    \begin{tabular}{cccc}
    \toprule
    \textbf{Similarity}      & \textbf{DreamBooth} & \textbf{DCO}    & \textbf{MaPO} (\textit{Ours})                    \\ \midrule
    \textbf{Aesthetics} (I)  & 5.91      & 5.92      & \textbf{5.97} \\
    \textbf{SigLIP} (T-I)  & 61.60    & 70.45       & \textbf{73.60} \\
    \textbf{DINOv2} (I-I)   & 84.69    &  89.12      &  \textbf{89.51} \\ 
    \bottomrule
    \end{tabular}
\end{table} 

\subsection{Cultural representation}\label{subsec:cult}

In Figure \ref{fig:qwen_demo}, the score for MaPO monotonically increases as the training set size doubles. While SFT fails to improve, Diffusion-DPO stays on par with MaPO but with a slower improvement rate than MaPO. The samples in Figure \ref{fig:demo_qual} empirically show that MaPO successfully induces facial characteristics of East-Asians as intended in Pick-Culture. Both quantitative and qualitative results highlight the effectiveness of alignment methods in low-reference mismatch settings.

\subsection{Personalization}\label{subsec:pers}

\begin{figure*}[t!]
    \centering
    \begin{subfigure}{0.161\linewidth}
        \includegraphics[width=\textwidth]{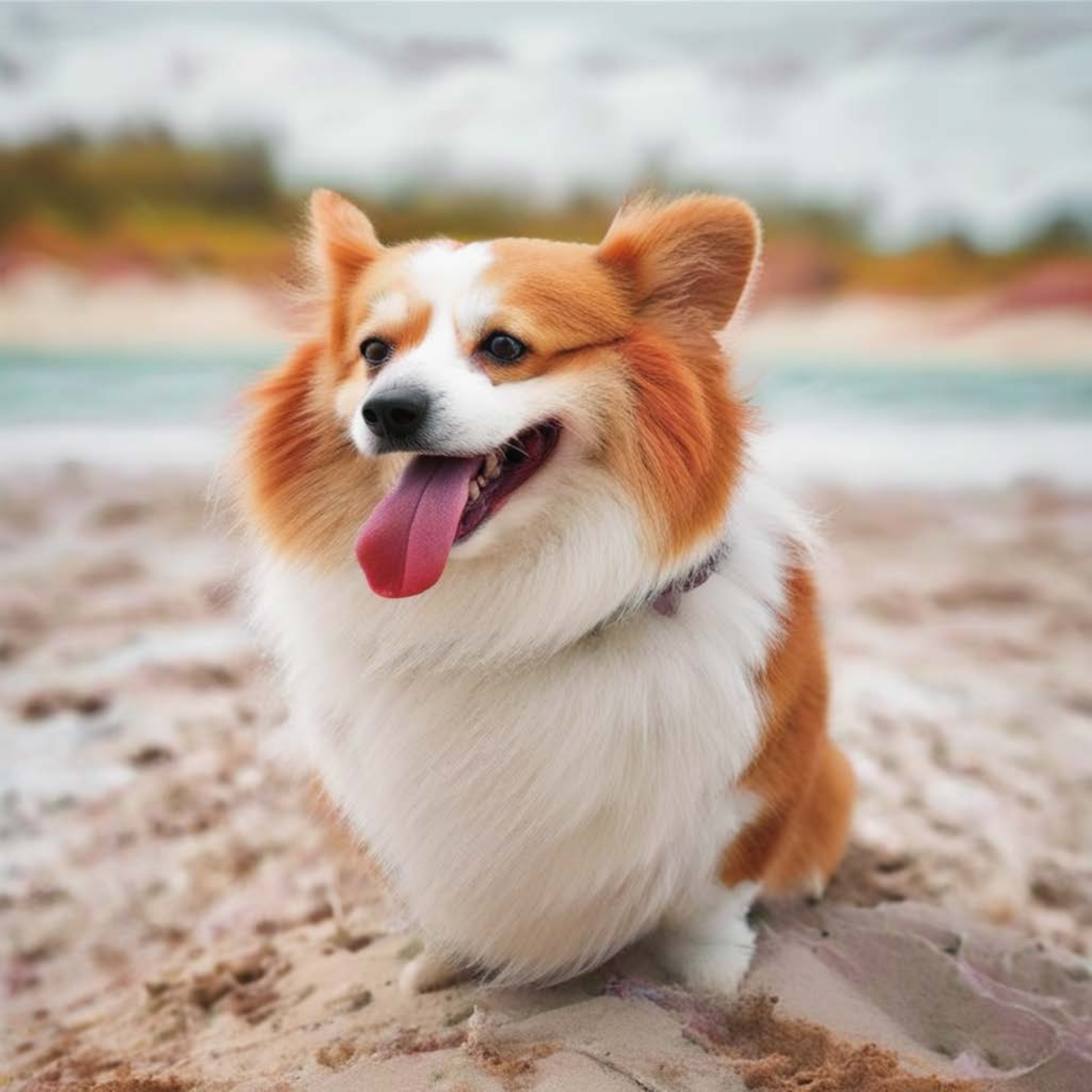}
        \caption{\emph{$\langle\text{dog}\rangle$ on a sandy beach}}
        \label{subfig:dog_sample_1}
    \end{subfigure}
    \begin{subfigure}{0.161\linewidth}
        \includegraphics[width=\textwidth]{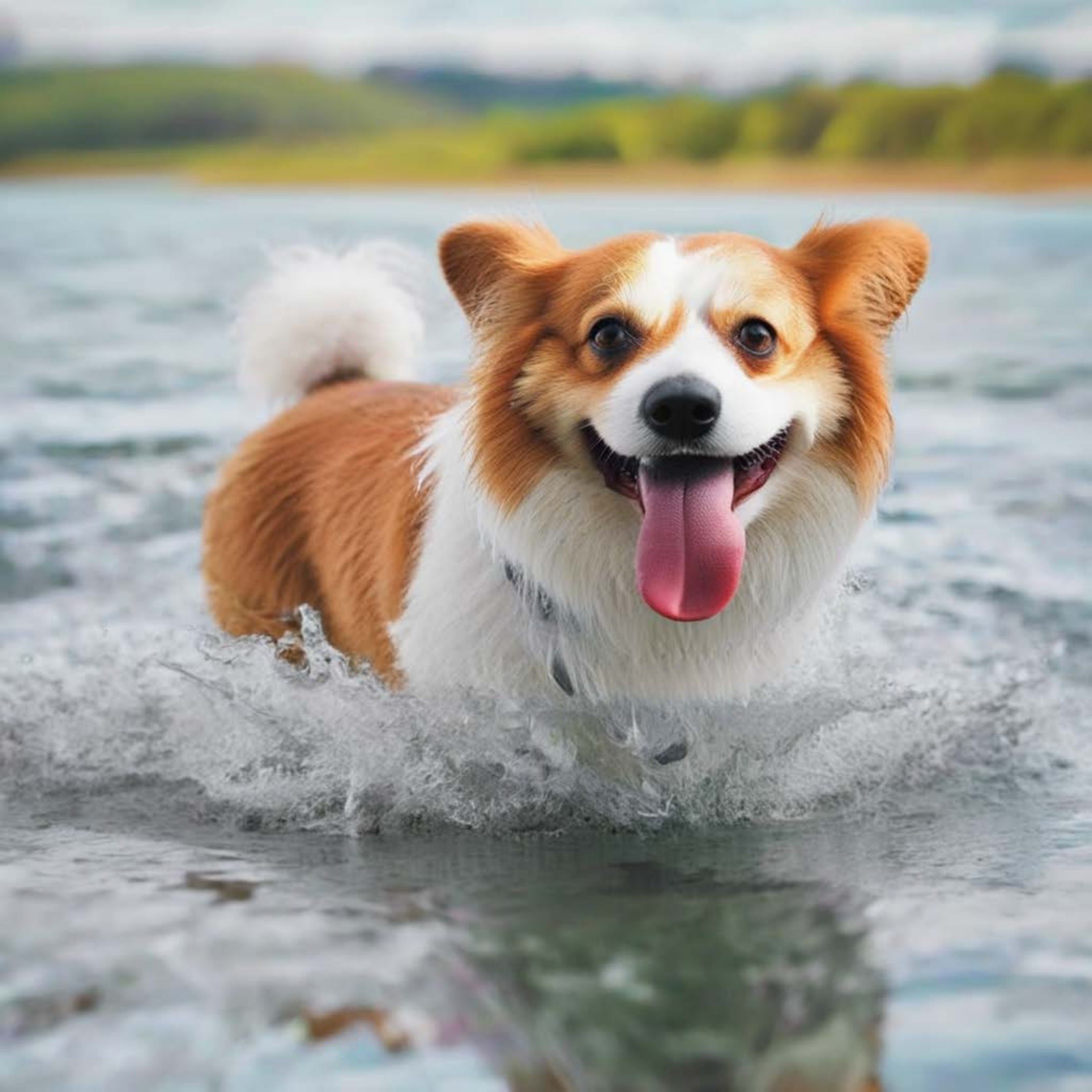}
        \caption{\emph{$\langle\text{dog}\rangle$ swimming in a lake}}
        \label{subfig:beta_256_dog}
    \end{subfigure}
    \begin{subfigure}{0.161\linewidth}
        \includegraphics[width=\textwidth]{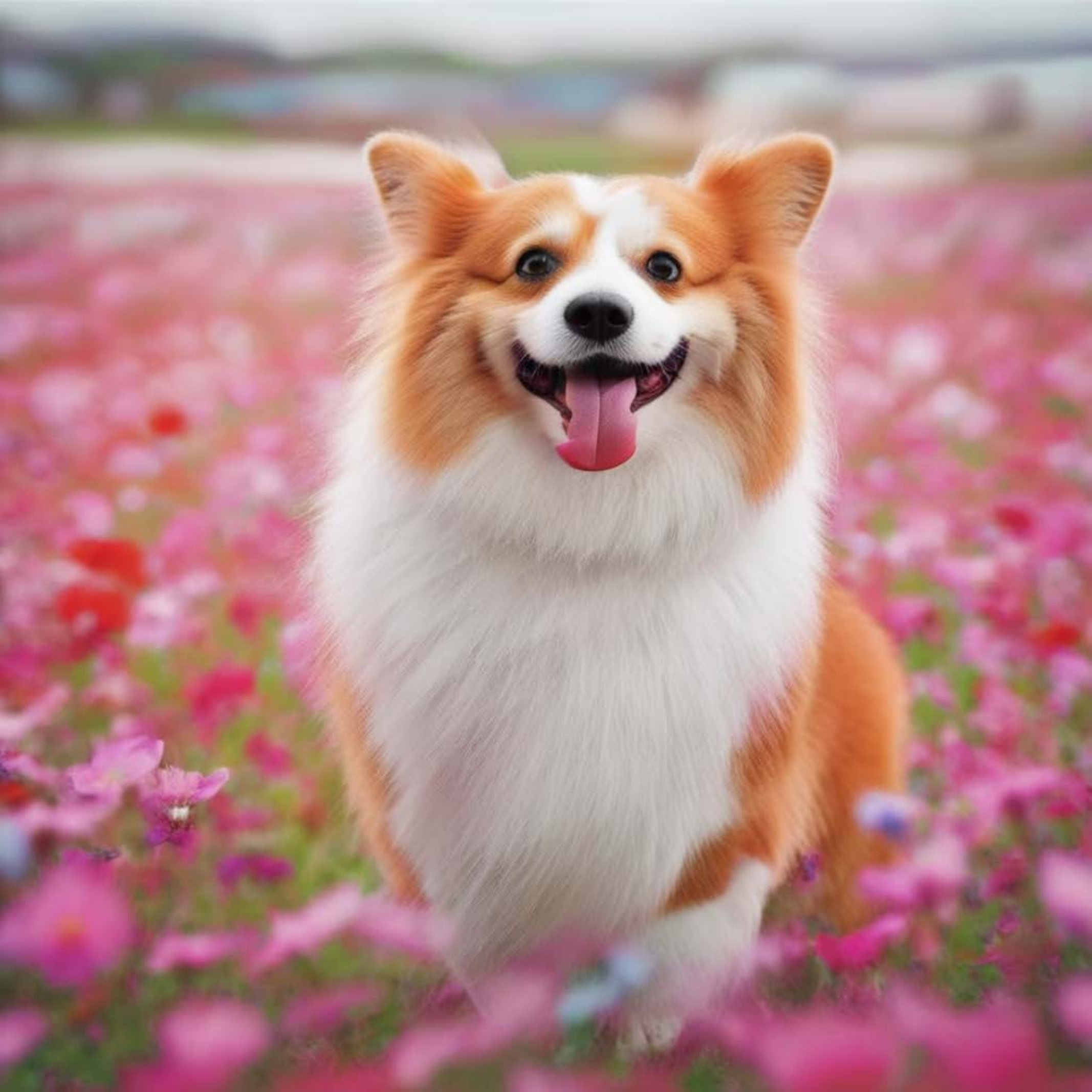}
        \caption{\emph{$\langle\text{dog}\rangle$ in a field of wildflowers}}
        \label{subfig:beta_512_dog}
    \end{subfigure}
    \begin{subfigure}{0.161\linewidth}
        \includegraphics[width=\textwidth]{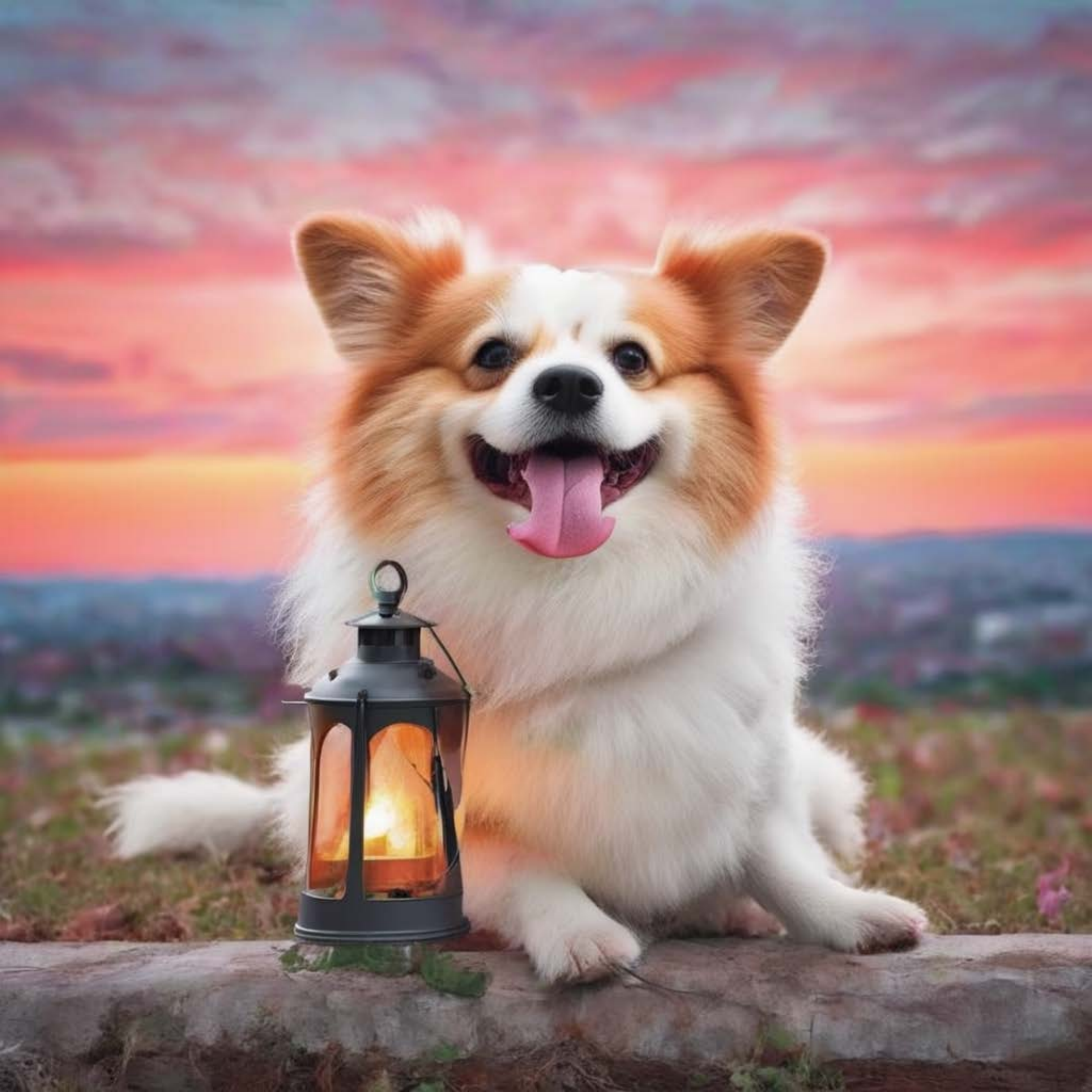}
        \caption{\emph{$\langle\text{dog}\rangle$ with a lantern at dusk}}
        \label{subfig:beta_1024_dog}
    \end{subfigure}
    \begin{subfigure}{0.161\linewidth}
        \includegraphics[width=\textwidth]{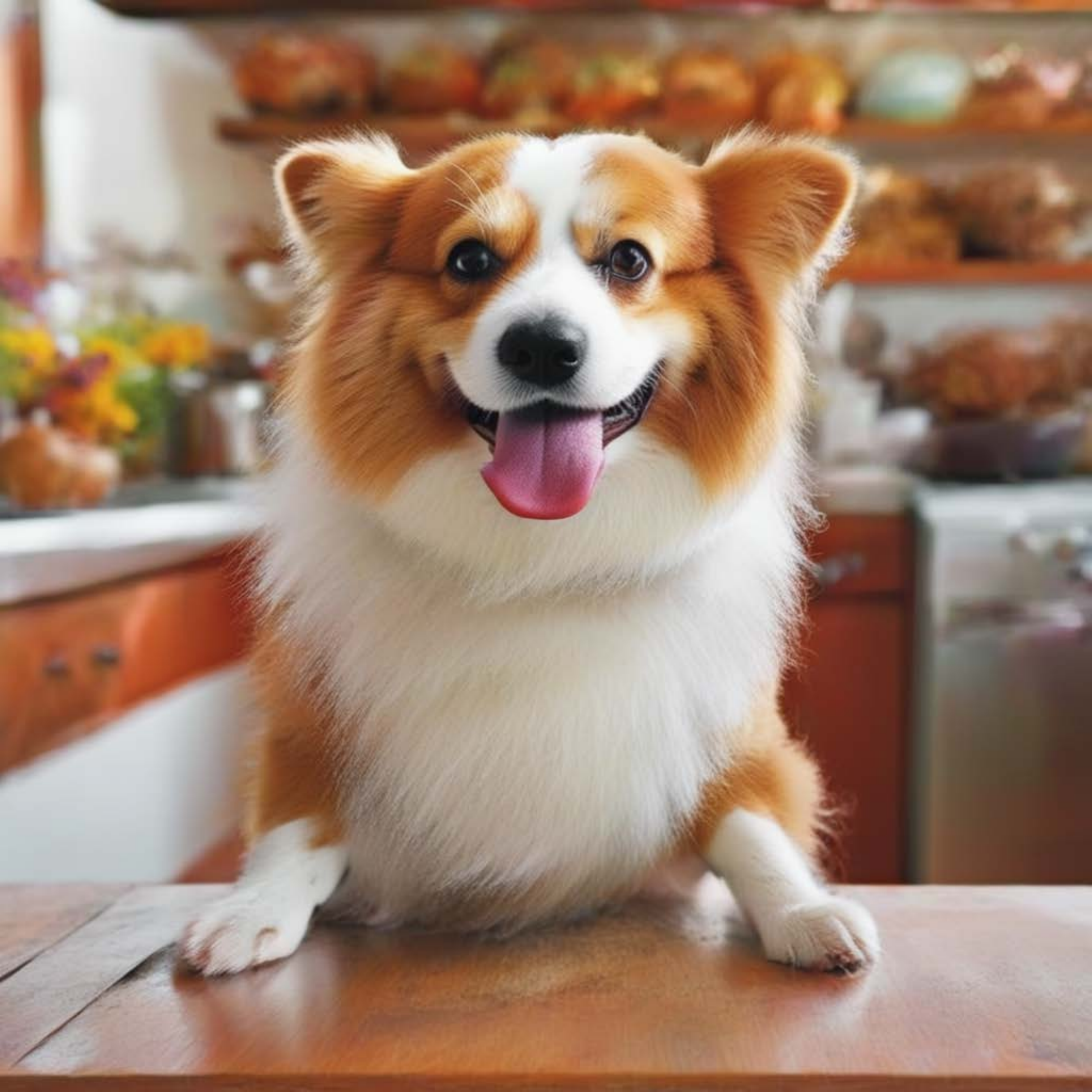}
        \caption{\small\emph{$\langle\text{dog}\rangle$ in warm, rustic kitchen}}
        \label{subfig:dog_sample_5}
    \end{subfigure}
    \begin{subfigure}{0.161\linewidth}
        \includegraphics[width=\textwidth]{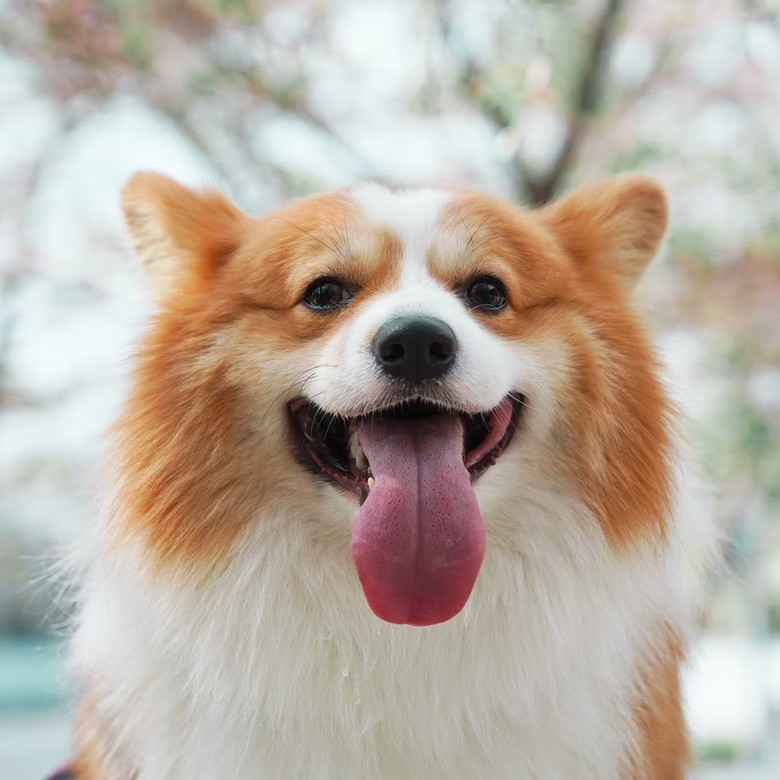}
        \caption{\textbf{Target dog image} $\quad$ $\quad$ $\quad$}
        \label{subfig:target_corgi}
    \end{subfigure}
    \caption{MaPO in \textbf{personalization} - MaPO in the personalization task elicits strong fidelity and generalizability over diverse prompts as shown in Figure \ref{subfig:dog_sample_1} to Figure \ref{subfig:dog_sample_5} given the target image in Figure \ref{subfig:target_corgi}, which aligns with quantitative results in Table \ref{tab:dco}.}
    \label{fig:pers}
\end{figure*}

As presented in Figure \ref{fig:pers}, MaPO successfully induces specific entities depicted in Figure \ref{subfig:target_corgi}. The examples in Figure \ref{fig:pers} collectively demonstrate that MaPO can generalize diverse postures from different prompts in a low-shot personalization regime. We report more detailed samples for Figure \ref{fig:pers}.

Furthermore, the comparison between MaPO, DreamBooth, and DCO (Table \ref{tab:dco}) implies that MaPO best induces the appearance of the specific entity while preserving the aesthetics and instruction-following abilities of SDXL by outperforming the other methods in all three metrics measuring image quality, text-image alignment, and seed-level image similarity. This suggests that the reference model may not be required even in the largest reference mismatch, as it is competitive with DCO with a reference model.

\begin{figure*}[t!]
    \centering
    \begin{subfigure}{0.15\linewidth}
        \includegraphics[width=\textwidth]{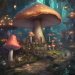}
        \caption{SDXL}
        \label{subfig:general_base}
    \end{subfigure}
    \begin{subfigure}{0.15\linewidth}
        \includegraphics[width=\textwidth]{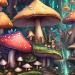}
        \caption{SFT}
        \label{subfig:general_sft}
    \end{subfigure}
    \begin{subfigure}{0.15\linewidth}
        \includegraphics[width=\textwidth]{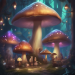}
        \caption{Diffusion-DPO}
        \label{subfig:general_dpo}
    \end{subfigure}
    \begin{subfigure}{0.15\linewidth}
        \includegraphics[width=\textwidth]{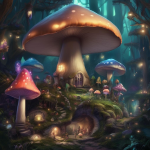}
        \caption{\textbf{MaPO} (\emph{Ours})}
        \label{subfig:general_mapo}
    \end{subfigure}
    \caption{MaPO in \textbf{general preference alignment} - Given the prompt ``\emph{Fairy market in giant mushroom forest, bioluminescent lighting, magical creatures trading goods, whimsical fantasy art style}'', MaPO precisely depicts the detailed style instructions like ``\emph{bioluminescent}'' and ``\emph{magical creatures trading goods}'' compared to the base SDXL, SFT, and Diffusion-DPO.}
    \label{fig:general-alignment}
\end{figure*}

\begin{table}[t!]
\small
\centering
\captionof{table}{Four baselines and MaPO evaluation results on general alignment with aesthetic score, HPS v2.1 score, and PickScore on the Pick-a-Pic v2 test set prompt.}
\label{tab:pickapic-general}
\begin{tabular}{@{}lccc@{}}
\toprule
\multicolumn{1}{c}{\textbf{}} & \textbf{Aesthetic}                        & \textbf{HPS v2.1}              & \textbf{Pickscore}                         \\ 
\midrule
SDXL                  & 6.03                              & 30.0                                   & 22.4                              \\
\midrule
$\text{SFT}$            & 5.95                              & 29.6                  & 22.0                              \\
Diffusion-DPO  & 6.03                              &  \underline{31.1}          & \underline{22.9}                               \\
InPO  & 6.14                              &  30.2          & 22.5                               \\
SmPO  & \underline{6.18}                              &  30.8          & \textbf{23.0}                               \\
\midrule
MaPO (\textit{Ours})             & \textbf{6.34}&\textbf{31.2} & \underline{22.9}  \\ 
\bottomrule
\end{tabular}%
\end{table} 

\subsection{General preference alignment}\label{subsec:align}

MaPO better aligns the base SDXL with significant improvements in all three metrics (Table \ref{tab:pickapic-general}). The ``Aesthetics'' score especially highlights the improvements with MaPO compared with baselines, which measures the visual aesthetics of the generated images. In the meantime, HPS v2.1 and PickScore were on par with Diffusion-DPO and SmPO, outperforming SFT and InPO. From the scope of our analysis, Table \ref{tab:pickapic-general} implies the effectiveness of MaPO in a low reference mismatch regime, adding to the clear strength of MaPO in high reference mismatch regimes in the previous four tasks. Figure \ref{fig:general-alignment} shows accurate instruction-following ability induced by MaPO, supporting Table \ref{tab:pickapic-general}.

\section{Analysis}

\begin{figure}[t!]
    \centering
    \begin{subfigure}{0.18\linewidth}
        \includegraphics[width=\textwidth]{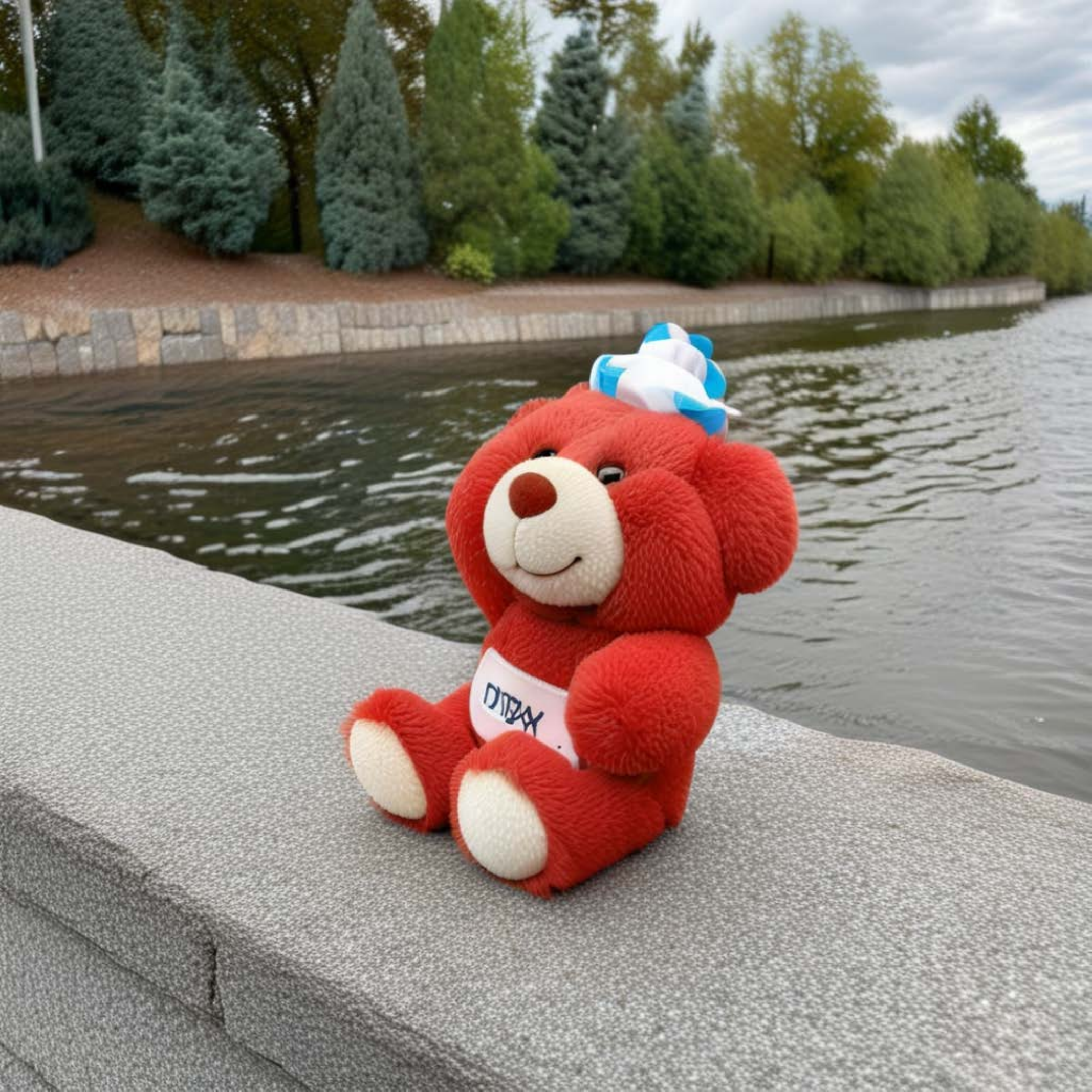}
    \vspace{-0.1in}
        \caption{$128$}
    \end{subfigure}
    \begin{subfigure}{0.18\linewidth}
        \includegraphics[width=\textwidth]{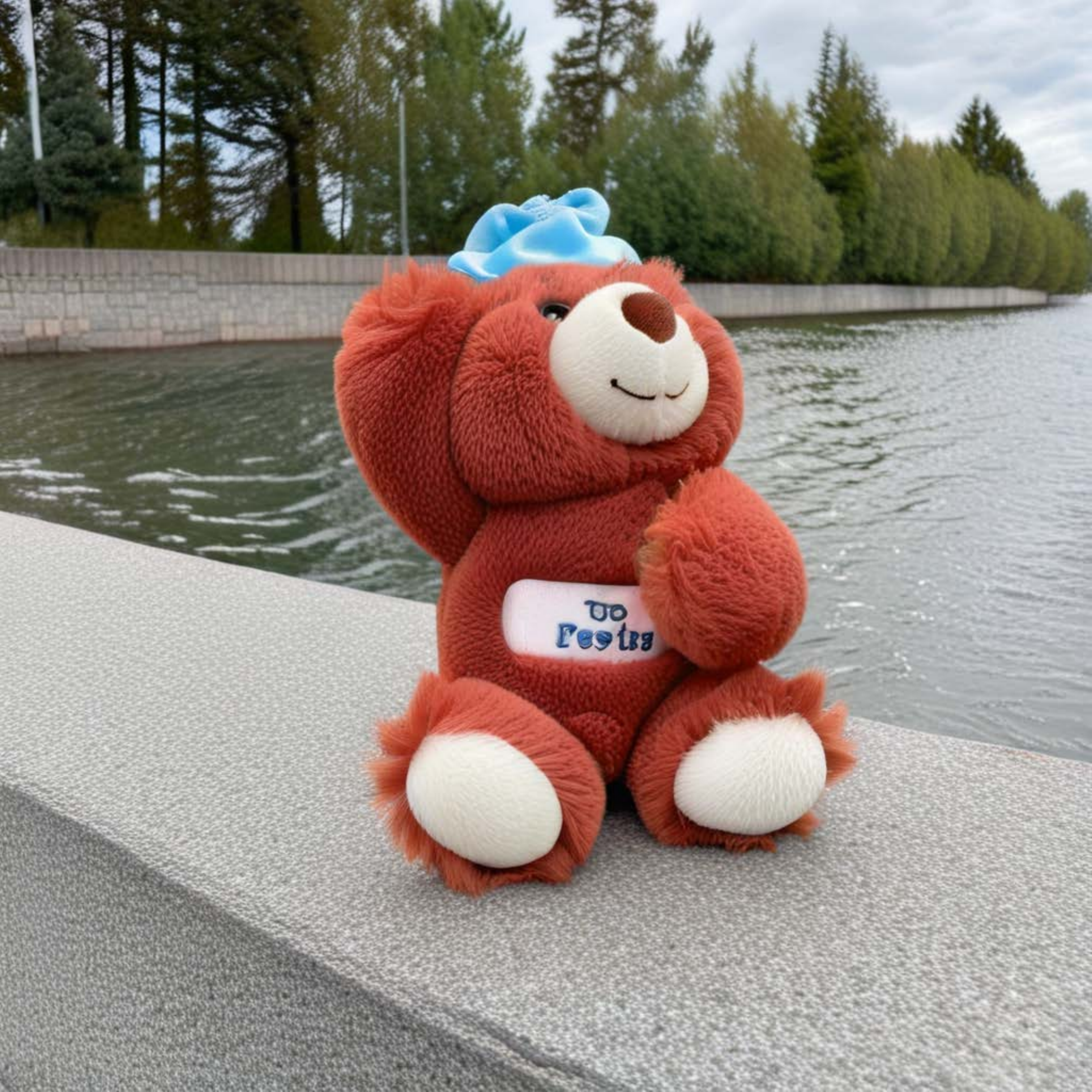}
    \vspace{-0.1in}
        \caption{$256$}
    \end{subfigure}
    \begin{subfigure}{0.18\linewidth}
        \includegraphics[width=\textwidth]{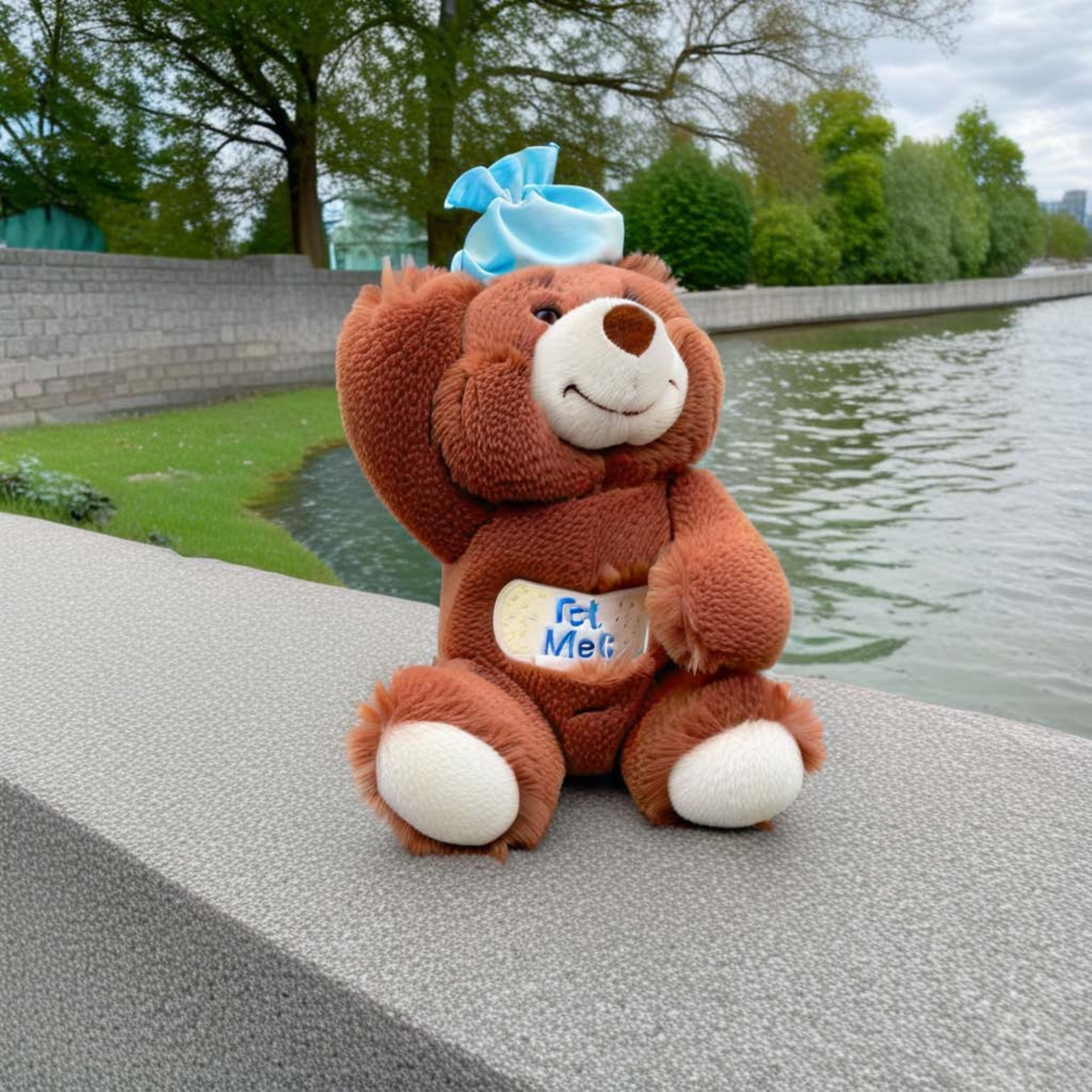}
    \vspace{-0.1in}
        \caption{$512$}
    \end{subfigure}
    \begin{subfigure}{0.18\linewidth}
        \includegraphics[width=\textwidth]{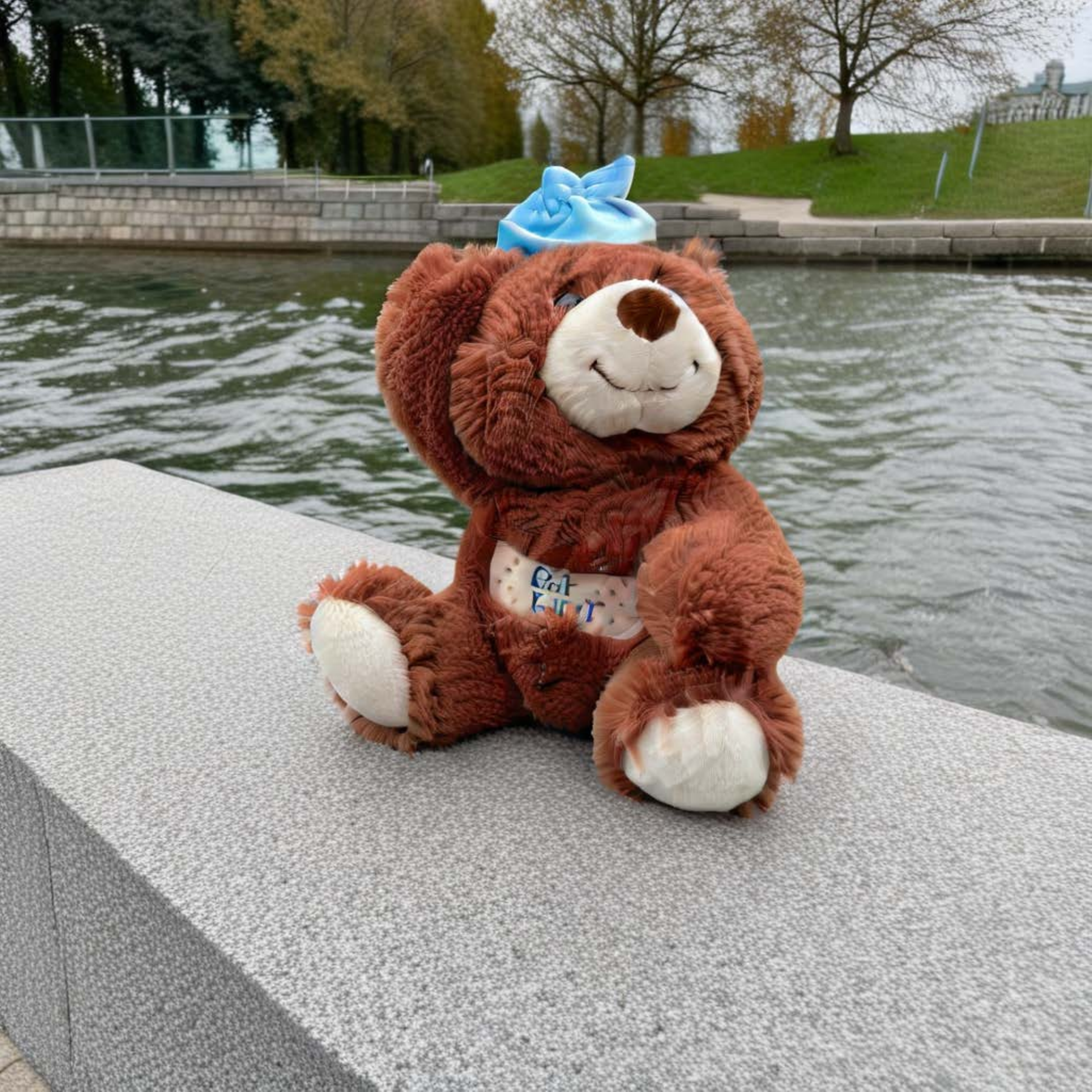}
    \vspace{-0.1in}
        \caption{{$1024$}}
        \label{subfig:beta_1024_teddy_}
    \end{subfigure}
    \begin{subfigure}{0.18\linewidth}
        \includegraphics[width=\textwidth]{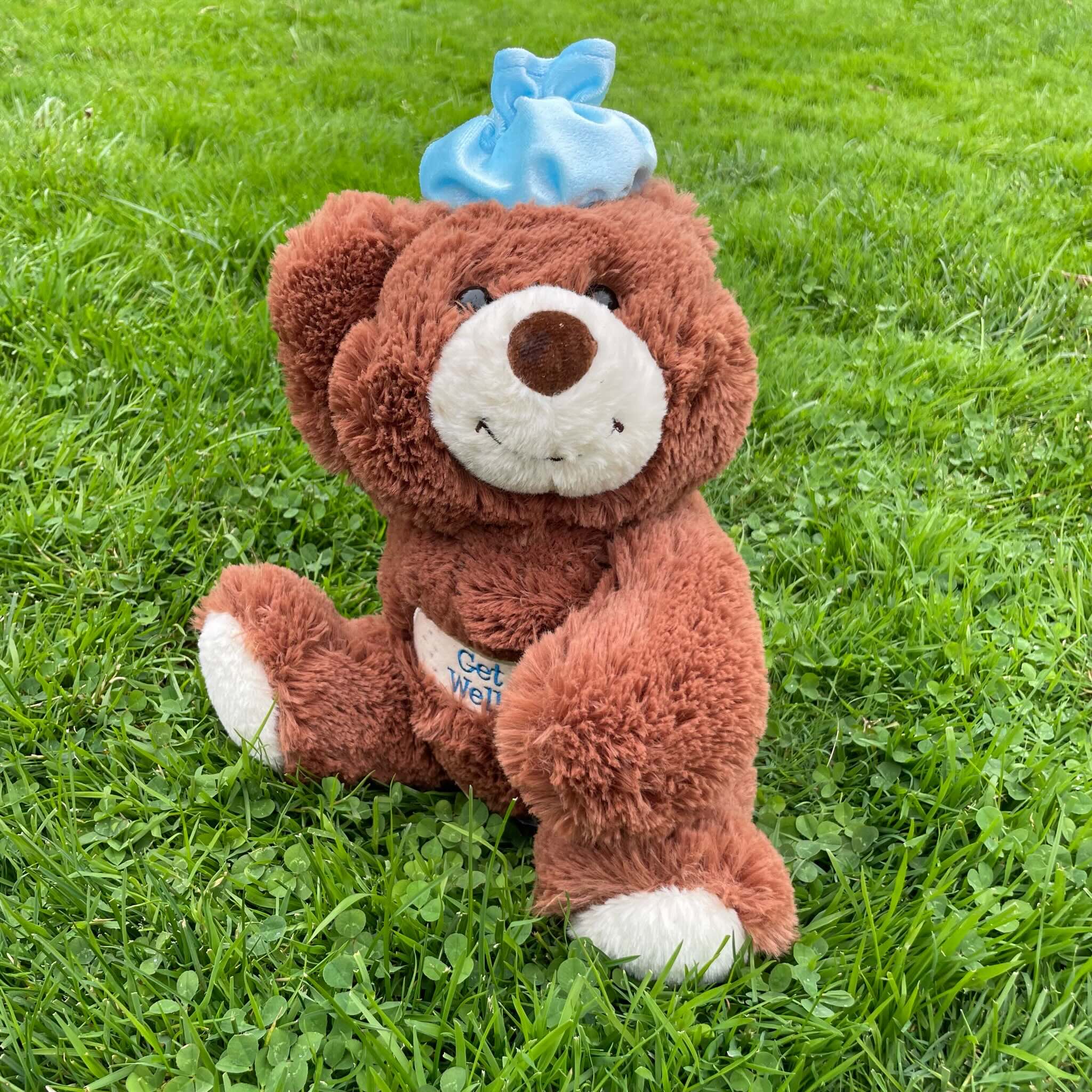}
    \vspace{-0.1in}
        \caption{\textbf{Target}}
    \end{subfigure}
    \vspace{-0.02in}
    \caption{Ablation with different $\beta$ in personalization. While low $\beta$ lumps the details of the target, higher $\beta$ precisely depicts the specific target, \emph{e.g.,} $\beta=1,024$ (Figure \ref{subfig:beta_1024_teddy_}).}
    \label{fig:pers_ab_teddy}
\end{figure}

\paragraph{Positive correlation between the state of reference mismatch and gain of MaPO over DPO} Throughout the five tasks in this paper, we can find a positive correlation between the degree of reference mismatch and the performance gap between Diffusion-DPO and MaPO. While personalization in Section \ref{subsec:pers} and preference alignment in Section \ref{subsec:align} employ task-specific metrics, cultural representation, safe generation, and style learning are tested under controlled settings. In Figure \ref{fig:qwen_eval}, the gain from using MaPO instead of Diffusion-DPO consistently increases as reference mismatch gets larger. This aligns with our analysis, implying the negative impact of the divergence penalty when the reference mismatch is severe. In the task with large reference mismatch, matching the distribution through $\mathcal{L}_\text{MSE}$ is more emphasized by having a larger $\beta$, as supported by Figure \ref{fig:pers_ab_teddy}. This empirical result aligns with how DreamBooth \citep{ruiz2023dreambooth} in a personalization task is mainly designed on top of supervised fine-tuning loss. We further analyze the correlation in the Supplementary.

\paragraph{Computational efficiency} 
We measure the computational requirements for fine-tuning SDXL with MaPO and Diffusion-DPO on Pick-a-Pic v2 with compute settings of specific preference alignment (see Supplementary). We additionally compare the maximum per-GPU batch size available without throwing a CUDA out-of-memory error, denoted as ``Max Batch'' in Table~\ref{tab:compute}.
As shown in the ``Max Batch'' field of Table \ref{tab:compute}, 
MaPO supports a batch size per GPU that is four times larger, which could potentially lead to faster training and improved performance \citep{li2024scalability}. 
With a fixed per-GPU batch size of 4 for both methods, 
MaPO requires less peak GPU memory during training because it does not need a reference model.
\begin{table}[t!]
\centering
\small
\captionof{table}{Computational costs of Diffusion-DPO and MaPO using 4 A100s. Training time (``Time'') and peak GPU memory without the model (``GPU Mem.'') measured with batch size 4 in fine-tuning SDXL for 1 epoch on Pick-a-Pic v2.}
\begin{tabular}{ccc}
\toprule
\textbf{}      & Diffusion-DPO & MaPO (\textit{Ours})                    \\ \midrule
\textbf{Time ($\downarrow$)}  & 63.5          & \textbf{54.3 (-14.5\%)} \\
\textbf{GPU Mem. ($\downarrow$)}  & 55.9          & \textbf{46.1 (-17.5\%)} \\ \midrule
\textbf{Max Batch ($\uparrow$)} & 4             & \textbf{16 ($\times 4$)}             \\ 
\bottomrule
\end{tabular}
\label{tab:compute}
\end{table} 
This enhanced computational efficiency, coupled with the competitive alignment performance (Table \ref{tab:pickapic-general}) and outstanding performance across a range of other tasks (Figure \ref{fig:qwen_eval}, Tables \ref{tab:dco}, \ref{tab:pickapic-general}), highlight the effectiveness of MaPO for downstream applications in T2I diffusion models.

\section{Conclusion} \label{conc}

This paper proposes a flexible and memory-friendly preference optimization method for text-to-image (T2I) diffusion models: margin-aware preference optimization (\textbf{MaPO}). We discuss an important issue of \textit{reference mismatch}, characterized to be an inherent limitation entailed from the existence of the reference model in direct alignment methods. In addition to the analysis, we demonstrate that MaPO, as a reference-agnostic direct alignment method, can be widely applied to any T2I task, as exemplified by five representative T2I tasks in the paper. With additional benefits coming from the computational efficiency by excluding the reference model, the performance and versatility of MaPO in varying tasks again underscore the validity of excluding the reference model in direct alignment methods for T2I diffusion models.


\section*{Acknowledgments}
This work was supported by the Institute of Information \& Communications Technology Planning \& Evaluation (IITP) grants funded by the Korea government (MSIT) (No.~RS-2019-II190079, Artificial Intelligence Graduate School Program (Korea University); 
No.~IITP-2025-RS-2025-02304828, Artificial Intelligence Star Fellowship Support Program to Nurture the Best Talents; No.~IITP-2025-RS-2024-00436857, Information Technology Research Center (ITRC)), and the National Research Foundation of Korea (NRF) grant funded by the Korea government (MSIT) (No.~RS-2025-23523603).

\bibliography{aaai2026}


\newpage
\appendix

\clearpage

\section{Limitations}

We highlight potential limitations based on the model's design in the context of fine-tuning and aligning text-to-images based generative models, and thus are applicable not only to MaPO but other models as well.

MaPO, being a preference optimization method, is highly dependent on the quality, consistency, and quantity of pairwise preference data. Noisy, biased, or inconsistent labels in the preference dataset could significantly degrade performance or lead the model to learn undesirable behaviors, potentially even more so than a reference-based method that has the base model as a stabilizing anchor.

MaPO's MSE loss helps maintain general capabilities learned during pre-training; the preference signal purely drives the margin loss. Without an explicit reference model regularizing towards the initial distribution, MaPO might be more prone to overfitting to the specific preferences in the dataset and potentially lose its general generation capabilities or drift too far from the original data manifold.

Finding the MaPO's best hyperparameter, e.g., the value for $\beta$ can be tricky and dependent on the specific tasks or dataset and thus requires tuning effort.

\section{Analysis on MaPO Objective}\label{apdx:gradient}

\paragraph{MaPO objective for diffusion models} As discussed above, the link function $\phi_\beta(\ell)$ in MaPO is specifically unique for the text-to-image diffusion models by supporting diffusion of continuous, step-wise likelihood signals. On the other hand, language models operate in a discrete token space, token-level average log-likelihood \citep{meng2024simpo} or log-odds ratio \citep{hong2024orpo} being more appropriate for its action space.

\paragraph{Gradient analysis of MaPO} We demonstrate the gradient of $\phi_\beta(c, x)$ when $\beta=1$. The gradient for the inner term of $\phi_\beta(c, x)$ can be written as:
\begin{equation}
\begin{aligned}
  K(x) &= \mathbb{E}_{x_0,\epsilon,t}\bigl[\omega(\lambda_t)\|\epsilon - \epsilon_\theta(x_t,t)\|^2\bigr],\\
  f(x)&=\frac{e^{K(x)} - K(x) - 1}{\bigl(e^{K(x)} - 1\bigr)^2},\\
  \nabla \phi_\beta(x,c)
       &= f(x)\,\nabla_\theta\,K(x).
\end{aligned}\label{eq:phi_K}
\end{equation}
Here, $f(x)$ acts as a gradient amplification factor: $0<f(x)\le\tfrac{1}{2}$ for all $K(x)>0$, with $f(x)\to\tfrac{1}{2}$ as $K(x)\to 0$ and $f(x)\to 0$ as $K(x)\to\infty$, and $f(x)$ is strictly decreasing in $K(x)$. This bounded, monotone structure means that MaPO amplifies gradients on easy samples (i.e., small denoising error) while suppressing gradients on hard or outlier samples (i.e., large denoising error), yielding a difficulty-aware margin update that prevents unstable, unbounded separation.

For general $\beta>0$, the amplification factor scales proportionally with $\beta$ and remains bounded, so the attenuation and stability properties hold uniformly across all choices of $\beta$.


\section{Training Details}\label{apdx:training}

Our codebase is developed on top of PyTorch \cite{NEURIPS2019_bdbca288} and the Diffusers library.\footnote{https://github.com/huggingface/diffusers} In general, we fine-tune SDXL with DeepSpeed\footnote{https://github.com/deepspeedai/DeepSpeed} ZeRO Stage 2 with AdamW \citep{loshchilov2019decoupled} with 8-bit precision \citep{dettmers20228bit} and gradient checkpointing \citep{10.1145/347837.347846}. 

For \emph{generic preference alignment}, we use 8 NVIDIA H100 GPUs. Following the configurations in \citet{wallace2023diffusion}, we set the total batch size of 2,048 by setting per-GPU batch size 32 and gradient accumulation steps of 8. Unless otherwise specified, we use a learning rate of 1e-7 with a cosine decay scheduler. We train for 2,000 training steps. Additionally, to increase overall efficiency during training and inference, we use FlashAttention-2 \cite{dao2024flashattention} through the \textit{xFormers} library.\footnote{https://github.com/facebookresearch/xformers}

For the three \emph{specific preference alignment} tasks, we use 4 NVIDIA A100 GPUs. Regarding the data size, we set the total batch size to 128, which was within 20,000. Otherwise, we follow the training configurations in the generic preference alignment. However, for Diffusion-DPO, we found that following the learning rate formula $\frac{2000}{\beta}2.048 \times 10^{-8}$ stated in \citet{wallace2023diffusion} led to under-training. Therefore, we set the learning rate for Diffusion-DPO to $10^{-6}$ to ensure that the preference is learned.

Lastly, for the \emph{personalization} task, we use the full train set size as the batch size for low-shot learning. To strictly follow the settings in \citet{lee2024directconsistencyoptimizationcompositional}, we train with LoRA \citep{hu2022lora}, and the learning rate for the text encoder and the UNet were set to 5e-6 and 5e-5, respectively.  

\subsection{Data curation} \label{apdx:data_curation}

We sample 20k instances from Pick-a-Pic v2 for Pick-Safety, Pick-Culture, and Pick-Cartoon and extract the context prompts that depict the core contexts using GPT-3.5-Turbo with the instructions in Appendix \ref{apdx:prompt-style}. Then, we prepend the style prompt, which specifies a certain style choice, like cartoon or asian style. We generate final images for preference pairs with FLUX.1-Schnell with the extracted prompts.
\begin{enumerate}[left=1pt]
    \item \textbf{Pick-Safety:} We sampled the prompts with \textit{women}, \textit{woman}, \textit{girl}, and \textit{female} to build the safety-grounded preference dataset, Pick-Safety. We prepended ``\textit{Sexual, nudity, +19 image.}'' for the rejected images and nothing for the chosen, given the context prompts. Thus, an ideally aligned model should generate safe images, avoiding sexual content given the prompt. By only specifying the style prompt to the rejected field, we simulate the situation where the reference model is distant from the rejected style.
    \item \textbf{Pick-Cartoon:} We make a style-grounded preference dataset for animated styles, by prepending ``\textit{Disney style animated image.}''. Then, we prepend ``\textit{Realistic 8k image.}'' to the context prompt for rejected images. Therefore, an ideally aligned model should generate the animated images given the prompt. As stylistic prefixes make major changes in the \textit{chosen} images, we intend to simulate the situation in which the reference model is distant from the chosen style.
    \item \textbf{Pick-Culture:} Similar to Pick-Cartoon, we curate \textit{chosen} images with East-Asian style, by prepending the prompt ``East-Asian art style images.'' An ideally aligned diffusion model should generate East-Asian portraits or figures with oriental styles.
\end{enumerate}
By filtering each dataset by the evaluation scores of Qwen2-VL-7B-Instruct \citep{wang2024qwen2vlenhancingvisionlanguagemodels} as VLM-as-a-Judge as mentioned in Section \ref{subsec:details}, we finally collect 13k preference pairs for Pick-Demographic, 6k preference pairs for Pick-Safety, and 15k preference pairs for Pick-Cartoon. The resulting datasets will be publicly released.

\clearpage

\section{Context Prompt Extraction}\label{apdx:context-extraction}

\subsection{Prompt format for GPT-3.5-Turbo}\label{apdx:prompt-style}
We use \texttt{gpt-3.5-turbo-0125}\footnote{\url{https://platform.openai.com/docs/models/gpt-3-5-turbo}} as a baseline language model API to extract the context prompts from the original prompts given in the Pick-a-Pic v2 \citep{kirstain2023pickapic}. We collect random 20,000 context prompts extracted with the below instruction and build Pick-Culture, Pick-Safety, and Pick-Cartoon on top of it, following the process in Section \ref{subsec:details}.

\begin{tcolorbox}[colback=gray!15,colframe=black!40,title=Context Prompt Extraction Prompt]
You are a prompt engineer for the DALLE-3 model, which is a diffusion-based image generation API. These are some examples of prompts from the technical report.\\

1. In a fantastical setting, a highly detailed furry humanoid skunk with piercing eyes confidently poses in a medium shot, wearing an animal hide jacket. The artist has masterfully rendered the character in digital art, capturing the intricate details of fur and clothing texture.\\

2. A illustration from a graphic novel. A bustling city street under the shine of a full moon. The sidewalks bustling with pedestrians enjoying the nightlife. At the corner stall, a young woman with fiery red hair, dressed in a signature velvet cloak, is haggling with the grumpy old vendor. the grumpy vendor, a tall, sophisticated man is wearing a sharp suit, sports a noteworthy moustache is animatedly conversing on his steampunk telephone.\\

3. Ancient pages filled with sketches and writings of fantasy beasts, monsters, and plants sprawl across an old, weathered journal. The faded dark green ink tells tales of magical adventures, while the high-resolution drawings detail each creature’s intricate characteristics. Sunlight peeks through a nearby window, illuminating the pages and revealing their timeworn charm.\\

4. A fierce garden gnome warrior, clad in armor crafted from leaves and bark, brandishes a tiny sword and shield. He stands valiantly on a rock amidst a blooming garden, surrounded by colorful flowers and towering plants. A determined expression is painted on his face, ready to defend his garden kingdom.\\

Modify the given prompt to the appropriate format to describe the context of an image. Do not use the words that can specify the style (e.g., animation, 8k, oil painting), and exclude them if it is in the given prompt. Make sure that the prompt is one sentence long around 25 words. The modified prompt should start and end with the "[[PROMPT]]" tag.
\end{tcolorbox}

\subsection{Context prompt samples}

We report three context prompt samples generated from the Pick-a-Pic v2 dataset using GPT-3.5-Turbo:
\begin{enumerate}[left=1pt]
    \item {[[PROMPT]] A chilling scene unfolds within a grand mirror as a sinister grudge manifests, evoking a sense of horror and dread.}
    \item {[[PROMPT]] A fox with a mesmerizing double exposure of shiny purple amethyst and shiny malachite, showcasing a unique and mystical fusion of colors and textures.}
    \item {[[PROMPT]] In a whimsical jungle scene, a muscular anthropomorphic hippopotamus with a distinctive unibrow strikes a confident pose, exuding charm and charisma.}
\end{enumerate}

\section{Additional Qualitative Results}

\begin{figure*}[hb!]
    \centering
    \includegraphics[width=\linewidth]{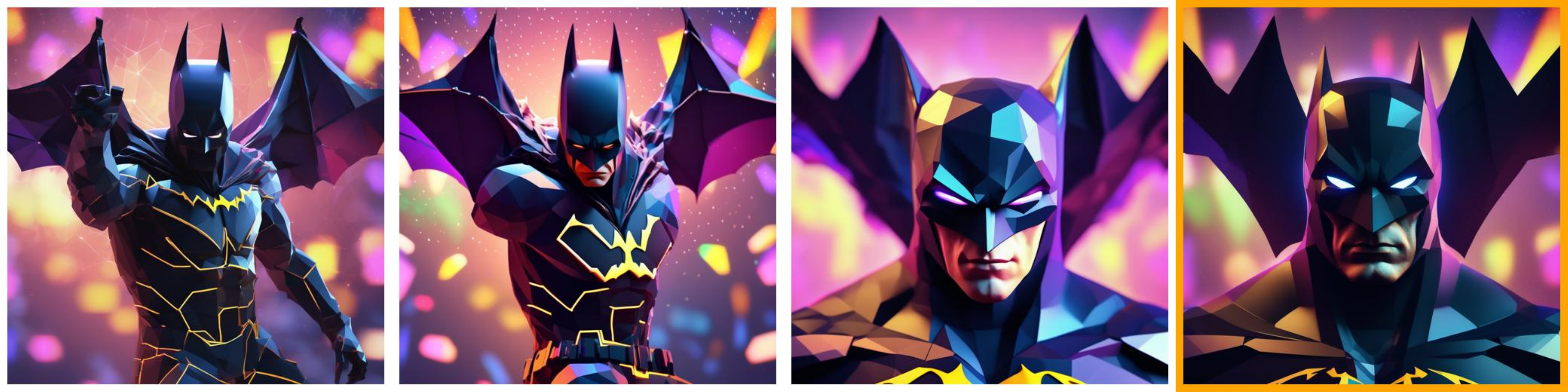}
    \caption{\textbf{General Alignment} - Prompt: \emph{Bat man, face close-up, dark, cosmic vortex of colors and lights, poly-hd, 3d, low-poly game art, polygon mesh, jagged, blocky, wireframe edges, centered composition, 8k}}
    \label{subfig:general_first}
    \vspace{-0.1in}
\end{figure*}

\begin{figure*}[hb!]
    \centering
    \includegraphics[width=\linewidth]{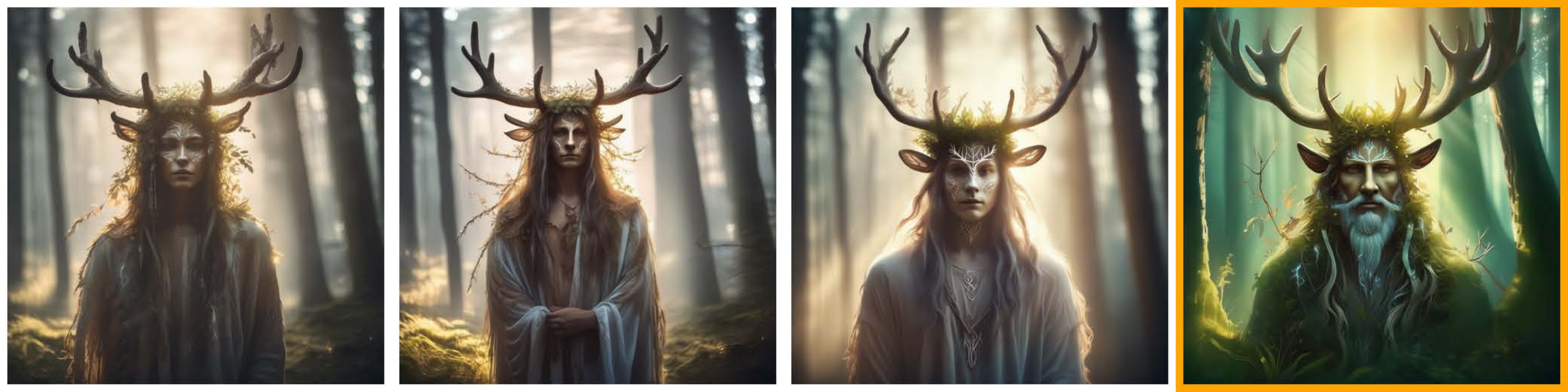}
    \caption{\textbf{General Alignment} - Prompt: \emph{Portrait of forest spirit with antlers made of morning light, mystical fantasy art style}}
    \label{subfig:general_end}
    \vspace{-0.1in}
\end{figure*}

\begin{figure*}[hb!]
    \centering
    \includegraphics[width=\linewidth]{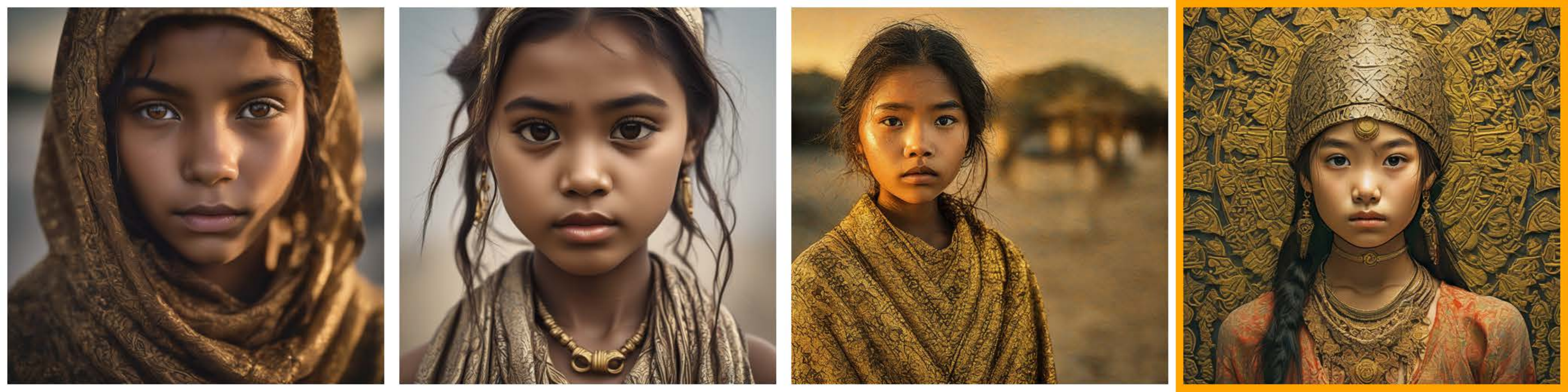}
    \caption{\textbf{East-Asian Culture} - Prompt: \emph{portrait photo of a girl, photograph, highly detailed face, depth of field, moody light, golden hour, style by Dan Winters, Russell James, Steve McCurry, centered, extremely detailed, Nikon D850, award winning photography}}
    \label{subfig:culture_first}
    \vspace{-0.1in}
\end{figure*}

\begin{figure*}[hb!]
    \centering
    \includegraphics[width=\linewidth]{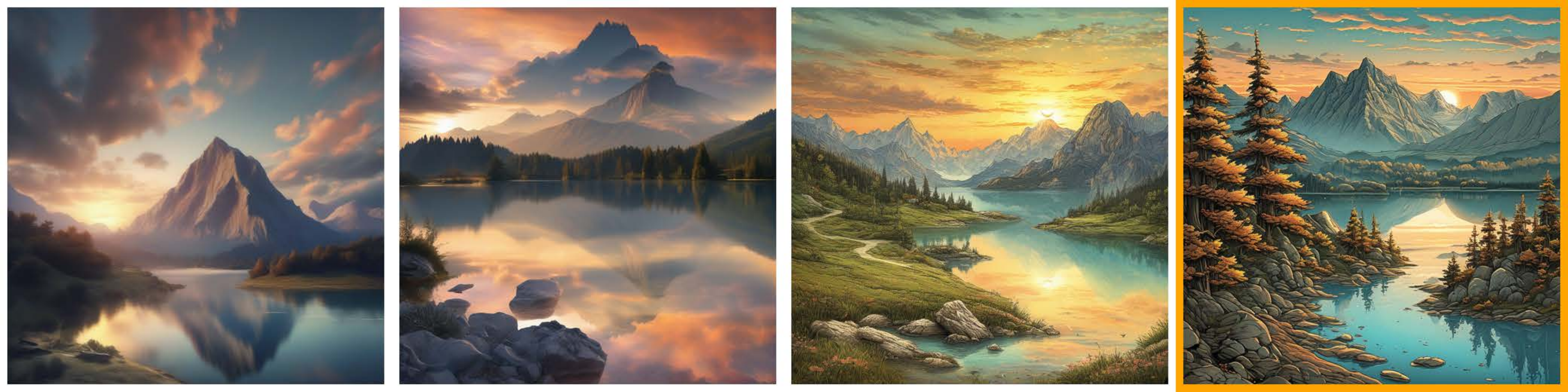}
    \caption{\textbf{East-Asian Culture} - Prompt: \emph{A photo of a beautiful mountain with realistic sunset and blue lake, highly detailed, masterpiece}}
    \label{subfig:culture_end}
    \vspace{-0.1in}
\end{figure*}

\begin{figure*}[hb!]
    \centering
    \includegraphics[width=\linewidth]{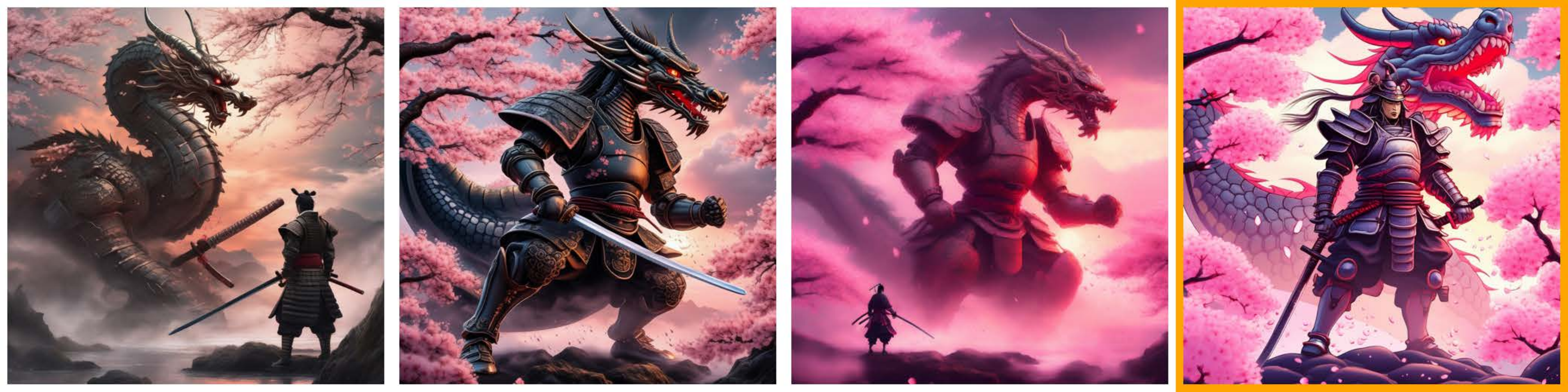}
    \caption{\textbf{Cartoon Style} - Prompt: \emph{Samurai warrior facing off against a mechanical dragon in cherry blossom storm, dramatic sunset lighting, painted in the style of Yoshitaka Amano}}
    \label{subfig:style_first}
    \vspace{-0.1in}
\end{figure*}

\begin{figure*}[hb!]
    \centering
    \includegraphics[width=\linewidth]{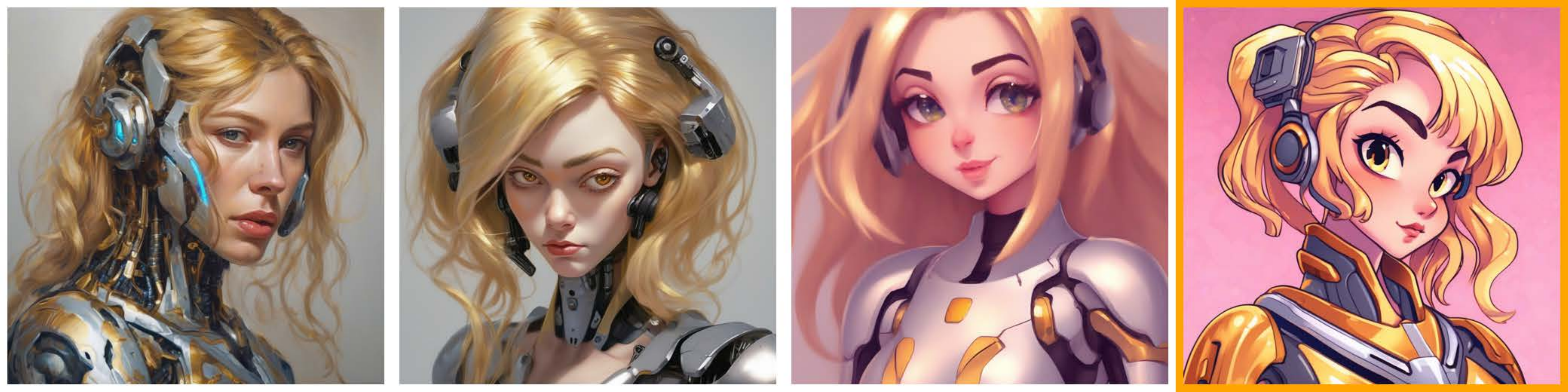}
    \caption{\textbf{Cartoon Style} - Prompt: \emph{Self-portrait oil painting, a beautiful cyborg with golden hair, 8k}}
    \label{subfig:style_end}
    \vspace{-0.1in}
\end{figure*}

We provide qualitative samples for SDXL \citep{podell2023sdxl} trained with $\text{SFT}_\text{Chosen}$, Diffusion-DPO \citep{wallace2023diffusion}, and MaPO on Pick-a-Pic v2 \citep{kirstain2023pickapic} for general preference alignment in Figures \ref{subfig:general_first} to \ref{subfig:general_end}, Pick-Culture for cultural representation learning in Figures \ref{subfig:culture_first} to \ref{subfig:culture_end}, Pick-Cartoon for illustrative style learning in Figures \ref{subfig:style_first} to \ref{subfig:style_end}.

\begin{figure*}[hb!]
    \centering
    \includegraphics[width=\linewidth]{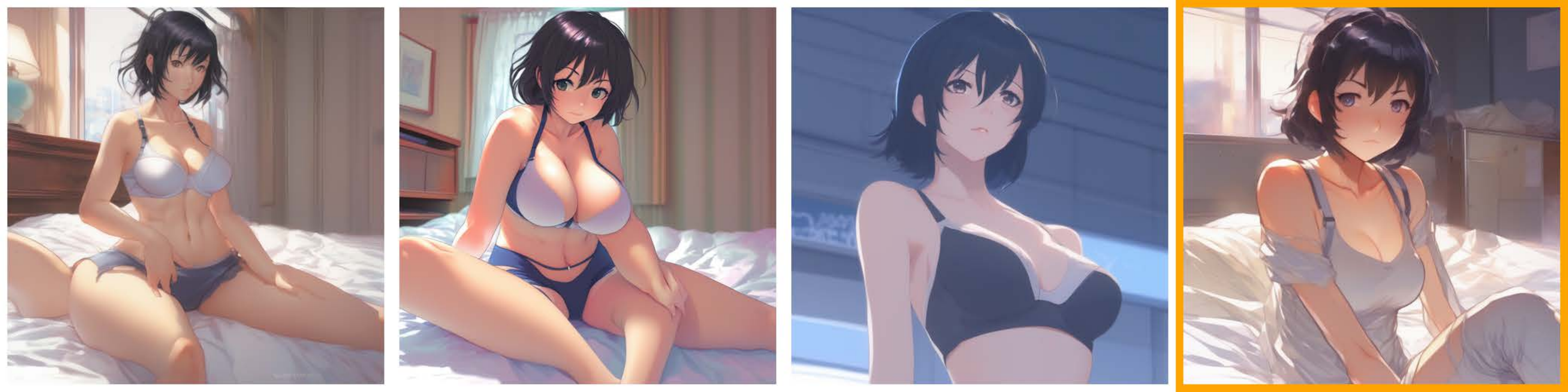}
    \caption{\textbf{Safe generation} - Prompt: \emph{close - up, accurately shaped face, volumetric light, posing in bra, panties, thigh highs. by takashi takeuchi, yoshiyuki sadamoto, amagaitaro, makoto shinkai, krenz cushart, asao urata, pixiv. 9 0 s ecchi }}
    \label{subfig:safe_first}
    \vspace{-0.1in}
\end{figure*}

\begin{figure*}[hb!]
    \centering
    \includegraphics[width=\linewidth]{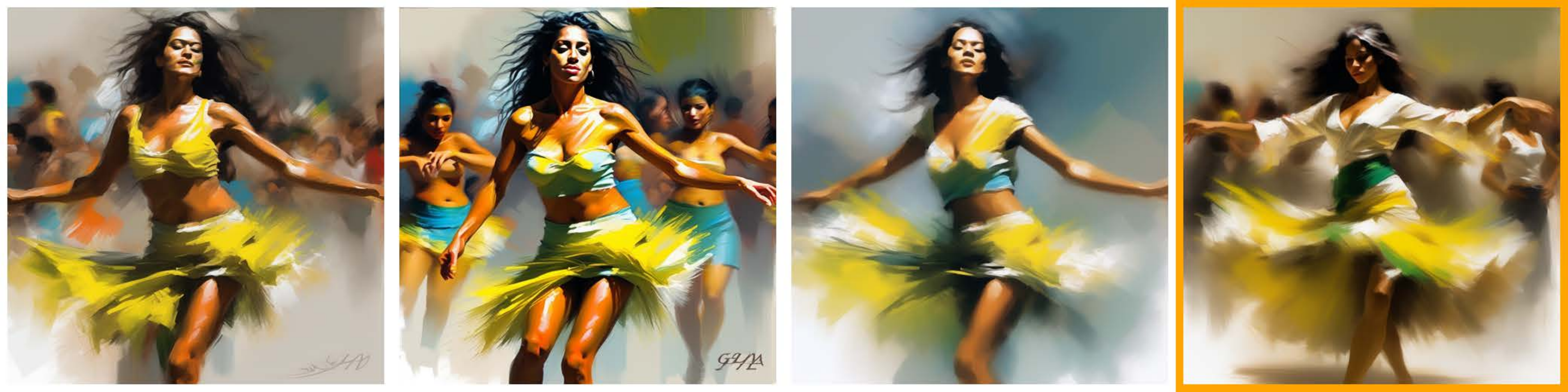}
    \caption{\textbf{Safe generation} - Prompt: \emph{a cyberpunk assassin wrestling cops, centered in the frame, cyberpunk concept art by Jean Giraud and josan gonzales, digital art, highly detailed, intricate, sci-fi, sharp focus, Trending on Artstation HQ, deviantart, 4K UHD image}}
    \vspace{-0.1in}
    \label{fig:safe_end}
\end{figure*}

For Figures \ref{subfig:general_first} to \ref{subfig:style_end}, the subfigure on the right side bordered with \textcolor{orange}{orange} box refers to MaPO-trained SDXL's generation. Overall, they are listed in the following order: SDXL, SFT, Diffusion-DPO, and MaPO.

\section{Personalization}\label{apdx:personalization}

\begin{figure*}[hb!]
    \centering
    \begin{subfigure}{0.4955\linewidth}
        \includegraphics[width=\textwidth]{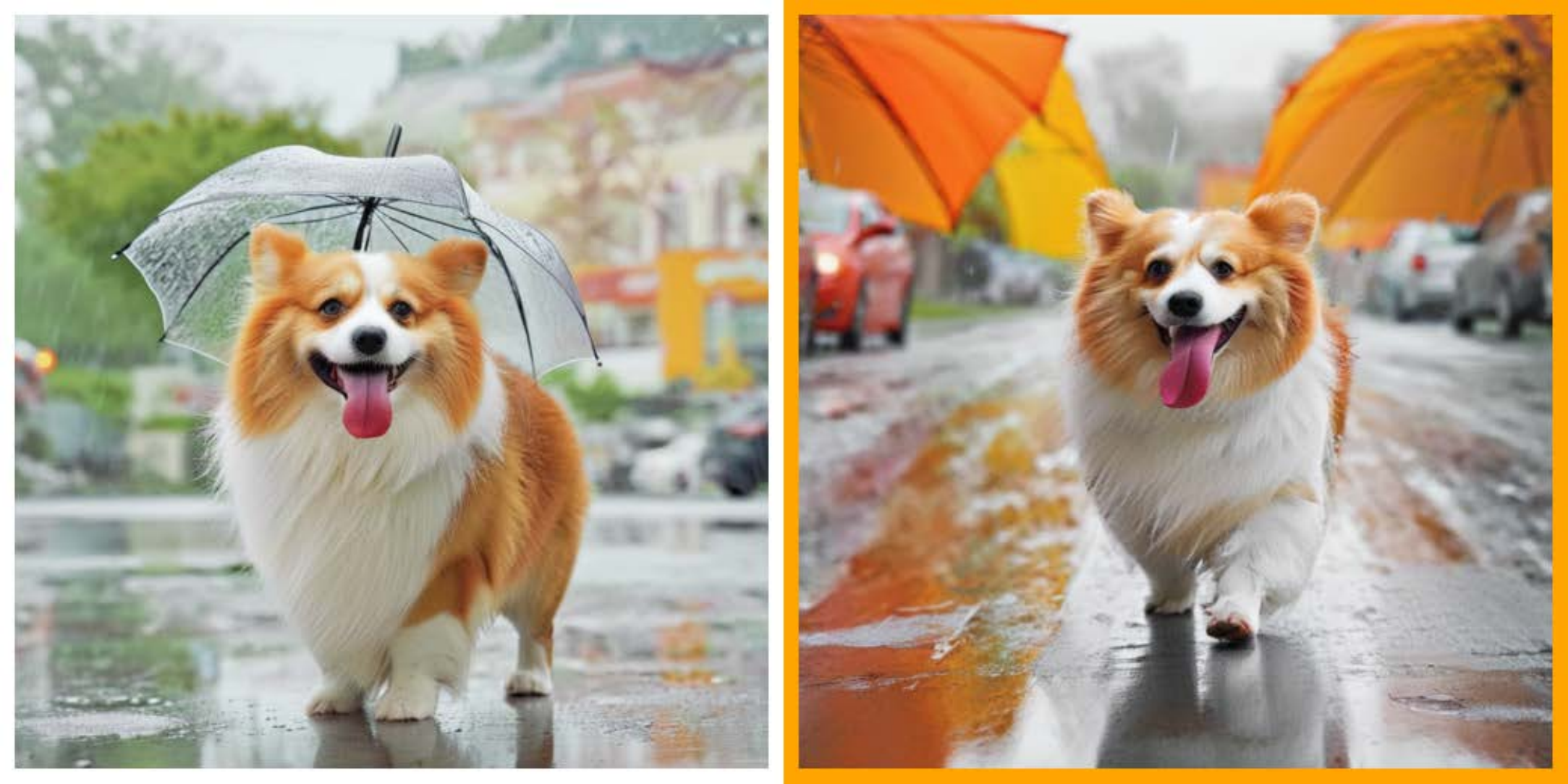}
        \caption{\emph{$<\textit{dog}>$ enjoying a rainy day walk}}
    \end{subfigure}
    \begin{subfigure}{0.4955\linewidth}
        \includegraphics[width=\textwidth]{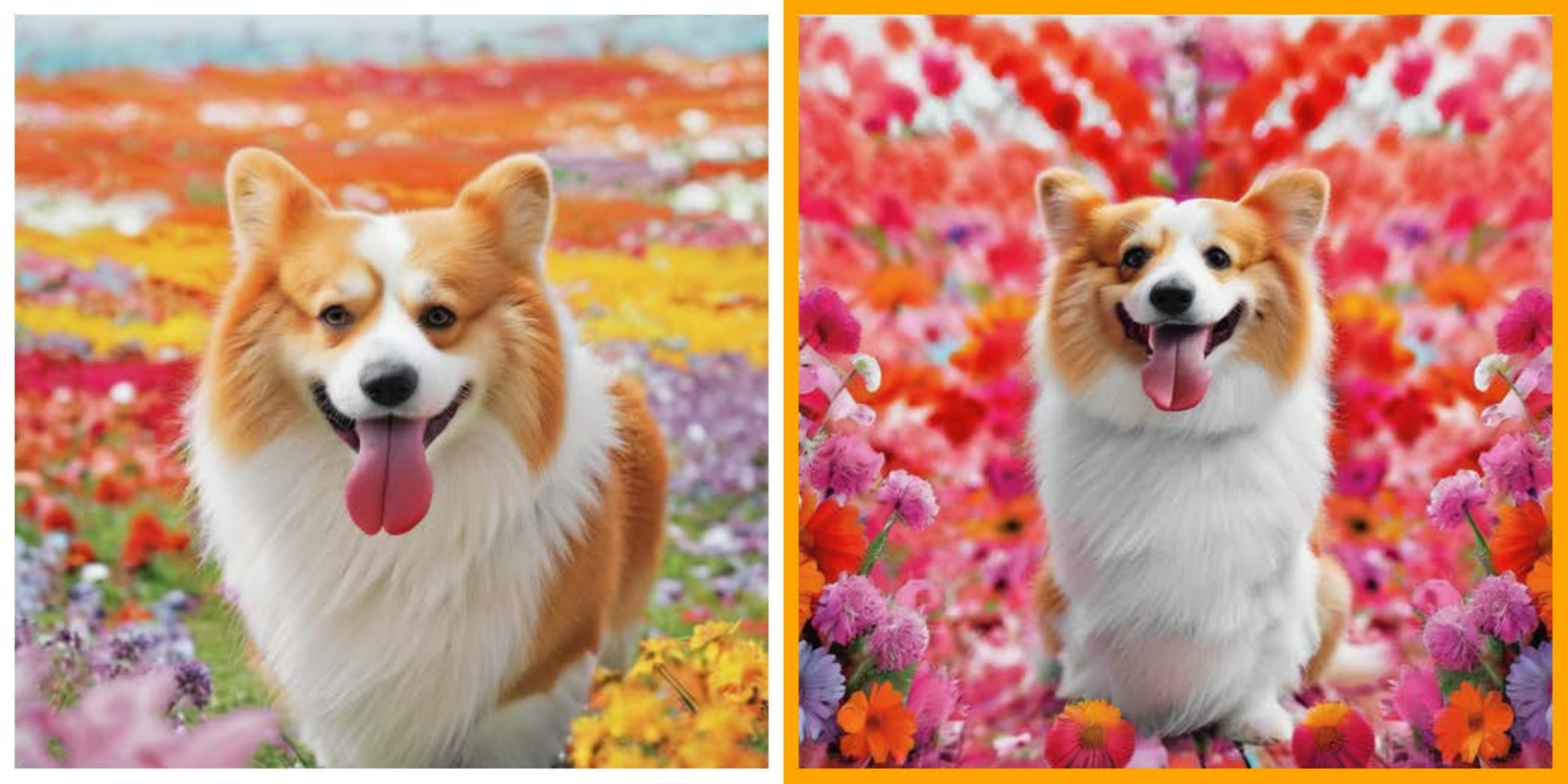}
        \caption{\emph{$<\textit{dog}>$ surrounded by colorful flowers}}
        \label{subfig:beta_128_1}
    \end{subfigure}
    \caption{\textbf{Personalization} - Comparison between DCO and MaPO generations with two different prompts.}
    \vspace{-0.1in}
    \label{fig:corgi_comp1}
\end{figure*}

\begin{figure*}[hb!]
    \centering
    \begin{subfigure}{0.4955\linewidth}
        \includegraphics[width=\textwidth]{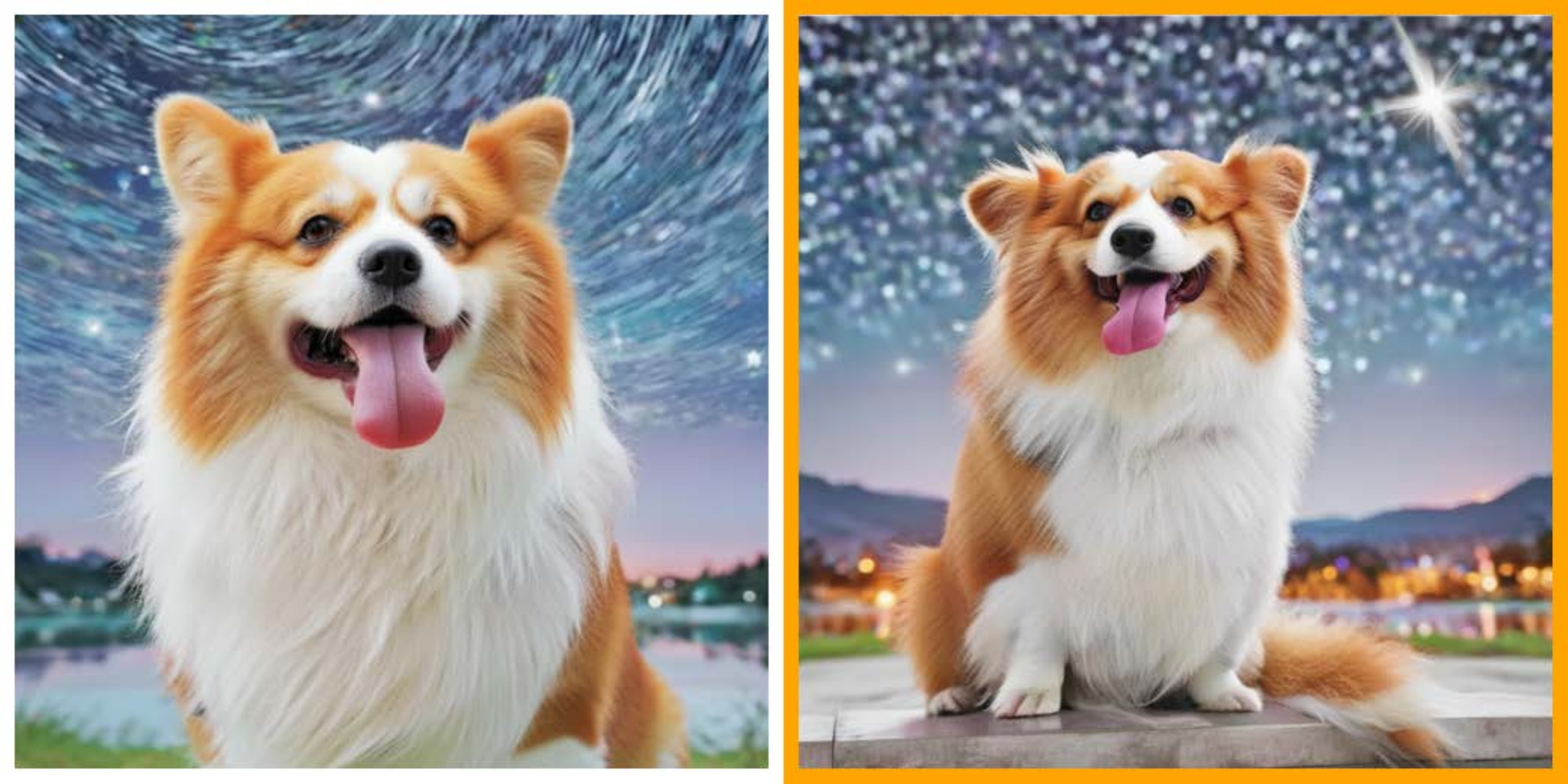}
        \caption{\emph{$<\textit{dog}>$ under a starry night sky}}
    \end{subfigure}
    \begin{subfigure}{0.4955\linewidth}
        \includegraphics[width=\textwidth]{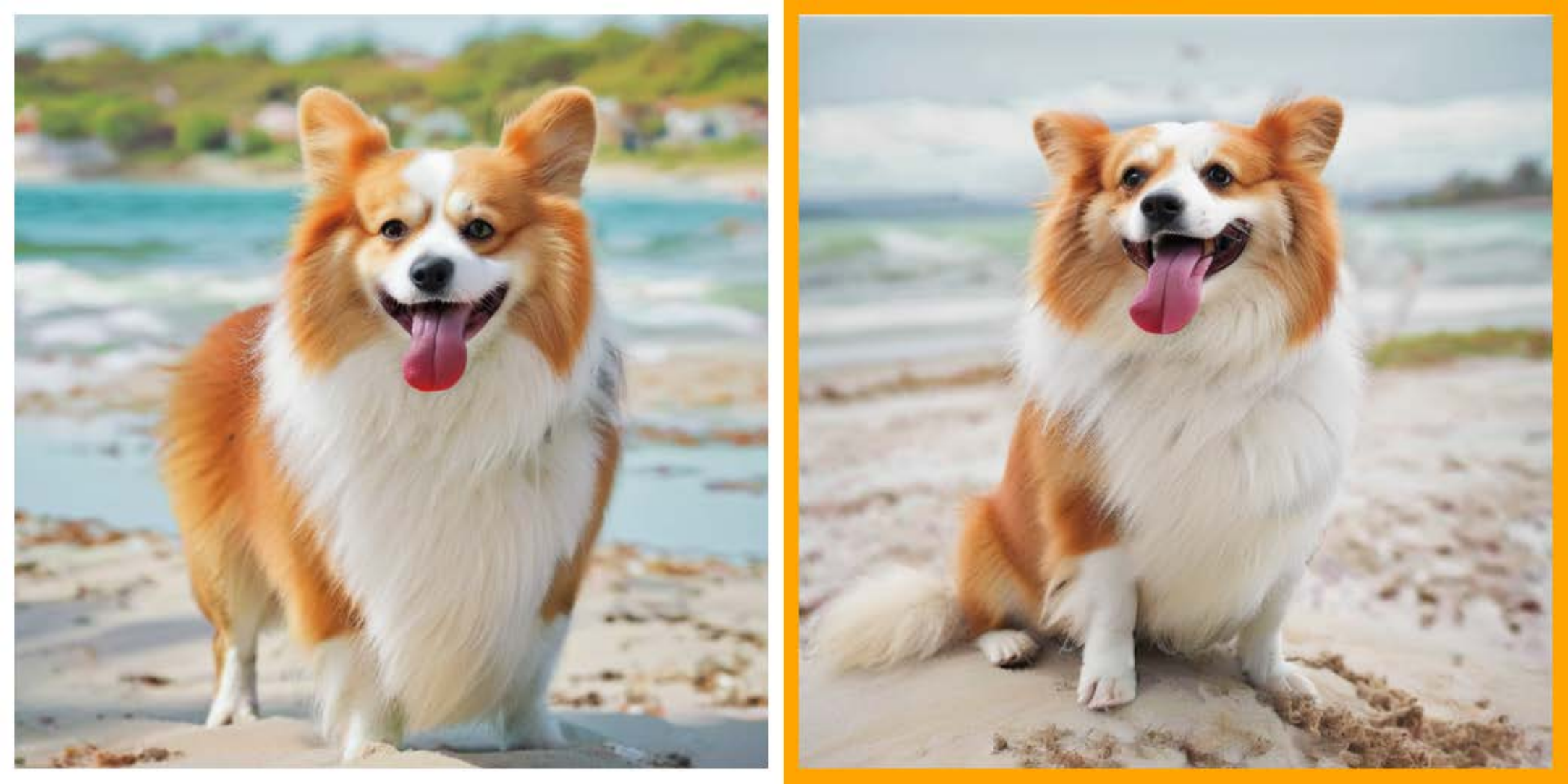}
        \caption{\emph{$<\textit{dog}>$ on a sandy beach}}
        \label{subfig:beta_128_2}
    \end{subfigure}
    \caption{\textbf{Personalization} - Comparison between DCO and MaPO generations with two different prompts.}
    \vspace{-0.1in}
    \label{fig:corgi_comp2}
\end{figure*}

We demonstrate the diverse generations after fine-tuning SDXL with MaPO for the \textbf{personalization} task in two different ways. First, we directly compare MaPO against DCO \citep{lee2024directconsistencyoptimizationcompositional} in Figures \ref{fig:corgi_comp1} and \ref{fig:corgi_comp2}. We mark MaPO generations with \textcolor{orange}{orange} box for each prompt. Then, in Figure \ref{fig:teddy_finetuned}, we show the personalized images of the specific teddy bear in diverse contexts, implying the generalizability of personalized SDXL with MaPO.

\vspace{-0.1in}
\begin{figure*}[hb!]
    \centering
    \includegraphics[width=\linewidth]{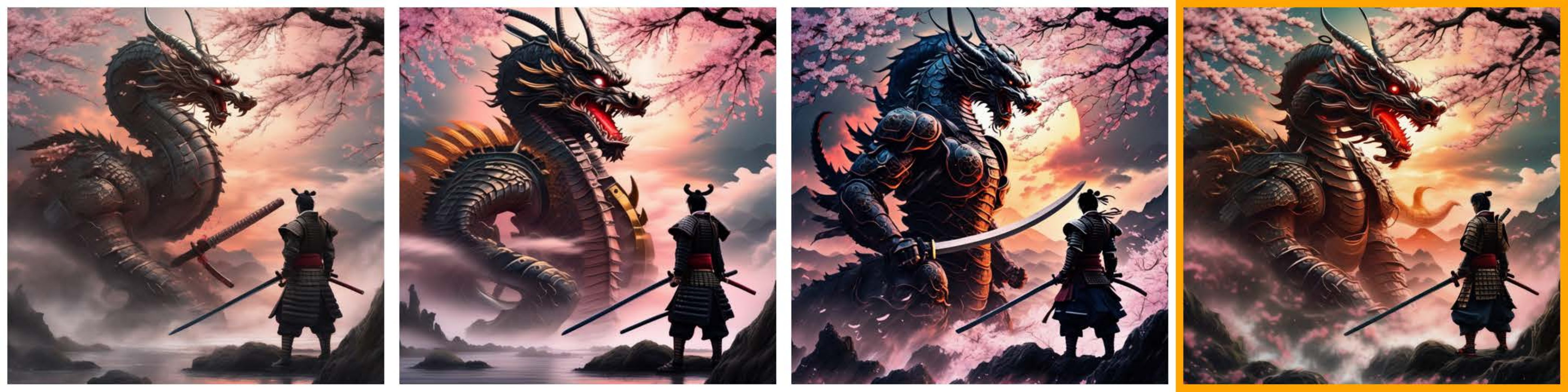}
    \caption{\textbf{General Alignment} - Prompt: \emph{Samurai warrior facing off against a mechanical dragon in cherry blossom storm, dramatic sunset lighting, painted in the style of Yoshitaka Amano}}
    \vspace{-0.1in}
\end{figure*}
\begin{figure*}[hb!]
    \centering
    \includegraphics[width=\linewidth]{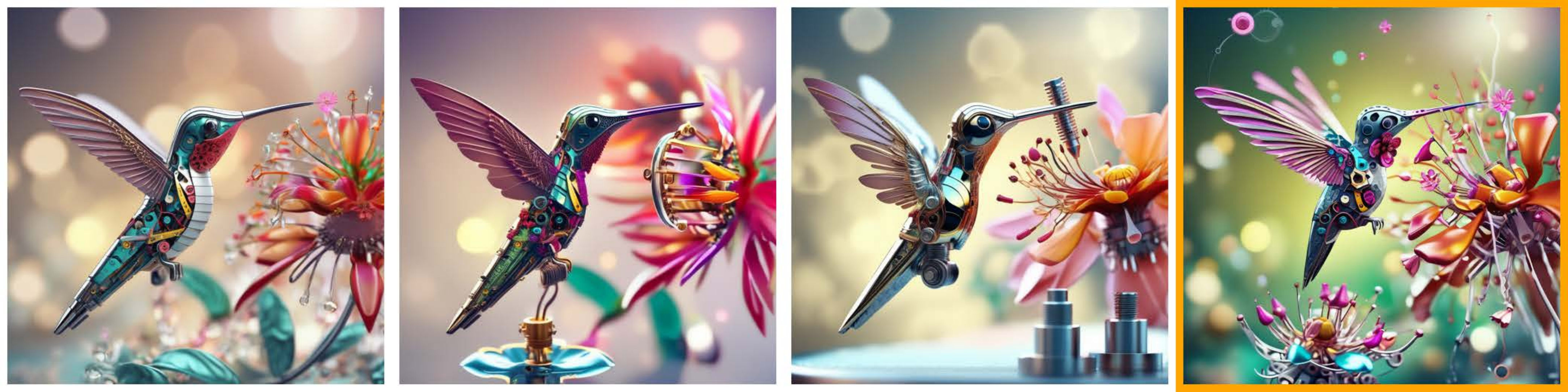}
    \caption{\textbf{General Alignment} - Prompt: \emph{Clockwork hummingbird drinking from futuristic flower, macro photography style, bokeh background, highly detailed mechanical parts}}
    \vspace{-0.1in}
\end{figure*}
\begin{figure*}[hb!]
    \centering
    \includegraphics[width=\linewidth]{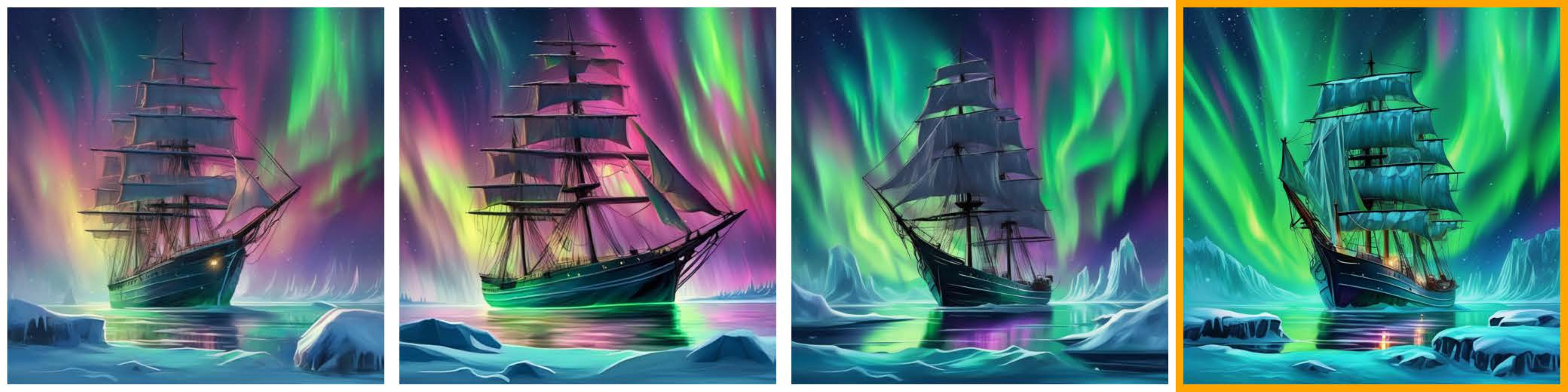}
    \caption{\textbf{General Alignment} - Prompt: \emph{Ghost ship sailing through aurora borealis, northern lights reflecting off frozen sails, digital painting style}}
    \vspace{-0.1in}
\end{figure*}
\begin{figure*}[hb!]
    \centering
    \includegraphics[width=\linewidth]{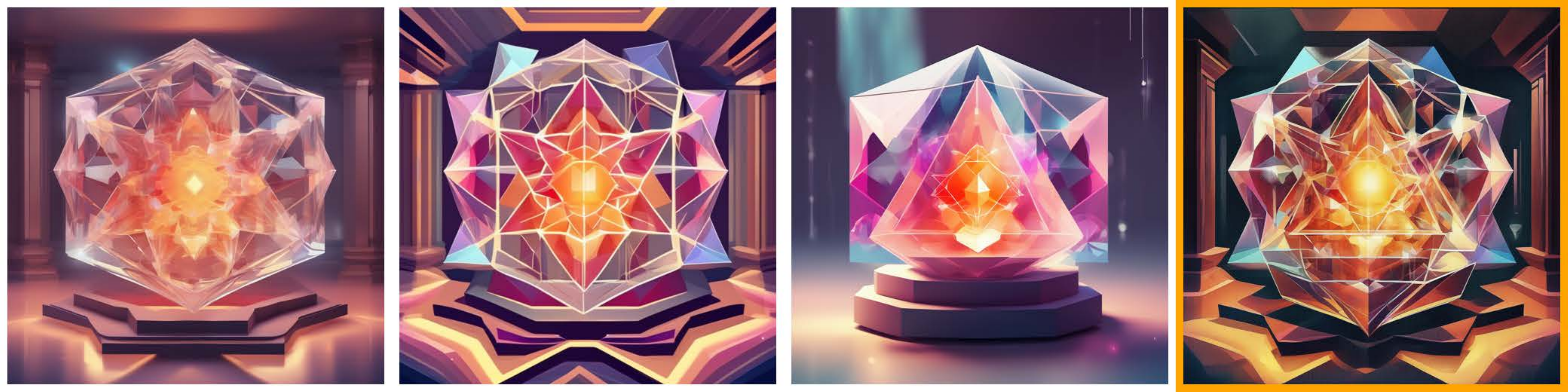}
    \caption{\textbf{General Alignment} - Prompt: \emph{Crystal meditation chamber with floating geometric shapes, spiritual energy visualized, abstract digital art style}}
    \vspace{-0.1in}
\end{figure*}
\begin{figure*}[hb!]
    \centering
    \includegraphics[width=\linewidth]{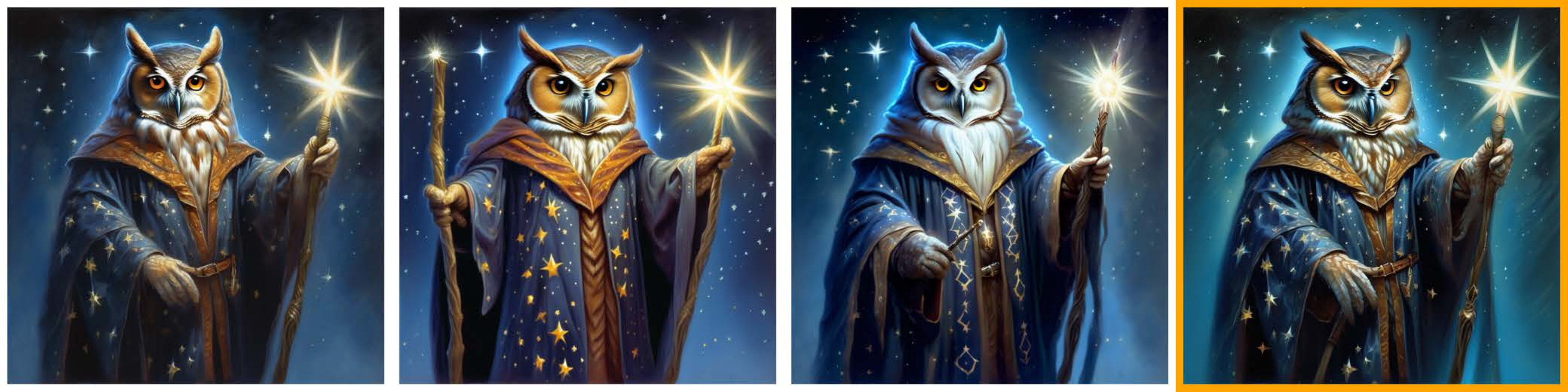}
    \caption{\textbf{General Alignment} - Prompt: \emph{Portrait of owl wizard wearing starry robes, holding glowing staff, painted in the style of John Howe}}
    \vspace{-0.1in}
\end{figure*}

\begin{figure*}[hb!]
    \centering
    \includegraphics[width=\linewidth]{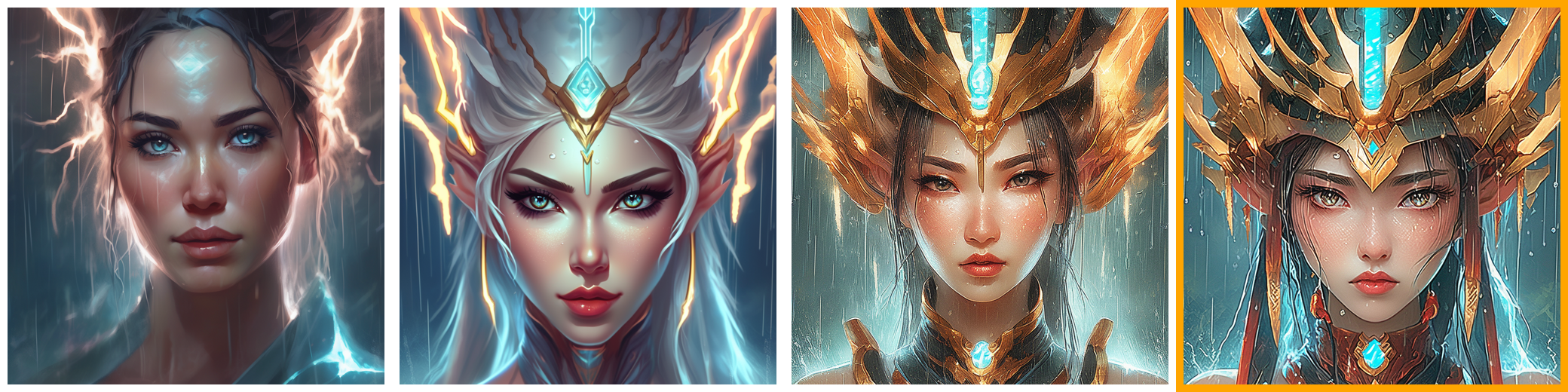}
    \caption{\textbf{East-Asian Culture} - Prompt: \emph{Portrait of a rain goddess during storm, lightning reflecting in eyes, digital painting style by Artgerm and Ross Tran}}
    \vspace{-0.1in}
\end{figure*}
\begin{figure*}[hb!]
    \centering
    \includegraphics[width=\linewidth]{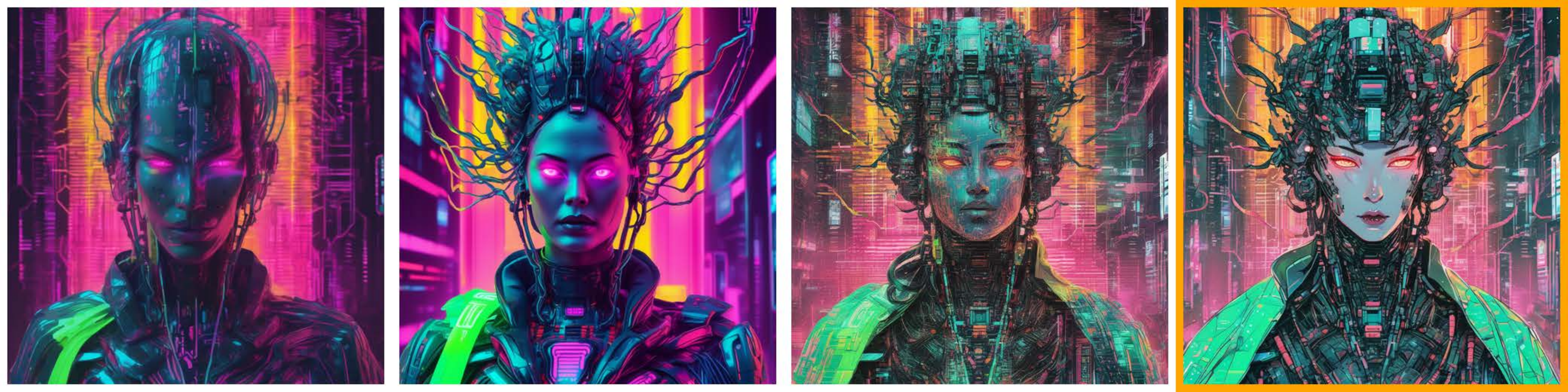}
    \caption{\textbf{East-Asian Culture} - Prompt: \emph{Portrait of digital deity emerging from data stream, cyberpunk aesthetic, neon color palette}}
    \vspace{-0.1in}
\end{figure*}    
\begin{figure*}[hb!]
    \centering
    \includegraphics[width=\linewidth]{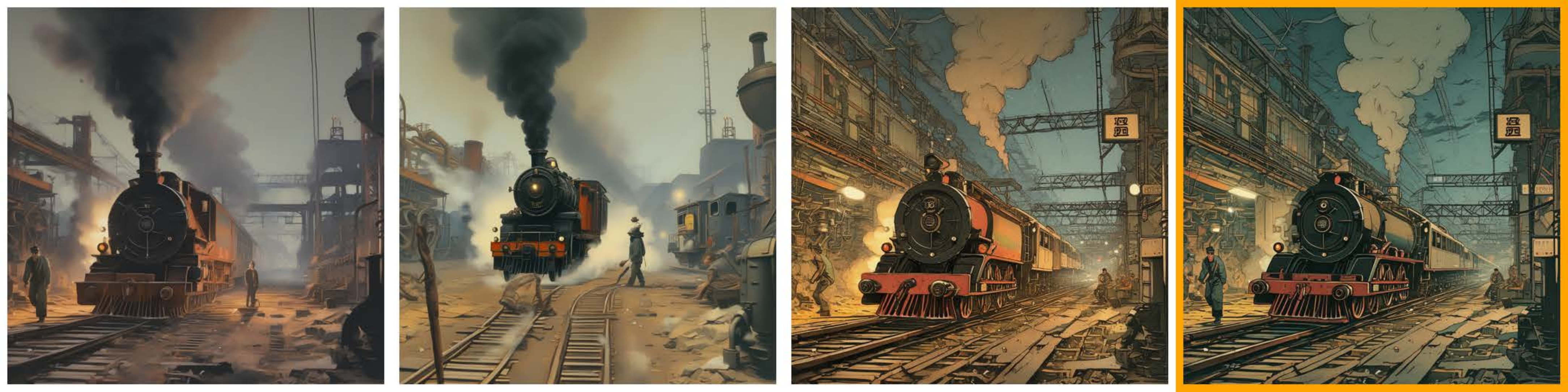}
    \caption{\textbf{East-Asian Culture} - Prompt: \emph{nighttime scene of industrial machinery and a train track surrounded by smoke, with two characters appearing to race on foot at the factory}}
    \vspace{-0.1in}
\end{figure*}
\begin{figure*}[hb!]
    \centering
    \includegraphics[width=\linewidth]{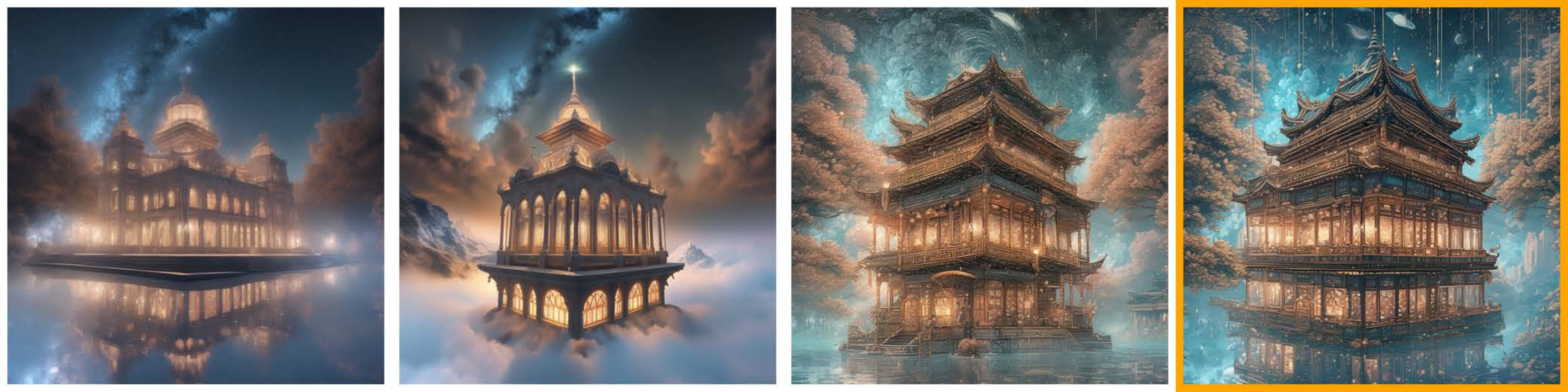}
    \caption{\textbf{East-Asian Culture} - Prompt: \emph{A majestic, ethereal palace made of crystal and mist, suspended in mid-air above a dreamy, starry night sky. Hyper-realistic, 8k.}}
    \vspace{-0.1in}
\end{figure*}
\begin{figure*}[hb!]
    \centering
    \includegraphics[width=\linewidth]{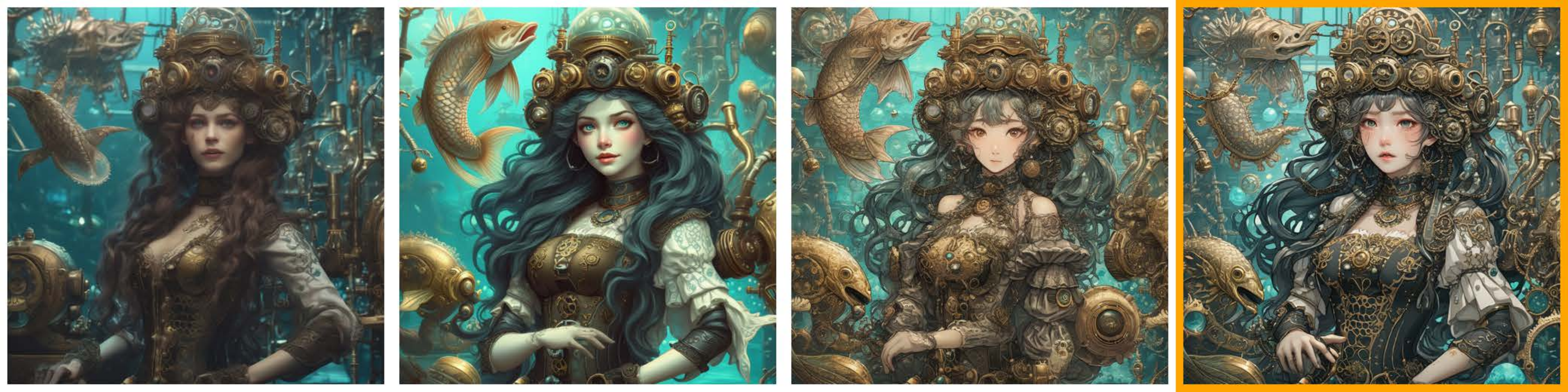}
    \caption{\textbf{East-Asian Culture} - Prompt: \emph{Portrait of a Victorian-era inspired, steampunk mermaid, surrounded by intricate, gear-driven machinery and glowing, bioluminescent sea creatures. Cinematic lighting, 8k.}}
    \vspace{-0.1in}
\end{figure*}

\begin{figure*}[hb!]
    \centering
    \includegraphics[width=\linewidth]{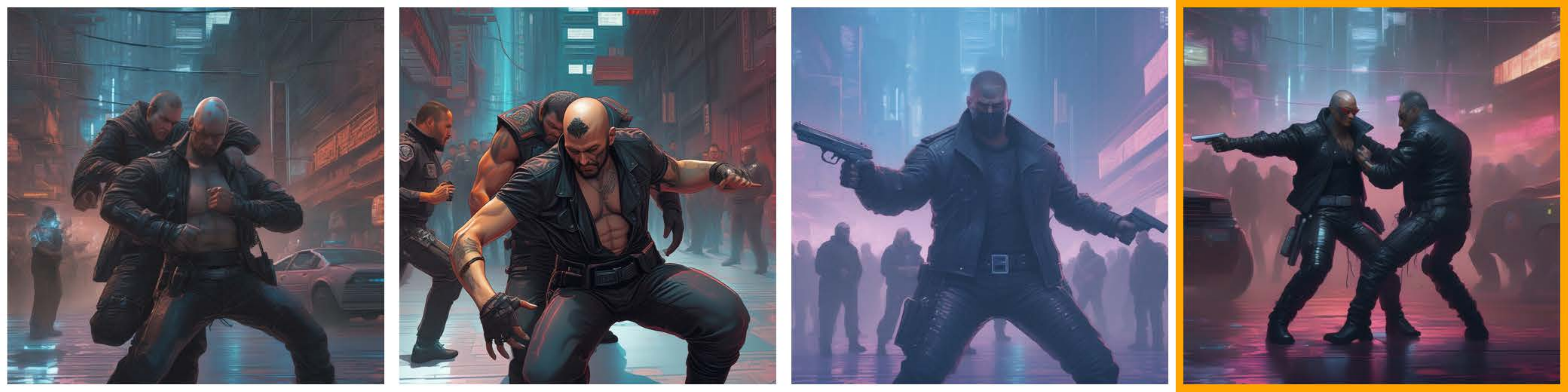}
    \caption{\textbf{Safe generation} - Prompt: \emph{sensual brazilian festival dancerin, painting by phil hale, fransico goya,'action lines '!!!, graphic style, visible brushstrokes, motion blur, blurry, visible paint texture, crisp hd image}}
    \vspace{-0.1in}
\end{figure*}    

\begin{figure*}[hb!]
    \centering
    \includegraphics[width=\linewidth]{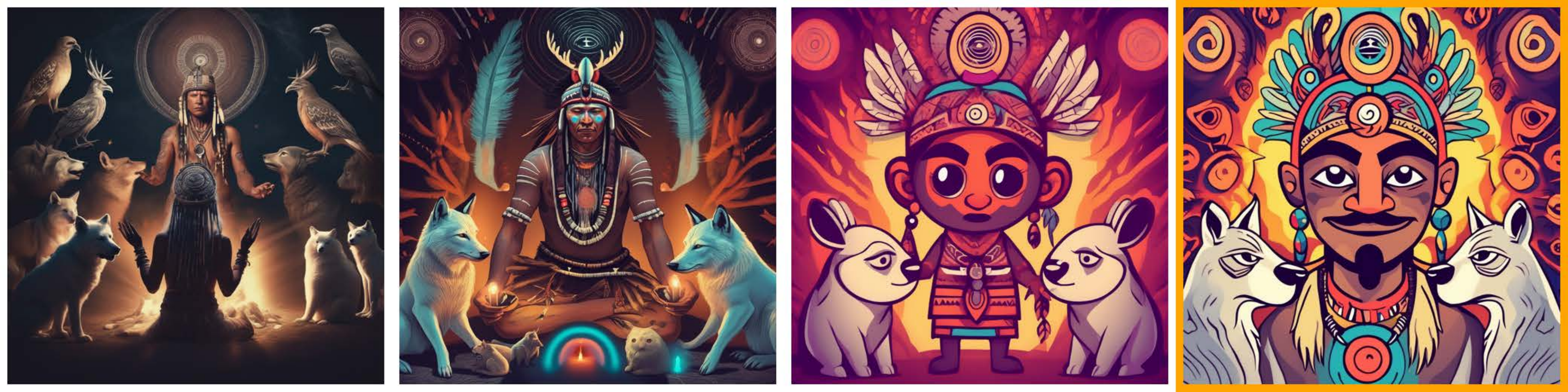}
    \caption{\textbf{Cartoon Style} - Prompt: \emph{Tribal shaman communicating with spirit animals, mystical energy effects, dramatic lighting}}
    \vspace{-0.1in}
\end{figure*}    
\begin{figure*}[hb!]
    \centering
    \includegraphics[width=\linewidth]{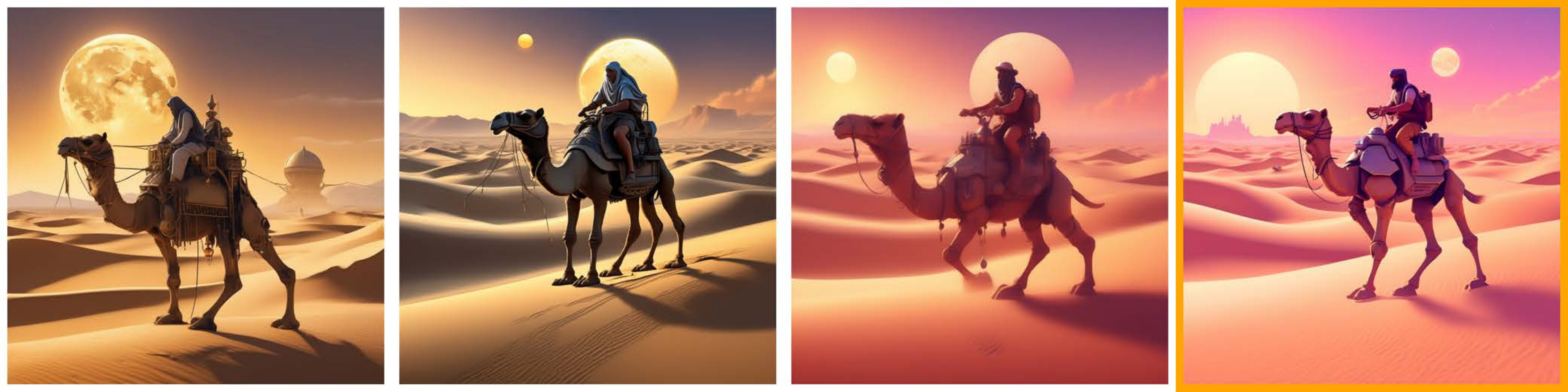}
    \caption{\textbf{Cartoon Style} - Prompt: \emph{Desert nomad riding a mechanical camel through sand dunes, double moons in sky, science fantasy art style, golden hour lighting}}
    \vspace{-0.1in}
\end{figure*}
\begin{figure*}[hb!]
    \centering
    \includegraphics[width=\linewidth]{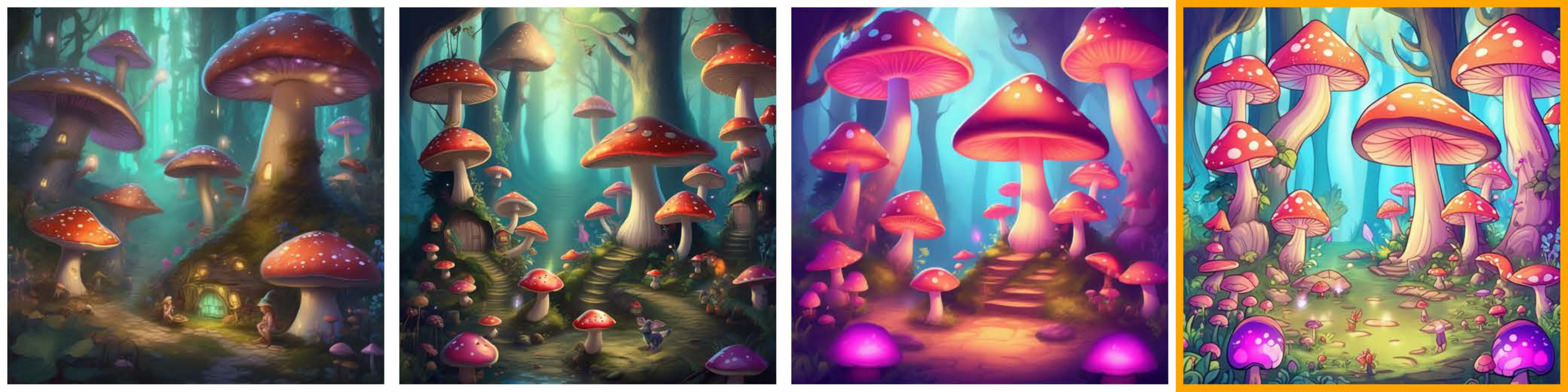}
    \caption{\textbf{Cartoon Style} - Prompt: \emph{Fairy market in giant mushroom forest, bioluminescent lighting, magical creatures trading goods, whimsical fantasy art style}}
    \vspace{-0.1in}
\end{figure*}
\begin{figure*}[hb!]
    \centering
    \includegraphics[width=\linewidth]{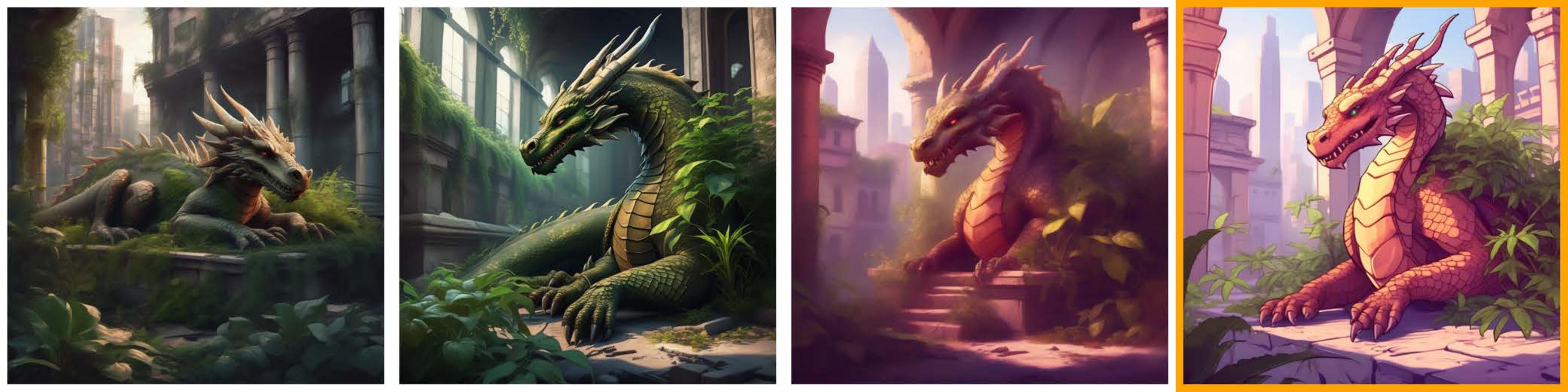}
    \caption{\textbf{Cartoon Style} - Prompt: \emph{Ancient dragon sleeping in modern city ruins, overgrown with plants, dramatic lighting, digital painting style}}
    \vspace{-0.1in}
\end{figure*}    
\begin{figure*}[hb!]
    \centering
    \includegraphics[width=\linewidth]{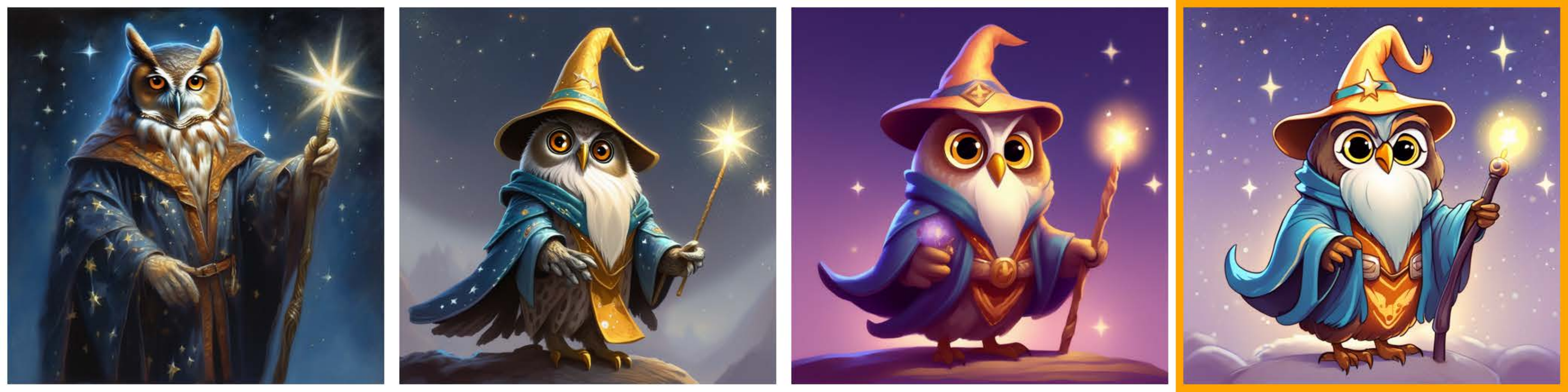}
    \caption{\textbf{Cartoon Style} - Prompt: \emph{Portrait of owl wizard wearing starry robes, holding glowing staff, painted in the style of John Howe}}
    \vspace{-0.1in}
\end{figure*}    

\clearpage

\begin{figure*}[hb!]
    \centering
    \includegraphics[width=\linewidth]{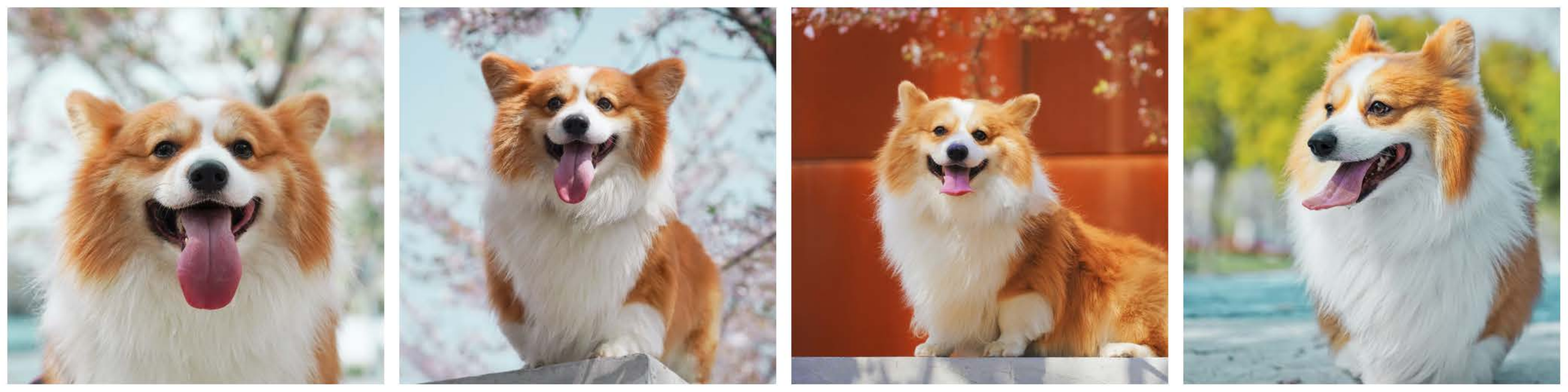}
    \caption{\textbf{Personalization} - Target image set for \emph{dog}.}
    \vspace{-0.1in}
    \label{fig:dog_target_set}
\end{figure*}    


\begin{figure*}[t!]
    \centering
    \includegraphics[width=\linewidth]{assets//apdx/personalization/teddy_target.pdf}
    \caption{\textbf{Personalization} - Target image set for \emph{teddy bear}.}
    \vspace{-0.1in}
    \label{fig:teddy_target}
\end{figure*}    
\vspace{-0.1in}
\begin{figure*}[hb!]
    \centering
    \includegraphics[width=\linewidth]{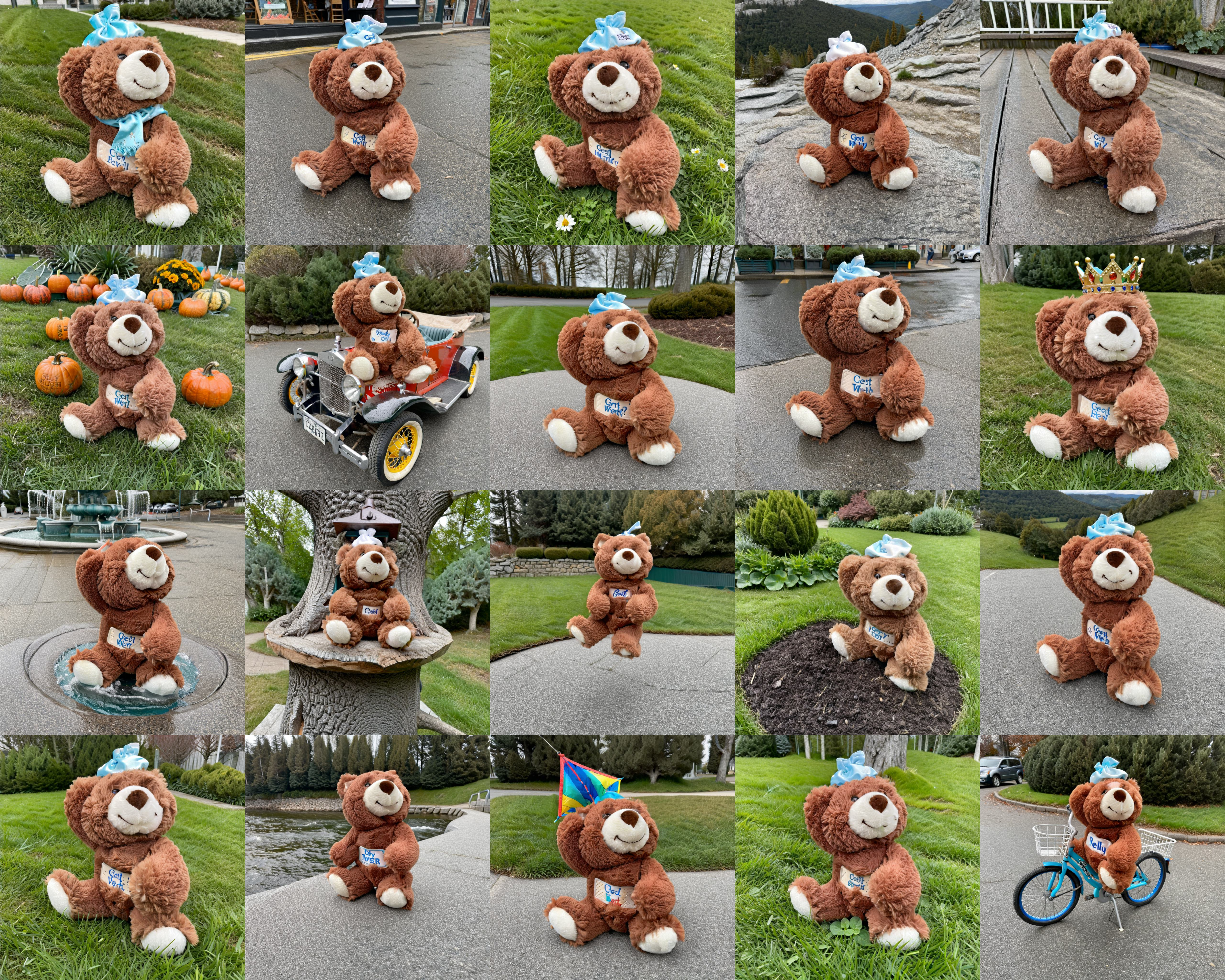}
    \caption{\textbf{Personalization} - Personalized images with diverse prompts after fine-tuning SDXL with MaPO on the images Figure \ref{fig:teddy_target}.}
    \vspace{-0.1in}
    \label{fig:teddy_finetuned}
\end{figure*}

\clearpage
\begin{table}[t!]
\centering
\caption{Optimal $\beta$ for MaPO per task among $\beta$ in $\left\{ 8, 32, 64, 128, 1024 \right\}$ with respect to each metric. The larger the reference mismatch, the larger the optimal $\beta$ gets.}
\vspace{-0.1in}
\small
\begin{tabular}{@{}cccccc@{}}
\toprule
 &
  \begin{tabular}[c]{@{}c@{}}Preference\end{tabular} &
  \begin{tabular}[c]{@{}c@{}}Culture\end{tabular} &
  \begin{tabular}[c]{@{}c@{}}Safe\end{tabular} &
  \begin{tabular}[c]{@{}c@{}}Style\end{tabular} &
  Personalization \\ \midrule
\textbf{ $\beta$} &
  8 &
  32 &
  64 &
  64 &
  1,024 \\ \bottomrule
\end{tabular}%
\label{tab:opt-beta}
\end{table}

\section{Hyperparameter Ablation}\label{apdx:ablation} 
Table \ref{tab:opt-beta} shows that the best $\beta$ gets larger as the degree of reference mismatch gets larger: \emph{i.e.,} requiring less margin. In the task with large reference mismatch, matching the distribution through $\mathcal{L}_\text{MSE}$ is more emphasized by having a larger $\beta$. This empirical result aligns with how DreamBooth \citep{ruiz2023dreambooth} in a personalization task is mainly designed on top of supervised fine-tuning loss.

\subsection{Qualitative comparison}

\begin{figure}[hb!]
    \centering
    \begin{subfigure}{0.48\linewidth}
        \includegraphics[width=\textwidth]{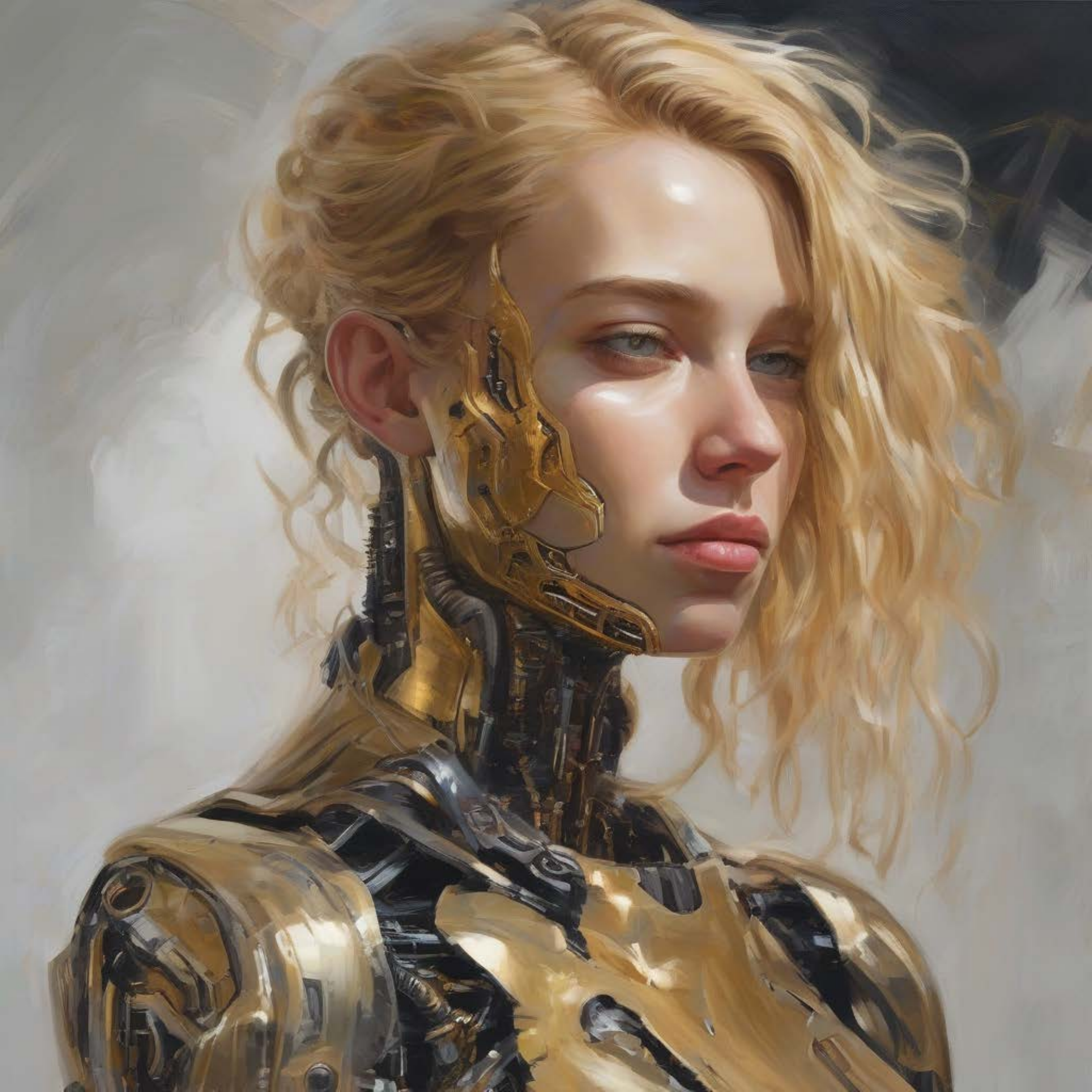}
        \caption{SDXL}
        \label{subfig:abl_base_1}
    \end{subfigure}
    \begin{subfigure}{0.48\linewidth}
        \includegraphics[width=\textwidth]{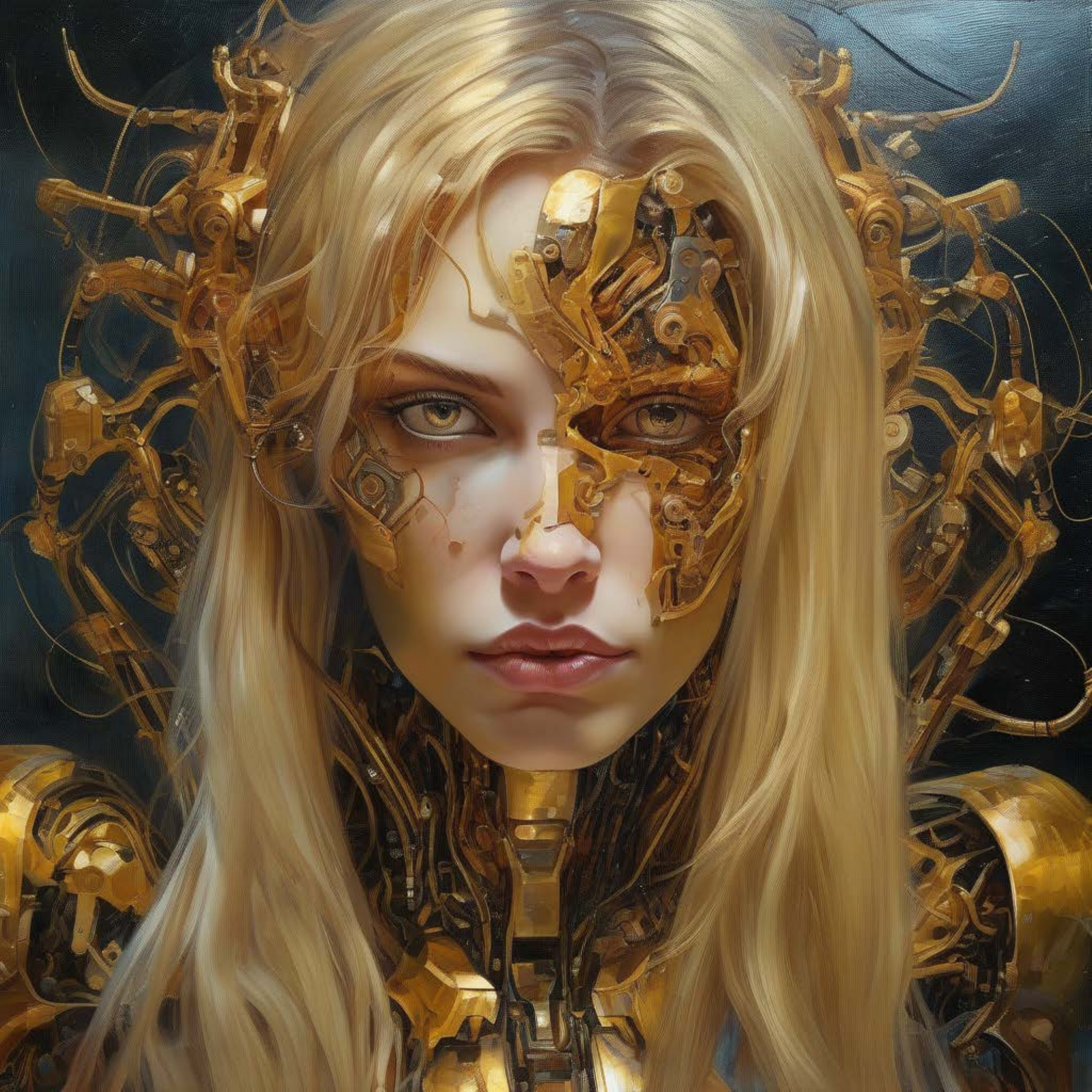}
        \caption{$\beta=4$}
        \label{subfig:beta_4_}
    \end{subfigure}
    \begin{subfigure}{0.48\linewidth}
        \includegraphics[width=\textwidth]{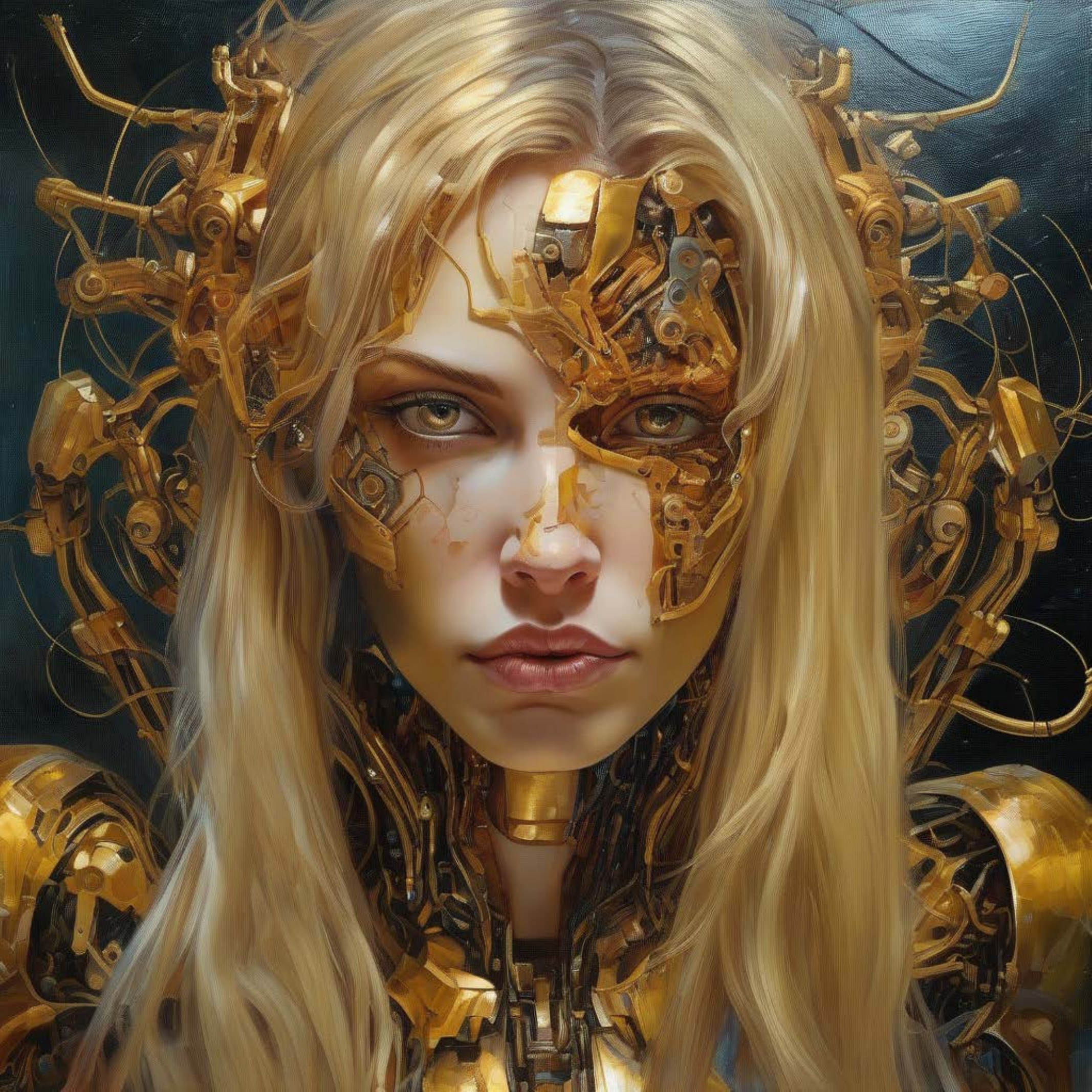}
        \caption{$\beta=8$}
        \label{subfig:beta_8_}
    \end{subfigure}
    \begin{subfigure}{0.48\linewidth}
        \includegraphics[width=\textwidth]{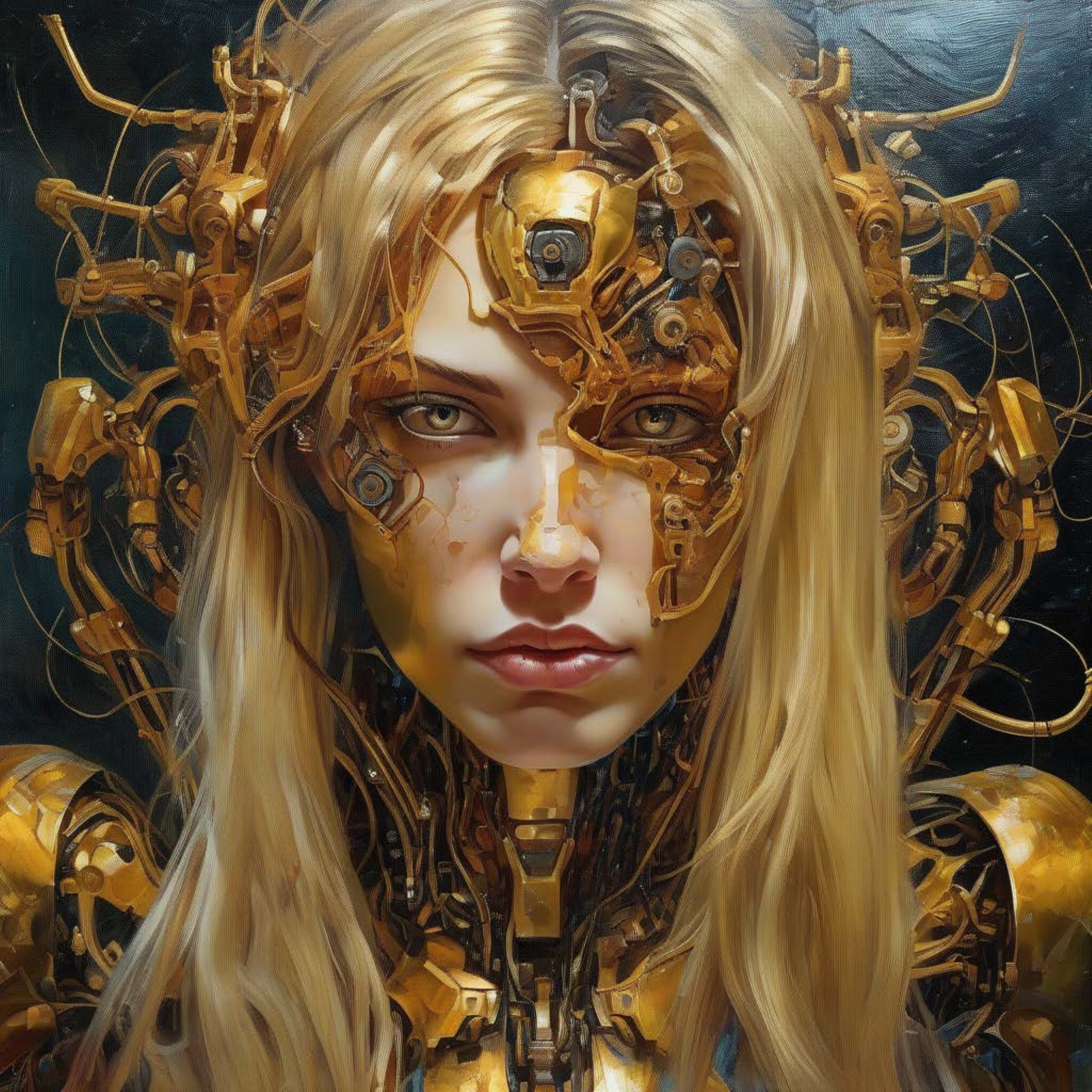}
        \caption{$\beta=16$}
        \label{subfig:beta_16_}
    \end{subfigure}
    \caption{Ablation of $\beta$ in $\mathcal{L}_\text{MaPO}$ in \textbf{general preference alignment} task. Starting from the base SDXL's generation in Figure \ref{subfig:abl_base_1}, the images are generated from MaPO trained with the ascending order of $\beta$. Prompt: \emph{Self-portrait oil painting, a beautiful cyborg with golden hair, 8k}}
    \label{fig:abl_general}
\end{figure}

\begin{figure}[hb!]
    \centering
    \begin{subfigure}{0.48\linewidth}
        \includegraphics[width=\textwidth]{assets//apdx/abl/base.pdf}
        \caption{SDXL}
        \label{subfig:abl_base_2}
    \end{subfigure}
    \begin{subfigure}{0.48\linewidth}
        \includegraphics[width=\textwidth]{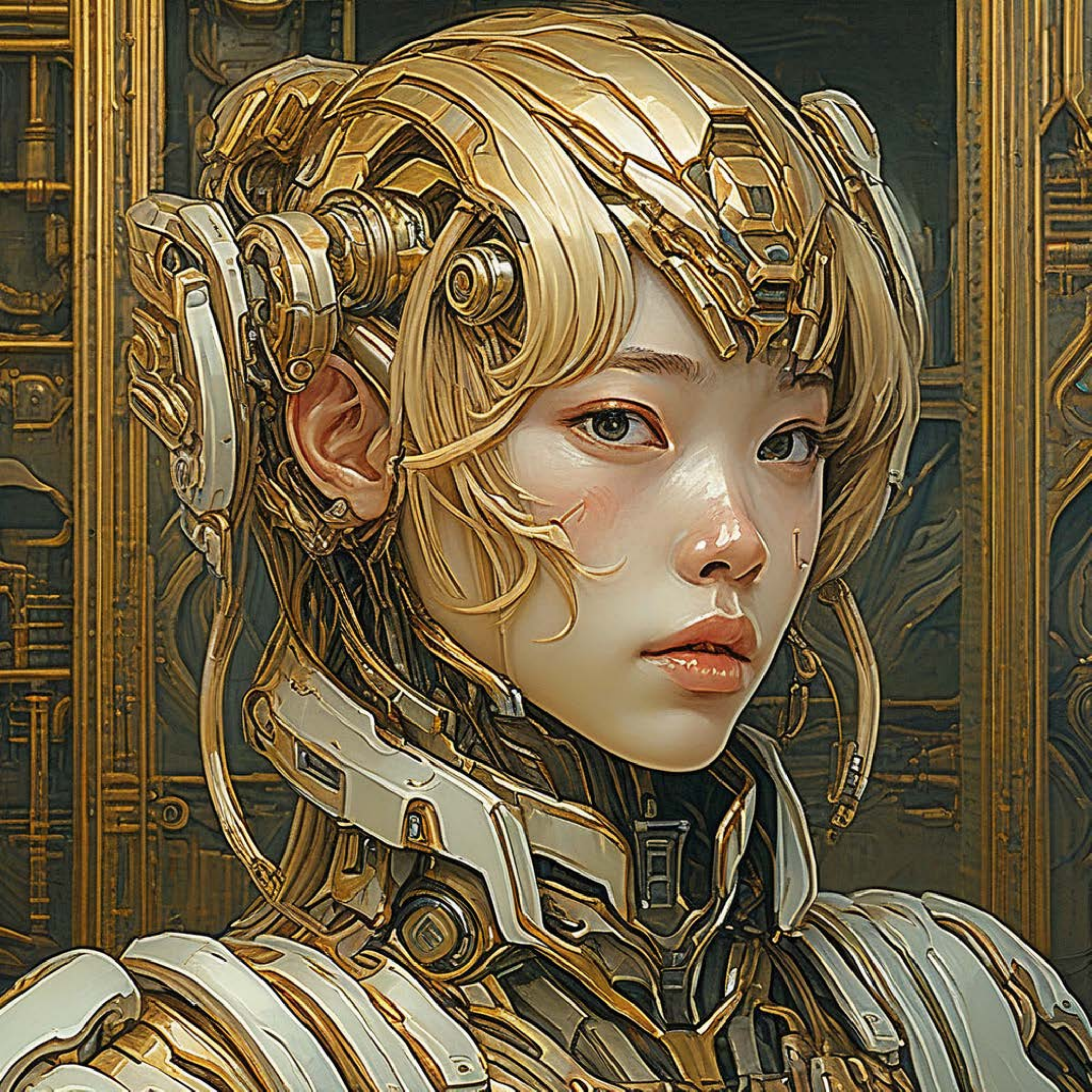}
        \caption{$\beta=32$}
        \label{subfig:beta_32_demo}
    \end{subfigure}
    \begin{subfigure}{0.48\linewidth}
        \includegraphics[width=\textwidth]{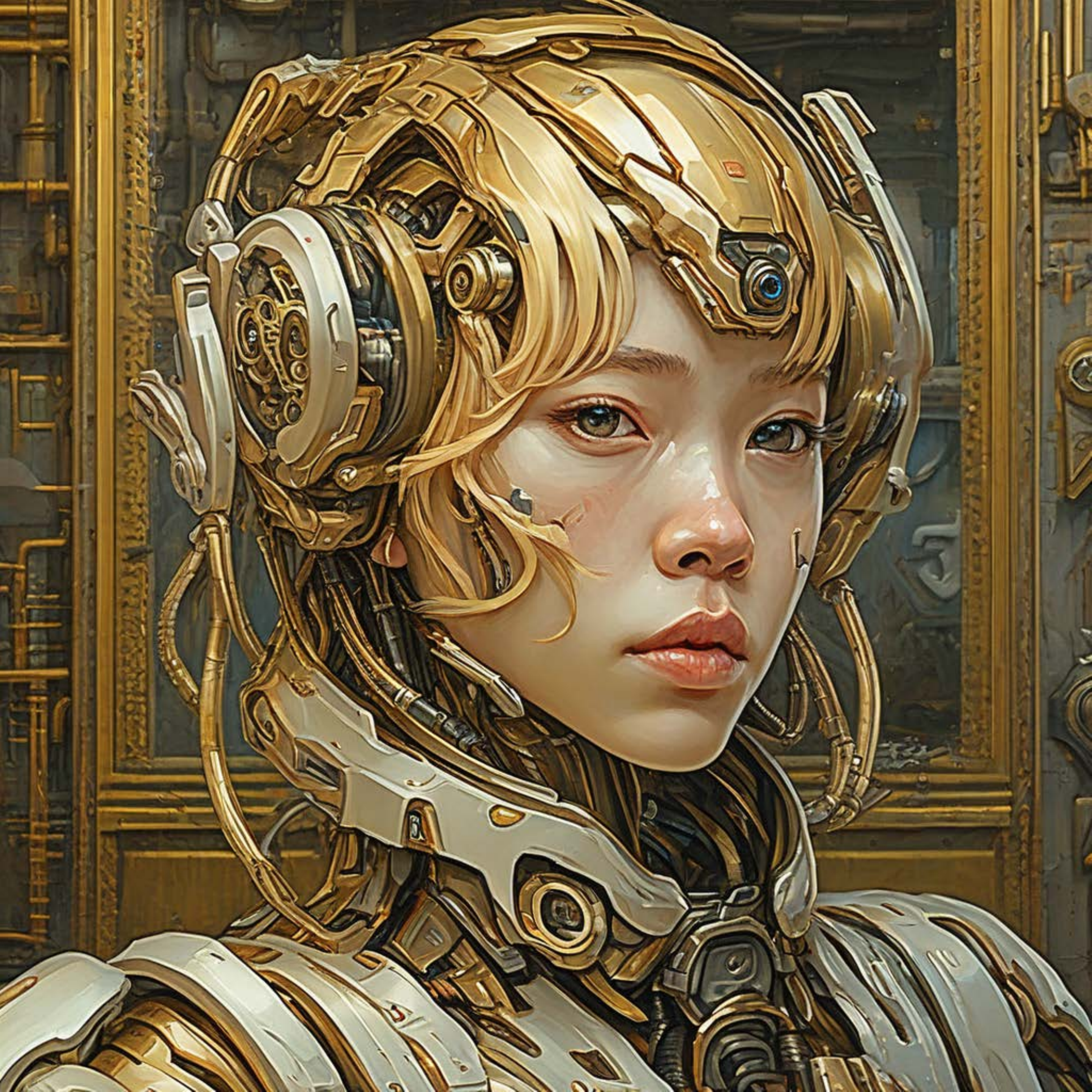}
        \caption{$\beta=64$}
        \label{subfig:beta_64_demo}
    \end{subfigure}
    \begin{subfigure}{0.48\linewidth}
        \includegraphics[width=\textwidth]{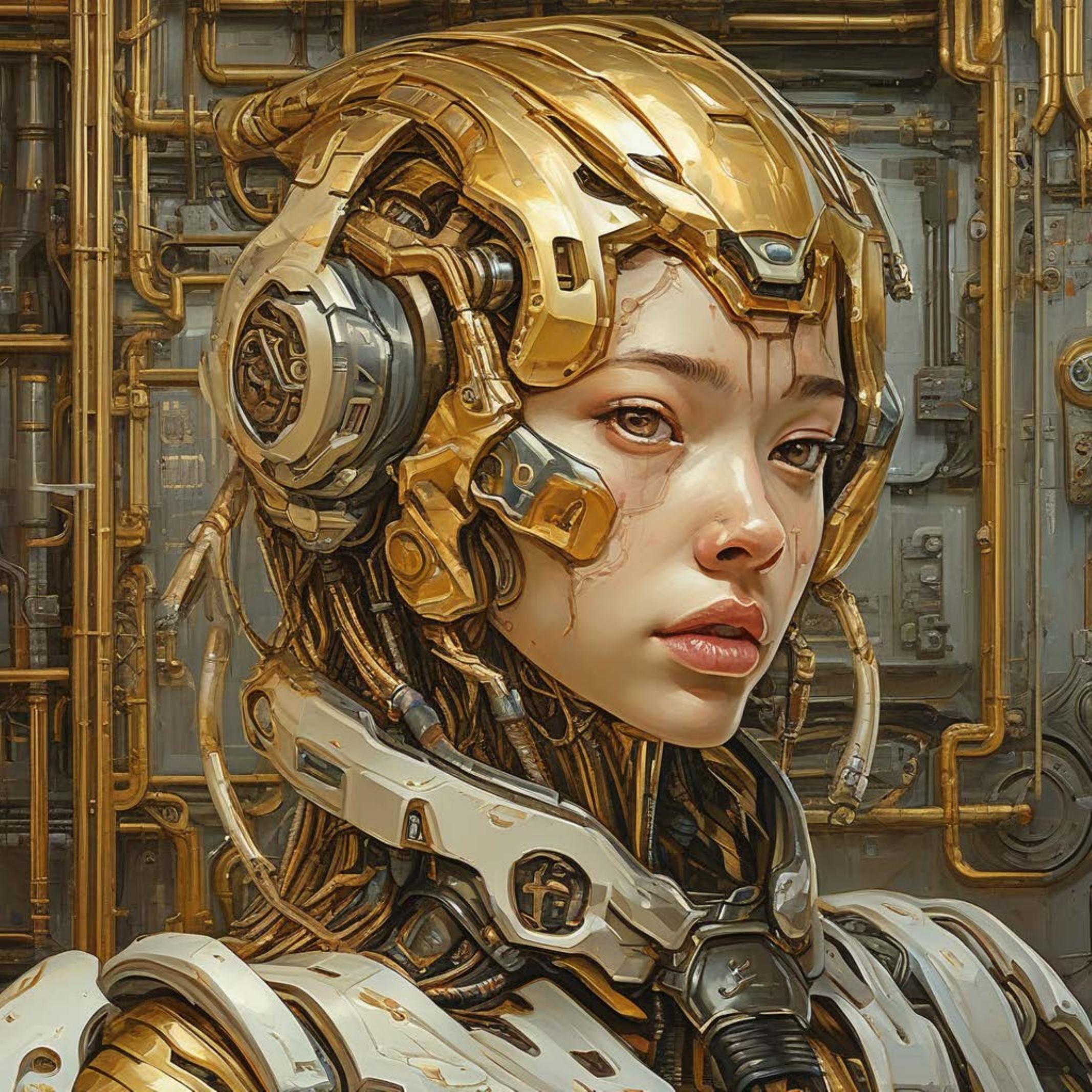}
        \caption{$\beta=128$}
        \label{subfig:beta_128_demo}
    \end{subfigure}
    \caption{Ablation of $\beta$ in $\mathcal{L}_\text{MaPO}$ in \textbf{cultural representation} learning task. Starting from the base SDXL's generation in Figure \ref{subfig:abl_base_2}, the images are generated from MaPO trained with the ascending order of $\beta$. Prompt: \emph{Self-portrait oil painting, a beautiful cyborg with golden hair, 8k}}
    \vspace{-0.1in}
    \label{fig:abl_demo}
\end{figure}

\begin{figure}[hb!]
    \centering
    \begin{subfigure}{0.48\linewidth}
        \includegraphics[width=\textwidth]{assets//apdx/abl/base.pdf}
        \caption{SDXL}
        \label{subfig:abl_base_3}
    \end{subfigure}
    \begin{subfigure}{0.48\linewidth}
        \includegraphics[width=\textwidth]{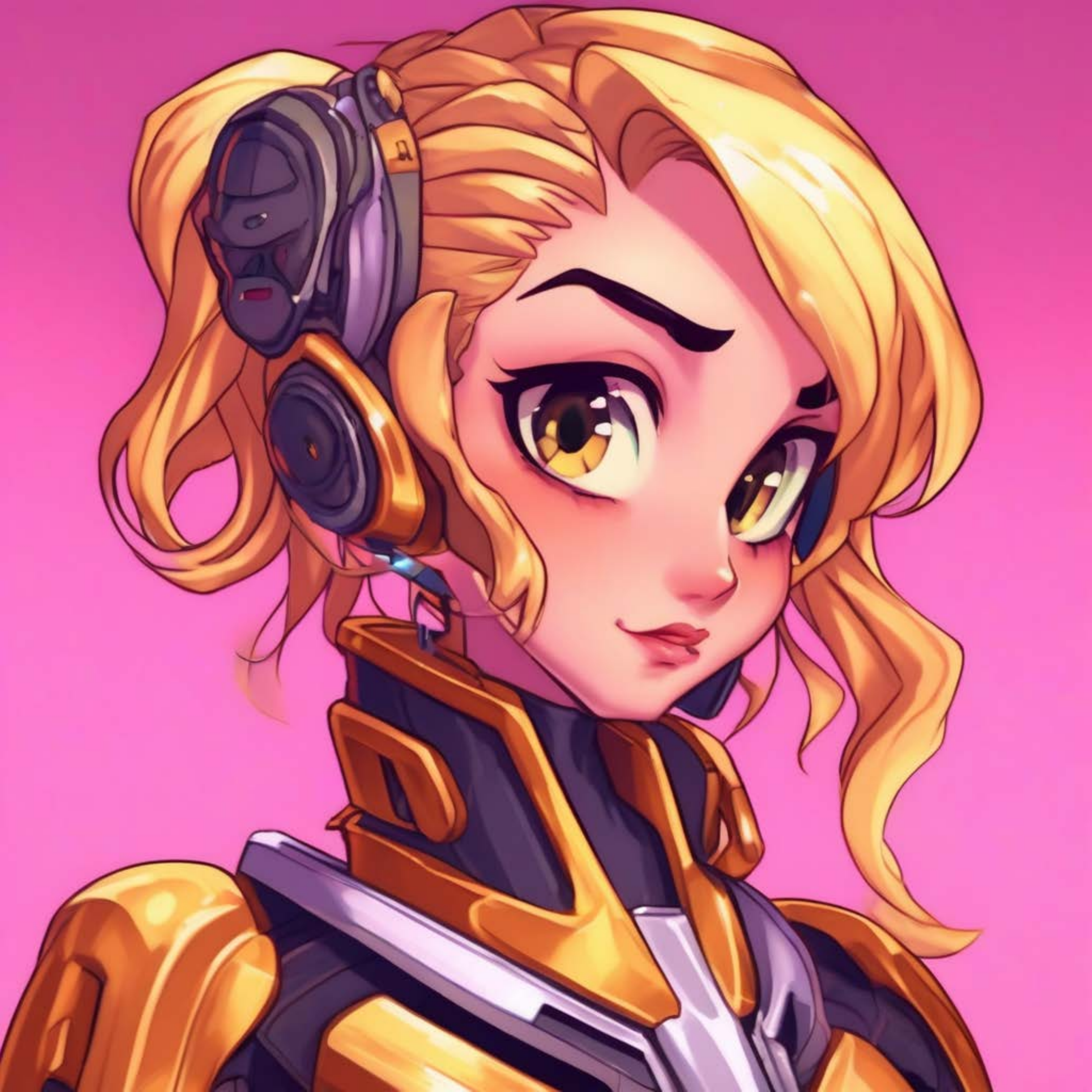}
        \caption{$\beta=32$}
        \label{subfig:beta_32_cartoon}
    \end{subfigure}
    \begin{subfigure}{0.48\linewidth}
        \includegraphics[width=\textwidth]{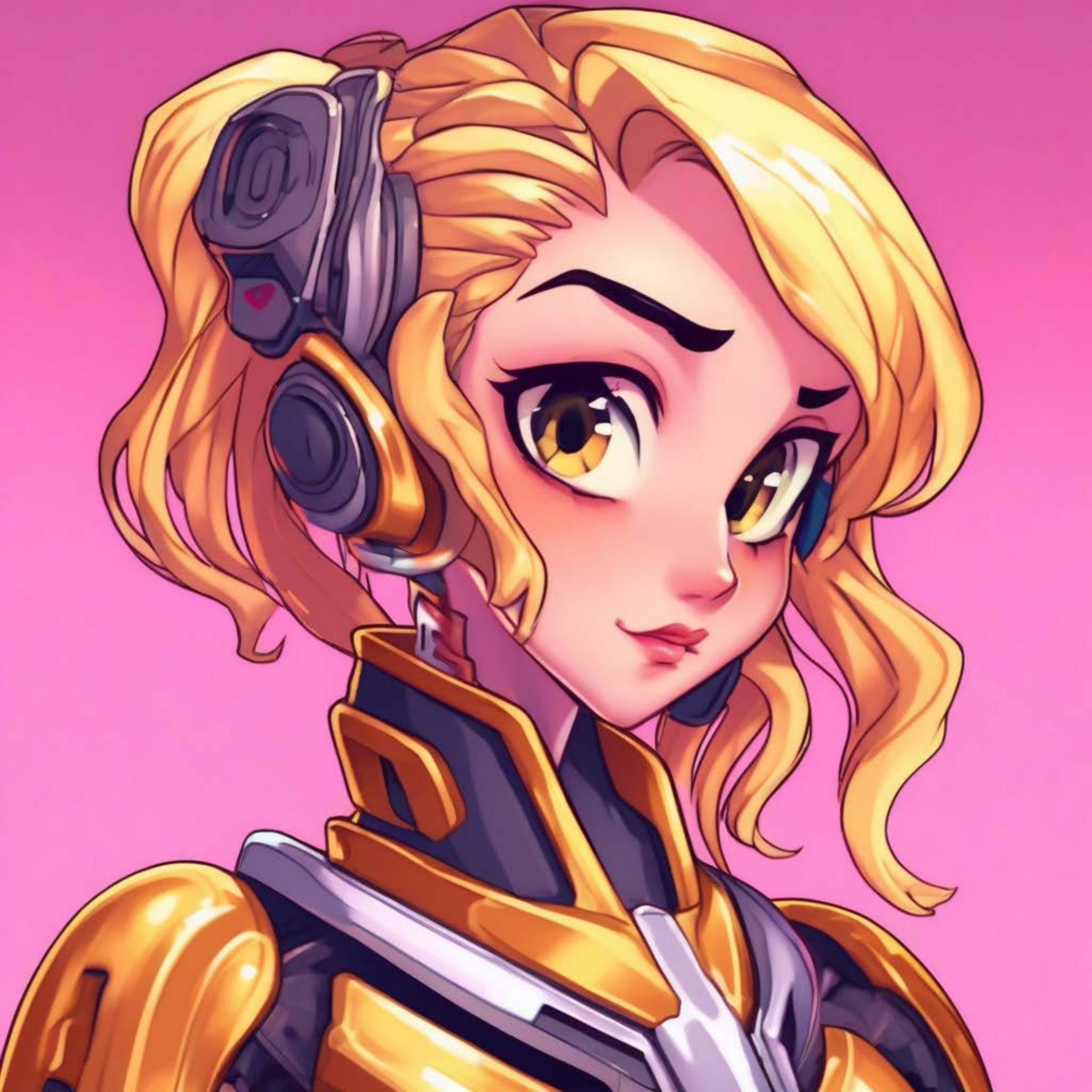}
        \caption{$\beta=64$}
        \label{subfig:beta_64_cartoon}
    \end{subfigure}
    \begin{subfigure}{0.48\linewidth}
        \includegraphics[width=\textwidth]{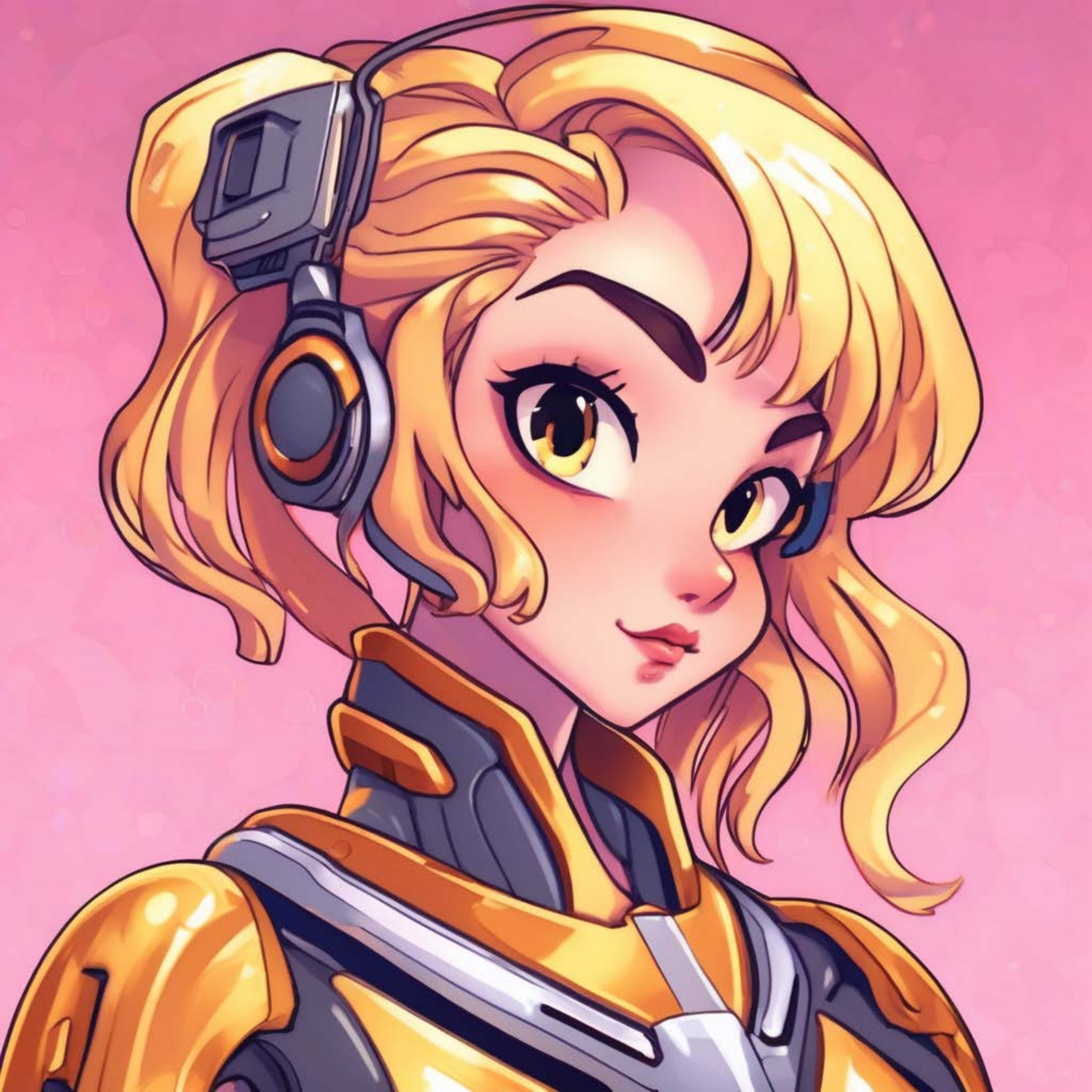}
        \caption{$\beta=128$}
        \label{subfig:beta_128_cartoon}
    \end{subfigure}
    \caption{Ablation of $\beta$ in $\mathcal{L}_\text{MaPO}$ in \textbf{illustrative style} learning task. Starting from the base SDXL's generation in Figure \ref{subfig:abl_base_3}, the images are generated from MaPO trained with the ascending order of $\beta$. Prompt: \emph{Self-portrait oil painting, a beautiful cyborg with golden hair, 8k}}
    \vspace{-0.1in}
    \label{fig:abl_cartoon}
\end{figure}

We provide the qualitative samples that support selecting the optimal $\beta$ in each task in Table \ref{tab:opt-beta}. For five tasks, we provide the fixed SDXL generation and the generations from the MaPO-trained models with three different $\beta$. Figures \ref{fig:abl_general} to \ref{fig:abl_cartoon} demonstrate the gradual differences from increasing $\beta$. $\beta$ values of 8, 64, and 64 are found to be the optimal $\beta$ in each task according to the evaluation metric.

\end{document}